%% file: main.tex
\PassOptionsToPackage{dvipsnames}{xcolor}
\documentclass[acmsmall, nonacm]{acmart}
\usepackage[T1]{fontenc}
\usepackage{scalefnt}
\usepackage{algorithm}
\usepackage{array}
\usepackage{url}
\usepackage{verbatim}
\usepackage{amsmath,amsfonts}
\usepackage{algorithmic}
\usepackage{graphicx}
\usepackage{textcomp}
\usepackage{comment}
\usepackage{xspace} 
\usepackage{booktabs}
\usepackage[nolist,nohyperlinks]{acronym}  
\usepackage{tabularx, array} 
\usepackage{multirow}
\hypersetup{
    colorlinks=true,
    linkcolor=blue,
    citecolor=blue,
    urlcolor=cyan,
  }
\usepackage[caption=false]{subfig}
\usepackage{wrapfig}
\usepackage{siunitx}
\usepackage{enumitem}
\usepackage{colortbl}

\input{content/macros.tex}

\begin{document}

\title[On-device Detection of Wearer-induced NLOS on Chest-worn UWB
Tags: ML, Link Indicators, or Both?]{On-device Detection of
  Wearer-induced NLOS on Chest-worn UWB Tags: Machine Learning, Link
  Indicators, or Both?}

\author{Markus Gallacher}
\authornote{Both authors have contributed equally to this work.}
\email{markus.gallacher@tugraz.at}
\orcid{0009-0000-9456-1092}
\affiliation{%
  \institution{Institute of Technical Informatics, Graz University of Technology}
  \city{Graz}
  \country{Austria}
}
\author{Vu Anh Minh Le}
\authornotemark[1] 
\email{vuanhminh.le@unitn.it}
\orcid{}
\affiliation{%
  \institution{Dept. of Information Eng. and Computer Science, University of Trento}
  \city{Trento}
  \country{Italy}
}

\author{Matteo Trobinger}
\email{matteo.trobinger@unitn.it}
\orcid{}
\affiliation{%
  \institution{Dept. of Information Eng. and Computer Science, University of Trento}
  \city{Trento}
  \country{Italy}
}

\author{Davide Vecchia}
\email{davide.vecchia@unitn.it}
\orcid{}
\affiliation{%
  \institution{Dept. of Information Eng. and Computer Science, University of Trento}
  \city{Trento}
  \country{Italy}
}

\author{Carlo Alberto Boano}
\email{cboano@tugraz.at}
\orcid{0000-0001-7647-3734}
\affiliation{%
  \institution{Institute of Technical Informatics, Graz University of Technology}
  \city{Graz}
  \country{Austria}
}
\author{Gian Pietro Picco}
\email{gianpietro.picco@unitn.it}
\orcid{}
\affiliation{%
  \institution{Dept. of Information Eng. and Computer Science, University of Trento}
  \city{Trento}
  \country{Italy}\vspace{+7.50mm}
}

\begin{abstract}
\input{content/abstract}
\end{abstract}

\input{content/preamble.tex}

\maketitle

\input{content/intro}
\input{content/background}
\input{content/research_questions}
\input{content/exp_setup}
\input{content/classifiers}
\input{content/eval}
\input{content/accuracy}

\input{content/error}
\input{content/runtime_energy}
\input{content/putting_together}

\input{content/recommendations}
\input{content/conclusions}
\input{content/acks}

\bibliographystyle{ACM-Reference-Format}
\bibliography{references_long}

\end{document}

%% file: content/macros.tex
\newcommand{\newnot}[1]{\textcolor{black}{#1}}

\newcommand{\question}[2]{\noindent{\textit{\textbf{(Q#1)}~#2}}}

\newcommand{\qone}{\question{1}{What is the accuracy of LI and ML
    classifiers under identical settings? How do they compare?}}

\newcommand{\qtwo}{\question{2}{Are LI and ML classifiers
    complementary? If so, can they be integrated effectively?}}

\newcommand{\qsix}{\question{6}{Do radio settings affect the classifier performance? How?}}

\newcommand{\qthree}{\question{3}{Does the amount of body occlusion 
affect classification?}}

\newcommand{\qfour}{\question{4}{What is
    the ranging error of samples classified as LOS?}} 

\newcommand{\qfive}{\question{5}{What is the runtime and energy cost of
    classification?}}

\newcommand{\qseven}{\question{7}{How to choose the ``best'' classifier?}}

\newcommand{\plen}{\texttt{PLEN}\xspace}

\newcommand{\uwb}{UWB\xspace}

\newcommand{\da}{$\Delta A$\,[\%]}
\newcommand{\dr}{$\Delta R$\,[$\mu s$]}
\newcommand{\de}{$\Delta E$\,[$\mu J$]}
\newcommand{\dpnn}{$\Delta\pnn$\,[cm]}

\newcommand{\orient}{{\newnot{\textsc{pose}}}\xspace}
\newcommand{\pose}{\orient}

\newcommand{\random}{\textsc{\newnot{mobile}}\xspace}
\newcommand{\mobile}{\random}

\newcommand{\office}{\textsc{office}\xspace}
\newcommand{\wh}{\textsc{warehouse}\xspace}
\newcommand{\clos}{c\textsc{los}\xspace}
\newcommand{\closlong}{c\textsc{los}\ensuremath{_{1024}}\xspace}
\newcommand{\cnlos}{c\textsc{nlos}\xspace}
\newcommand{\chtwo}{\textsc{ch2}\xspace}
\newcommand{\chfour}{\textsc{ch4}\xspace}
\newcommand{\chtwofour}{\textsc{ch2,\,ch4}\xspace}
\newcommand{\chfive}{\textsc{ch5}\xspace}

\newcommand{\trans}{transitional\xspace}
\newcommand{\pn}{\ensuremath{p90}\xspace}
\newcommand{\pnn}{\ensuremath{p99}\xspace}

\newcommand{\kfold}{\mbox{$k$-fold}\xspace}

\newcommand{\class}[1]{\newnot{\scalefont{0.96}\ensuremath{\mathsf{#1}}}}
\newcommand{\qual}[1]{\ensuremath{_\mathsf{#1}}}
\newcommand{\ind}[1]{\ensuremath{\mathit{#1}}}
\newcommand{\ifsym}{\class{if}}
\newcommand{\ifcombo}[2]{\class{\ifsym\qual{#1,#2}}}
\newcommand{\andsym}{\class{\wedge}}
\newcommand{\andcombo}[2]{\class{\andsym\qual{#1,#2}}}

\newcommand{\gtlabel}{\class{Labeling}\xspace}

\newcommand{\pdtxt}{PD}
\newcommand{\cltxt}{CL}
\newcommand{\clnltxt}{\ensuremath{\cltxt^{*}}}%
\newcommand{\litxt}{\ensuremath{LI}}

\newcommand{\pd}{\newnot{\ind{\pdtxt}}\xspace}
\newcommand{\cl}{\newnot{\ind{\cltxt}}\xspace}
\newcommand{\clnl}{\newnot{\ind{\clnltxt}}\xspace}

\newcommand{\pdthr}{\class{\ensuremath{\pdtxt\qual{0.5}}}\xspace}
\newcommand{\clthr}{\class{\ensuremath{\cltxt\qual{0.5}}}\xspace}
\newcommand{\clnlthr}{\class{\ensuremath{\clnltxt\qual{0.5}}}\xspace}
\newcommand{\lithr}{\class{\ensuremath{\litxt\qual{0.5}}}\xspace}

\newcommand{\xgb}{\class{XGBoost}\xspace}
\newcommand{\rem}{\class{REMNet}\xspace}

\newcommand{\pdxgb}{\class{\ensuremath{\pdtxt\qual{XGB}}}\xspace}
\newcommand{\clxgb}{\class{\ensuremath{\cltxt\qual{XGB}}}\xspace}
\newcommand{\clnlxgb}{\class{\ensuremath{\clnltxt\qual{XGB}}}\xspace}
\newcommand{\lixgb}{\class{\ensuremath{\litxt\qual{XGB}}}\xspace}

\newcommand{\ifelsepdxgb}{\ifcombo{\pdtxt}{XGB}\xspace}
\newcommand{\ifelsepdrem}{\ifcombo{\pdtxt}{REM}\xspace}
\newcommand{\ifelseclxgb}{\ifcombo{\cltxt}{XGB}\xspace}
\newcommand{\ifelseclrem}{\ifcombo{\cltxt}{REM}\xspace}
\newcommand{\ifelseclnlxgb}{\ifcombo{\clnltxt}{XGB}\xspace}
\newcommand{\ifelseclnlrem}{\ifcombo{\clnltxt}{REM}\xspace}

\newcommand{\xgbclnlpd}{\class{XGB\qual{\pdtxt,\clnltxt}}\xspace}

\newcommand{\pdandxgb}{\andcombo{\pdtxt}{XGB}\xspace}
\newcommand{\clandxgb}{\andcombo{\cltxt}{XGB}\xspace}
\newcommand{\clnlandxgb}{\andcombo{\clnltxt}{XGB}\xspace}
\newcommand{\pdandrem}{\andcombo{\pdtxt}{REM}\xspace}
\newcommand{\clandrem}{\andcombo{\cltxt}{REM}\xspace}
\newcommand{\clnland}{\andcombo{\clnltxt}{REM}\xspace}
\newcommand{\clnlandrem}{\clnland}

\newcommand{\rss}{\ensuremath{\mathit{RSS}}\xspace}
\newcommand{\fppl}{\ensuremath{\mathit{FPPL}}\xspace}
\newcommand{\prnlos}{\ensuremath{\mathit{PrNLOS}}\xspace}
\newcommand{\luep}{\ensuremath{\mathit{LUEP}}\xspace}
\newcommand{\mc}{\ensuremath{\mathit{M_C}}\xspace}

\newcommand{\degree}{\ensuremath{^{\circ}}\xspace}
\newcommand{\rangl}{\ensuremath{\theta}\xspace}
\newcommand{\nlosth}{\ensuremath{\theta_{\mathit{th}}}\xspace}

\newcommand{\errbin}{\ensuremath{\varepsilon_{\mathit{bin}}}\xspace}
\newcommand{\errth}{\ensuremath{\varepsilon_{\mathit{th}}}\xspace}

\newcommand{\rectool}{\textsc{\mbox{Libra}}\xspace}
\newcommand{\libra}{\rectool}
\newcommand{\recwg}[2]{\ensuremath{(#1\,{:}\,#2)}\xspace}

\newcommand{\conf}[2]{{#1{}\,|\,#2}\xspace}

\begin{acronym}
    \acro{NLOS}{non-line-of-sight}
    \acro{LOS}{line-of-sight}
    \acro{WNLOS}{human non-line-of-sight}
    \acro{MAE}{mean absolute error}
    \acro{ML}{machine learning}
    \acro{LI}{link indicators}
    \acro{SS-TWR}{single-sided two-way-ranging}
    \acro{DS-TWR}{double-sided two-way-ranging}
    \acro{PSR}{preamble symbol repetition}
    \acro{UWB}{ultra-wideband}
    \acro{SVM}{support-vector machine}
    \acro{CIR}{channel impulse response}
    \acro{DL}{deep learning}
\end{acronym}

\newcommand{\vs}{vs.\xspace}

\newcommand{\wrt}{w.r.t.\xspace}
\newcommand{\ie}{i.e.,\xspace}
\newcommand{\eg}{e.g.,\xspace}

\newcommand{\opt}{OptiTrack\xspace}

\newcommand{\sstwr}{SS-TWR\xspace}

\newcommand{\mdek}{MDEK1001\xspace}

\newcommand{\pnnval}[1]{\ensuremath{\pnn\,{=}\,#1}\,cm\xspace}
\newcommand{\cmval}[1]{\ensuremath{#1}\,cm\xspace}
\newcommand{\perval}[1]{\ensuremath{#1}\%\xspace}

\newcommand{\fakepar}{}
\newcommand{\boldpar}[1]{\vspace{0.50mm}\noindent\textbf{#1}}
\newcommand{\italicpar}[1]{\vspace{0.50mm}\noindent\textit{#1}}

\usepackage{paralist}
\setdefaultenum{\em i)}{(a)}{1.}{A.}
\makeatletter
\renewcommand*{\p@section}{\S}
\renewcommand*{\p@subsection}{\S}
\renewcommand*{\p@subsubsection}{\S}
\makeatother

\usepackage{ifthen}
\newboolean{authnotes}

\setboolean{authnotes}{true}
\setboolean{authnotes}{false}

\ifthenelse{\boolean{authnotes}}
{

\newcommand{\vule}[1]{{\color{orange}\footnote{{\bf {\color{orange} Vu}: #1}}}}
\newcommand{\matteo}[1]{{\color{orange}\footnote{{\bf {\color{orange} Matteo}: #1}}}}
\newcommand{\davide}[1]{{\color{orange}\footnote{{\bf {\color{orange} Davide}: #1}}}}
\newcommand{\gp}[1]{{\color{orange}\footnote{{\bf {\color{orange} GP}: #1}}}}
\newcommand{\mg}[1]{{\color{orange}\footnote{{\bf {\color{orange} Markus}: #1}}}}
\newcommand{\cab}[1]{{\color{orange}\footnote{{\bf {\color{orange} Carlo}: #1}}}}
\newcommand{\todo}[1]{{\color{red}[TODO@#1]:}}

\usepackage{draftwatermark}
\SetWatermarkLightness{0.85}
}
{
\newcommand{\vule}[1]{}
\newcommand{\matteo}[1]{}
\newcommand{\davide}[1]{}
\newcommand{\gp}[1]{}
\newcommand{\mg}[1]{}
\newcommand{\cab}[1]{}
\newcommand{\todo}[1]{}         
}

\usepackage{tabularx,ragged2e}
\newcolumntype{B}{>{\RaggedRight}X}    
\newcolumntype{C}{>{\Centering  \hsize=.35\hsize}X}
\newcolumntype{V}{>{\Centering  \hsize=.1\hsize}X}
\newcolumntype{A}{>{\Centering  \hsize=.45\hsize}X}
\newcolumntype{L}{>{\Centering  \hsize=.5\hsize}X}
\newcolumntype{T}{>{\Centering  \hsize=.6\hsize}X}

\newcolumntype{M}[1]{>{\centering\arraybackslash}m{#1}}
\newcolumntype{P}[1]{>{\arraybackslash}m{#1}}

\usepackage{tikz}


%% file: content/abstract.tex
Ultra-wideband (UWB) enables high-accuracy distance estimation
(\emph{ranging}), but performance degrades in non-line-of-sight (NLOS)
conditions. When UWB tags are worn, the human body inevitably causes a
\emph{wearer-induced NLOS (WNLOS)}, yielding significant errors;
still, despite its relevance, WNLOS remains understudied. Moreover,
WNLOS detection typically must occur \emph{on-device} at the tag, requiring
classification accuracy to be balanced against tight runtime and
energy constraints often hard to meet with machine learning (ML)---the
dominating NLOS classification approach. Meanwhile, UWB radio link
indicators (LI) offer a faster and lightweight alternative that has
rarely been explored.
We study existing on-device ML and LI classifiers when applied to
WNLOS, and propose novel \emph{hybrid} ones exploiting their
complementarity. We focus on chest-worn tags and analyze
21~classifiers on a dataset of 1.5~million samples across channels,
PHY settings, environments, subjects, and orientations. Results
show that while all of them achieve good classification accuracy,
relative performance, including ranging errors due to
misclassification, depends heavily on channel and PHY
configuration. Runtime and energy efficiency also vary dramatically;
some techniques offer marginal accuracy gains while consuming up to
$39.6\times$ more energy. Therefore, we also introduce a recommender tool to
identify optimal classifiers under user-defined constraints, publicly
released with our dataset.


%% file: content/preamble.tex
\renewcommand{\shortauthors}{M. Gallacher et al.}

\ccsdesc[500]{Networks~Location based services}
\ccsdesc[500]{Human-centered computing~Ubiquitous and mobile computing}
\ccsdesc[300]{Human-centered computing~Empirical studies in ubiquitous and mobile computing}
\ccsdesc[300]{Computer systems organization~Sensor networks}
\ccsdesc[100]{Computing methodologies~Classification and regression trees}
\ccsdesc[100]{Computing methodologies~Neural networks}

\keywords{Body occlusion, Classification, Energy efficiency, Line-of-sight~(LOS), Link indicator, Machine learning, Non-line-of-sight~(NLOS), OptiTrack, DW1000, Ranging, Ultra-wideband.}

\setcopyright{none} 

\settopmatter{printacmref=false}


%% file: content/intro.tex
\vspace{+7.50mm}
\section{Introduction}
\label{sec:intro}

Ultra-wideband~(UWB) has become a reference technology for
location-aware Internet of Things applications in indoor and GNSS-denied environments. 
Its integration into consumer devices, such as smartphones and smartwatches, has recently driven wider adoption and significant market growth~\cite{marketandmarkets}, enabling disruptive UWB-based applications including keyless entry systems~\cite{kalyanaraman2020keyless}, secure payments~\cite{UWBAlliance_Applications}, robot navigation~\cite{2016_ledergerber_robot, yu2021robot}, industrial process optimization~\cite{FiRaConsortium_UseCases}, and social interaction tracking~\cite{2020_biri_socitrack}. 
By transmitting signals over a wide bandwidth, UWB
enables to estimate with high-resolution the signal's time of
arrival and corresponding time of flight, enabling
\textit{decimeter-level} distance measurements (ranging) and
localization---a performance that narrowband wireless technologies
such as ZigBee, Bluetooth, and \mbox{Wi-Fi} cannot
match~\cite{Leitch_2023_OnIndoorLocalization}.

However, this accuracy is typically attained in \textit{\ac{LOS}} conditions, \ie when no obstacles obstruct the direct propagation path between the transmitting and receiving devices. 
Instead,~in~cluttered indoor environments, where walls, doors, furniture, or other obstacles obstruct the direct path---\ie in \textit{\ac{NLOS}} conditions---the UWB signal may be attenuated or completely blocked, leading to ranging and localization errors on the order of several decimeters to meters, depending on the obstruction severity and nature~\cite{wang23survey, iya2011ObstructedPropogation, marano2010NLOSIdentificationMit, stocker2021PerformanceofSupport, VanNguyen15ML, gallacher2023InSight}. 

\boldpar{Human body occlusion on chest-worn UWB tags: Impact and
  relevance.} Among common obstructing materials, the human body is
known to have a particularly strong impact on UWB signal
propagation~\cite{Ferreira_2021_FeatureSelection,
  makellaraki25material}, primarily attributed to the high water
content of human tissue, which causes substantial signal
absorption~\cite{yusoff09absorption, klemm06absorption}, and
  to the coupling between the body and the antenna, reducing its
  efficiency~\cite{christ06tissue}.
  
  This has been confirmed by studies of the NLOS conditions induced by various materials placed between transmitting and receiving devices, consistently reporting stronger attenuation for human body obstruction than for other common materials like, \eg concrete walls, glass surfaces, or wooden elements~\cite{heydariaan18nlosbenchmark}.

The impact of body occlusion is exacerbated when the UWB device is
\textit{worn} by a person, as the wearer’s own body \emph{directly}
obstructs the UWB link.  We refer to this condition as
\textit{wearer-induced NLOS (WNLOS)}, to distinguish it from the
generic case where the human obstruction occurs in between and away from the two communicating devices---a
scenario that has already been
investigated~\cite{Kim_2023_UWBPositioning,
  Tian_2019_HumanBodyShadowing, Wei_2022_NLOSIdentification,
  Li_2024_HumanBodyImpact} and is outside the scope of this paper.

WNLOS is relevant in many human-centric UWB applications where tags
are often worn on the body or carried 
in a bag or pocket~\cite{sung23position}. Examples include assisted
living systems for fall detection and activity
monitoring~\cite{bregar2021AssistedLiving,
  Rodrigues_2023_WearableSmartSensing}, proximity detection solutions
for social distancing~\cite{cao20206FitAPartProtocolPhysical,
  leoni22covid}, 
geofencing systems for safe human–machine operations in industrial
settings~\cite{locklin20geofencing, Liukkonen25geofencing}, sports
performance \mbox{monitoring for} elite athletes~\cite{Hulka_02092020,
  yang25tennis}, as well as fine-grained tracking and behaviour
analysis of shoppers in retail
environments~\cite{contigiani2016SmartRetail} or visitors at public
exhibitions~\cite{hachem25tosn}. Among the possible placements, UWB
tags worn on the \textit{chest} (\eg an audio guide or proximity
device hanging from a lanyard) are not only a very common case, but
also the one experiencing the most severe signal degradation, as the torso maximizes signal attenuation~\cite{Otim_2019_ImpactofBodyWearable,
  ridolfi_sport2018}. Therefore, in line with other studies~\cite{yogesh2024UWBDistanceEstimationa,
  cao20206FitAPartProtocolPhysical, Mirama_2023_NLOSDetection,
  Han_2025_RobustUWBPositioning, Yang_2025_FuzzyTransformer,
  le22hnlos}, we focus on a chest-worn configuration.

\begin{figure}[!t]
  \centering 
  \includegraphics[width=0.74\columnwidth]{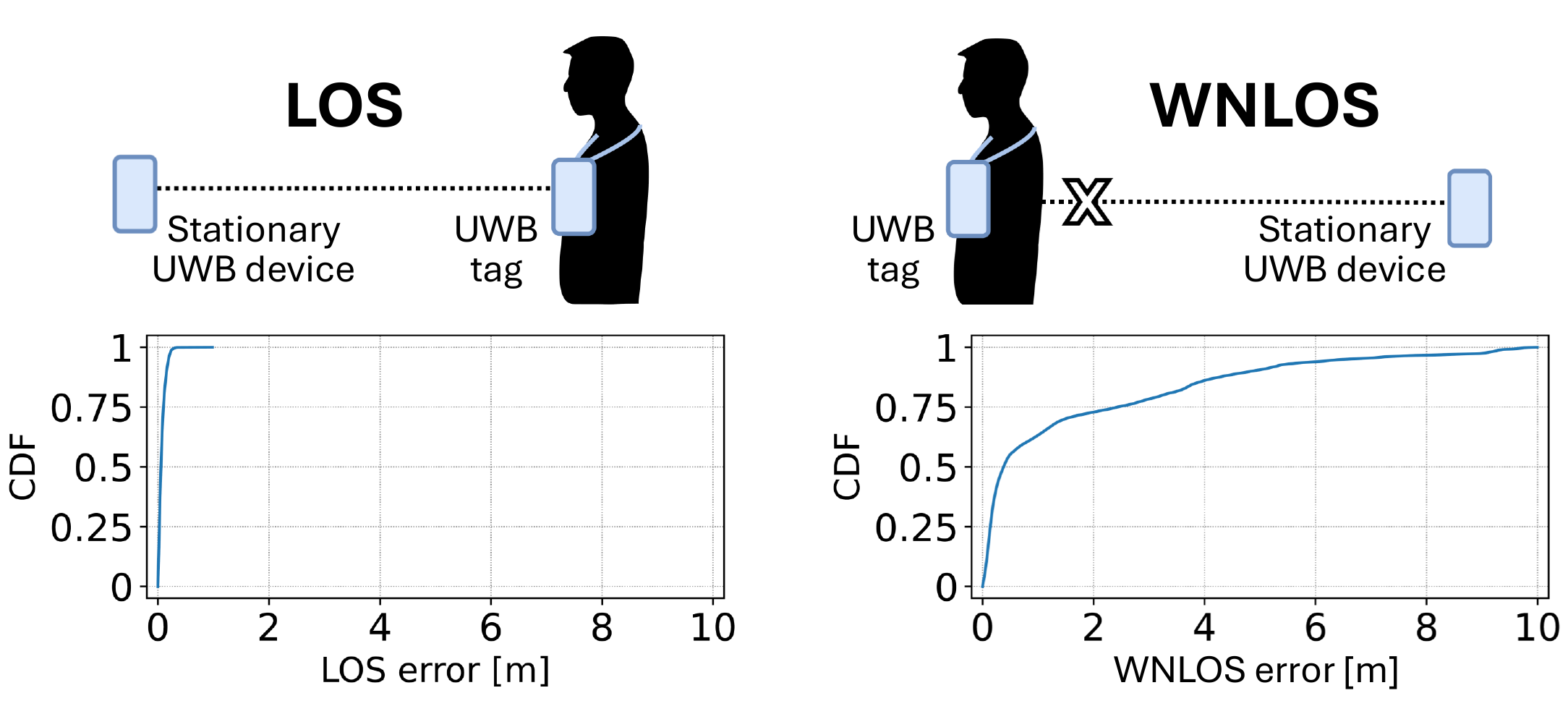}
  \vspace{-4.00mm}
  \caption{NLOS conditions induced by a human body occluding the direct path between UWB devices (WNLOS) can cause meter-level ranging errors, in contrast to centimeter-level errors achieved under LOS conditions.} 
  \label{fig:hnlos_impact}
\end{figure}

To concisely yet concretely illustrate the nature and magnitude of the
problem, we focus on a proximity warning scenario where a person wears a tag\footnote{The tag is worn with a lanyard on the chest of a male subject (1.74\,m, BMI~23.1); the stationary device is mounted on a wooden pole 1.35\,m above ground. Hardware and ranging scheme (\sstwr) are described in~\ref{sec:exp_settings}.} that measures the distance to a stationary device~and triggers an
alert when it gets too close.  The body orientation
(Fig.\,\ref{fig:hnlos_impact}) determines whether the two
communicating devices are in LOS or instead the body of the wearer
obstructs the UWB link, \ie they are in WNLOS. The charts
quantitatively show the dramatic impact of this second case: when the
device is fully obstructed, ranging errors are as high as
$10$\,m,
    \ie two orders of magnitude higher \wrt unobstructed LOS
    conditions where errors are confined to $0.16$\,m at the
    $90^{\text{th}}$ percentile---enough to make the
      proximity warning system severely unreliable and therefore
      potentially unsafe.

\boldpar{Limitations of existing work.}  Despite the strong impact of
wearer's body on UWB ranging and the practical relevance of its
effects, the performance of real-world applications and the
effectiveness of existing NLOS classifiers in the presence of WNLOS
remain poorly understood.

\italicpar{Impact of WNLOS: Lack of systematic empirical evidence.}
Several studies have examined how the presence of a body on the link
line-of-sight, including WNLOS, affects UWB
ranging~\cite{Otim_2019_ImpactofBodyWearable, Kim_2023_UWBPositioning,
  Ferreira_2021_FeatureSelection, Mirama_2023_NLOSDetection,
  sung23position, Han_2025_RobustUWBPositioning,
  Wei_2022_NLOSIdentification, chenC_10422747,
  Tian_2019_HumanBodyShadowing, le22hnlos,
  yogesh2024UWBDistanceEstimationa}. 
However, their results are
difficult to generalize due to inconsistent radio configurations and
limitations in the experimental setup. Indeed, most works focus on a
single UWB channel, often unspecified or different across studies; the
same holds for other PHY~settings\footnote{\label{fn:problem}To convey
  the scale of the problem, the UWB channel used is
  2~\cite{Otim_2019_ImpactofBodyWearable}, 
  3~\cite{Kim_2023_UWBPositioning},
  4~\cite{Ferreira_2021_FeatureSelection},
  5~\cite{Mirama_2023_NLOSDetection, sung23position},
  9~\cite{Han_2025_RobustUWBPositioning}, or is
  unspecified~\cite{Wei_2022_NLOSIdentification, chenC_10422747,
    Tian_2019_HumanBodyShadowing}.  The preamble length varies between
  128 symbols~\cite{Wei_2022_NLOSIdentification,
    Han_2025_RobustUWBPositioning} and 1024
  symbols~\cite{Ferreira_2021_FeatureSelection,
    Kim_2023_UWBPositioning}, and is often
  unspecified~\cite{Otim_2019_ImpactofBodyWearable,
    Mirama_2023_NLOSDetection, chenC_10422747,
    Tian_2019_HumanBodyShadowing}.  The LDE noise multiplier
  (\ref{sec:sec:uwb_bg}) is reported and studied only
  in~\cite{le22hnlos}, despite its careful fine-tuning being highly
  recommended by chip manufacturers~\cite{dw1000-nlos-2}.}. 
Moreover, body occlusion introduces varying degrees of signal
attenuation depending on the person's 
orientation; yet, this factor has rarely been studied in detail. 
This holds also for the subset of studies focusing specifically on
WNLOS conditions~\cite{Otim_2019_ImpactofBodyWearable,
  Kim_2023_UWBPositioning, Tian_2019_HumanBodyShadowing,
  chenC_10422747, Ferreira_2021_FeatureSelection,
  Mirama_2023_NLOSDetection, le22hnlos}, which mostly rely on data
collected by a \emph{single} person at predefined waypoints with fixed
body orientation (\eg $0^{\circ}$, $90^{\circ}$, $180^{\circ}$,
$270^{\circ}$)---too coarse-grained compared to the relative
  position of body and link in real scenarios. In general, these
limitations typically stem from the labor-intensive nature of data
collection and the lack of ground-truth equipment capable of
accurately tracking the person's heading. 

\italicpar{Detecting WNLOS: No comparison among
    classifiers.}  Several research works have explored how to
classify UWB links and distinguish between LOS and (W)NLOS conditions,
critical for discarding erroneous ranging measurements, optimizing
anchor selection, and guiding error~\mbox{correction~(\ref{sec:bg})}.
The literature is dominated~by computationally-heavy \textit{machine
  learning~(ML)} solutions, analyzing the \emph{channel impulse
  response} (CIR,~\ref{sec:bg}) using either
manually-crafted~\mbox{features} (\eg support vector machines and
decision trees\mbox{~\cite{marano2010NLOSIdentificationMit,
    stocker2021PerformanceofSupport, VanNguyen15ML}}) or
automatically-learned ones (\eg deep neural
networks~\cite{bregar2018CNN, jiang_9108193, angarano2021REMNet,
  Hwang_2025_UltrawidebandNonLineofSightClassification,
  Deng_2023_UWBNLOSIdentificationAndMitigation}).  However, a few
studies~\cite{Albaidhani16nlos, le22hnlos} showed that LOS and (W)NLOS    
conditions can be distinguished also via the simpler \textit{link
  indicators~(LI)} defined by radio
manufacturers~\cite{dw1000:metrics}, without incurring the
resource-intensive CIR extraction over SPI and subsequent analysis
required by ML. Yet, there is still no quantitative comparison between
the two approaches in the same setting, hindering the understanding of
their limits, benefits, and 
complementarity. 

\italicpar{Supporting application requirements for WNLOS: Accuracy-only
    focus, no on-device support.} Prior studies have focused almost
exclusively on maximizing the \textit{accuracy} of classification,
often neglecting the practical constraints imposed by real-world
\mbox{applications}.  Yet, UWB is used in a wide range of scenarios
with highly diverse requirements that may limit the applicability of
classifiers optimized solely for maximum accuracy.  Human-centric
applications often require 
UWB tags able to perform classification \textit{entirely on-device}
(\eg to preserve user privacy and reduce latency), preventing
computation-intensive classifiers, like many ML-based ones, to offload
the task to a more powerful edge
infrastructure~\cite{Jones23TinyButMighty}.  Industrial
\mbox{geo-fencing} systems often require a \textit{high update rate}
to ensure timely detection of hazards and prevent
accidents~\cite{locklin20geofencing, Liukkonen25geofencing}, making
fast runtime classification a top priority, even if this comes at the
price of lower accuracy.  A similar trade-off is present in
assisted-living applications for elderly
care~\cite{bregar2021AssistedLiving,
  Rodrigues_2023_WearableSmartSensing}, where tags must prioritize
\textit{high energy efficiency} to extend battery life and enable
continuous, reliable monitoring. 
Other applications, \eg fine-grained tracking in shopping malls or
exhibitions~\cite{contigiani2016SmartRetail, hachem25tosn}, demand
highly accurate distance estimates and must therefore \textit{minimize
  the error} of valid ranging samples\footnote{UWB systems often
  consider ranges as valid only when classified as LOS, as these
  typically exhibit lower error. However, due to misclassified
  samples, using a classifier maximizing accuracy does not necessarily
  imply that the ranging error of LOS-classified samples is minimized,
  as we show in~\ref{sec:los-hnlos_class}.}. 
In~contrast, applications where information about facing direction or
mutual visibility is essential, \eg sport monitoring assessing player
attention and positioning~\cite{Hulka_02092020} or safety systems
determining whether workers can see an approaching vehicle or have
their back turned to it~\cite{Liukkonen25geofencing},
require very \textit{high classification accuracy}. 
Unfortunately, existing research has not yet explored how to reconcile
the different and often conflicting priorities of these real-world
applications, leaving application developers without clear guidance on
the best-suited classification methods.

\boldpar{Methodology and contributions.} In this paper, we tackle the
aforementioned gaps in the literature, and offer concrete answers to
the related open research questions, further detailed in
\ref{sec:rq}.  As our target platform, we select the Qorvo
MDEK1001~\cite{qorvo_mdek1001}, based on the DW1000 transceiver,
because it represents a mature and widely adopted UWB solution with a
large installed ecosystem and extensive prior use in real-world
localization systems, while offering access to a wider set of UWB
channels than newer DW3000 variants~\cite{schuh24complex}.

We provide the systematic empirical evidence necessary to elicit and
support our findings via a rigorous experimental methodology yielding
a massive, multi-faceted, and fine-grained dataset. We leverage an
OptiTrack~\cite{optitrack} motion capture system for mm-accurate
ground truth and full angular link characterization
(0--360$^{\circ}$), and collect over 1.5~million ranging samples
across multiple UWB channels and PHY~settings, and~five human subjects
moving in two distinct environments, a warehouse and an office space.
Experiments are performed using \textit{controlled} motion with
360$^{\circ}$ rotations, to exhaustively generate occlusions without
the bias of specific mobility patterns, and~ \textit{unconstrained}
movements of the subjects, to capture the full range of link dynamics
in more realistic conditions.  This mix of experimental rigor and
diversity across environments, human subjects, radio settings, and
motion patterns---a total of 198~distinct configurations---is
unprecedented in the literature~(\ref{sec:exp_settings}).

Moreover, we address the current lack of comparison between
\emph{existing} approaches to WNLOS detection by considering three
LI-based and two ML-based classifiers. These are selected to support
\textit{entirely \mbox{on-device}} execution on off-the-shelf UWB
modules, to reflect the aforementioned requirements of human-centric
applications. Moreover, we contribute \emph{novel} classifiers that
exploit a \emph{hybrid} design, currently unexplored in the NLOS
literature, 
aimed at
exploiting the complementary properties of the pure LI and ML
classifiers via different strategies (\ref{sec:classifiers}).

Our analysis (\ref{sec:los-hnlos_class}), based on the
\emph{quantitative} comparison of the overall \emph{21~classifiers}
applied to our datasets, reveals that \emph{all} of them, including
our hybrid ones, achieve good accuracy, on average between \perval{85.8}
and \perval{93.1}, and up to \perval{\approx{100}} when it matters the
  most, \ie when occlusions are prominent and WNLOS-induced ranging
  errors the highest.  However, results also highlight that accuracy
is strongly affected by the radio configuration, \ie wireless channel
and PHY-level optimizations. Interestingly, the radio configuration
impacts also the residual error of ranging samples classified as LOS,
crucial for applications that discard WNLOS samples to retain accurate
distance estimates~\cite{contigiani2016SmartRetail, hachem25tosn},
albeit in a different way: 
the classifiers yielding the highest detection accuracy are not always
those yielding the lowest error of LOS samples---and vice
versa. Finally, classifiers strongly differ also in their overhead,
with some offering a $1$\% increase in classification accuracy while
incurring a $27.8\times$~longer runtime and $39.6\times$~higher energy
consumption.

These findings, which we distill for the first time in the literature,
are highly valuable for system designers. On the other hand,
identifying the best trade-offs among the several important and
interrelated dimensions we elicited and specific application
requirements 
can be challenging---as we experienced ourselves. Therefore, we
present a recommender tool called \rectool\footnote{Meaning ``scale''
  or ``balance'' in Latin, it evokes the careful weighing of
  alternatives to support quantitative evaluation and optimal
  trade-off selection.}  that, based on given application performance
requirements and their relative importance, uses a multi-objective
decomposition framework to select and rank classifiers in both
graphical and tabular form. This rich output provides developers with
a clear overview of which WNLOS classification techniques best fit
their requirements and why~(\ref{sec:recommendations}).

We publicly release\footnote{\label{fn:links}Upon article acceptance.} the main artifacts in this study, \ie the datasets
of our large experimental campaign and
\rectool.  The latter is
designed to operate with datasets and classifiers beyond those used in
this study enabling, \eg practitioners to integrate their own
measurements collected in their target application environment.
We~believe that these open-source resources, along with the
generality of our methodology, enable valuable opportunities for
follow-up research, \eg performing similar studies with other
obstructing materials, classifiers, or different 
UWB radios~(\ref{sec:conclusion}). 

%% file: content/background.tex
\section{Background and Related Work}
\label{sec:bg}

\noindent
This section provides relevant background on \uwb technology and state-of-the-art classification techniques for (W)NLOS. 

\subsection{UWB Distance Estimation and (W)NLOS Conditions}
\label{sec:sec:uwb_bg}
\noindent
UWB radios like the popular Qorvo DW1000~\cite{dw1000-manual} used in
this work can return cm-accurate ranging estimates in LOS 
conditions~\cite{malajner2015UWBRangingAccuracy}.  This stems from the
high time resolution enabled by UWB’s large bandwidth
($\geq$\,500\,MHz) and short signal pulses ($\approx$\,2\,ns), which
allow a receiving device to acquire a fine-grained CIR.  The latter
contains detailed information about the propagation paths of the
signal, including the direct path between transmitter and receiver,
called \emph{first path} (FP), and other multipath components (MPCs)
originating from reflections in the environment.

\fakepar
CIRs are processed on-chip using a proprietary leading-edge detection~(LDE) 
algorithm to identify the FP of the incoming signal and record its time of arrival (ToA) as a reception (RX) timestamp. 
Coupled with the ability of UWB radios to schedule transmissions (TX) with high precision, these RX timestamps enable accurate estimation of the time of flight (ToF)~$\tau$ between two devices, and their distance $d=c\times\tau$, with $c$ being the speed of light. 

Among the several schemes to estimate distance between devices,
single-sided two-way ranging \mbox{(SS-TWR)} is one of the most
widely used. In SS-TWR, the initiator sends a \texttt{POLL} message to
the responder, which replies with a \texttt{RESP} message after a
predefined delay. Extensions to this basic exchange, such as
double-sided two-way ranging \mbox{(DS-TWR)}, introduce additional
messages to mitigate errors induced by clock drift, at the cost of
increased latency and energy consumption.  Since WNLOS effects
originate from signal propagation, the resulting ranging bias is
primarily determined by the propagation path and is not fundamentally
affected by the choice of two-way ranging scheme. The
clock-related errors, which can be mitigated through additional
ranging exchanges, are indeed negligible compared with the propagation
delays introduced by WNLOS conditions~\cite{dw1000:aps011}.
Therefore, in this work, we use \mbox{SS-TWR} as the reference ranging
procedure, as it provides a representative baseline for studying
WNLOS-induced ranging errors while minimizing communication overhead
and implementation complexity, which are key considerations in
low-power wearable deployments.

\begin{figure}[!t]
    \centering
    \includegraphics[width=0.75\columnwidth]{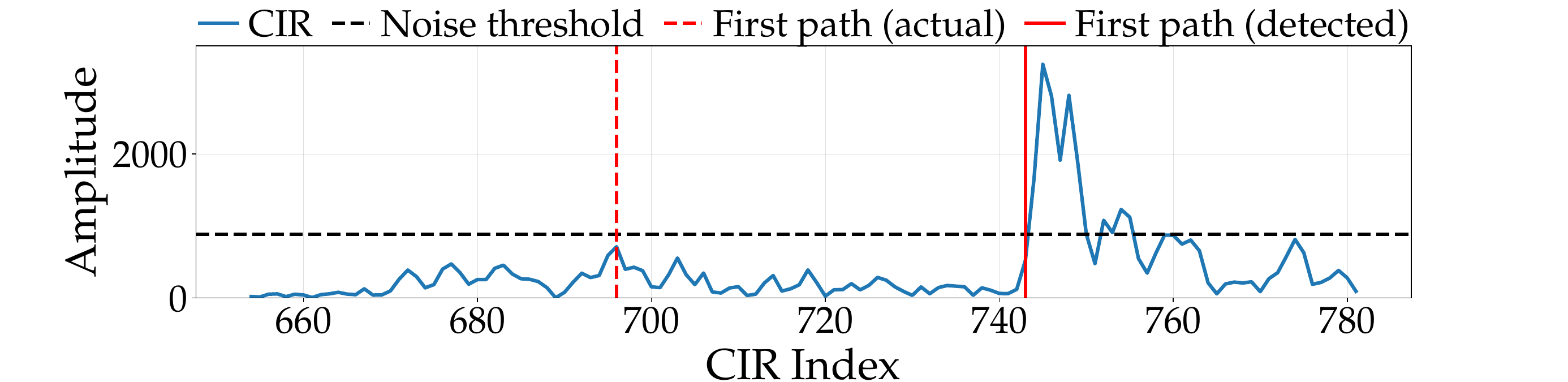}
    \vspace{-2.00mm}
    \caption{Example of a CIR acquired in WNLOS. The LDE algorithm fails~to~detect the true FP (dashed red line) and wrongly selects a later peak (solid red line).}
    \label{fig:cir}
    \vspace{-2.00mm}
\end{figure}

\boldpar{Ranging in NLOS.}  When obstacles block the signal direct
path, UWB ranging accuracy may degrade severely, with errors reaching
up to several meters~\cite{wang23survey, iya2011ObstructedPropogation,
  marano2010NLOSIdentificationMit, VanNguyen15ML,
  stocker2021PerformanceofSupport, gallacher2023InSight}.  The problem
stems from the inability of the LDE algorithm to correctly identify
the FP. Obstacles typically attenuate its peak, causing LDE to lock
onto a later-arriving MPC; this yields an overestimated ToF and
therefore measured distance.  Fig.\,\ref{fig:cir} shows a CIR we
acquired in WNLOS conditions, where the true FP (dashed red) is so
attenuated that it falls below the threshold (dashed black) used by
the LDE to discern valid
peaks from noise.  As a result, a later, stronger reflection (solid red) is
mistakenly selected as FP, introducing a significant
error\footnote{The CIR shown in Fig.\,\ref{fig:cir}, collected from a
  DW1000 radio, has a resolution of $\approx$\,1\,ns. In this case,
  the detected FP appears 48 CIR samples after the true FP, yielding a
  timestamping error of 48.08\,ns and a ranging error of 14.42\,m.}.

\boldpar{Radio optimizations for NLOS.}  UWB radios often provide
configurable parameters enabling optimized performance in NLOS.  In
the DW1000 radio, these are the \textit{noise multiplier}, a
multiplicative factor used to set the noise threshold for LDE, and the
\textit{preamble length} (\plen) of transmitted
frames~\cite{dw1000-nlos-2}.  A~high noise multiplier improves FP
detection when the noise level is high, but can miss attenuated
signals typical~of NLOS conditions (Fig.\,\ref{fig:cir}).  Conversely,
a low multiplier increases sensitivity to weak signals but also the
risk of false detections where noise is mistakenly detected as
FP~\cite{yang24nlos, le22hnlos}.  \plen, used for radio
synchronization and CIR accumulation, determines the number of symbol
repetitions.  A long \plen allows more symbols to be captured and
accumulated in the CIR, improving the reliability of FP detection, but
increases both ranging time and energy consumption.

\subsection{NLOS Classification}
\label{subsec:relwork_hnlos_classification}

\noindent
The poor ranging performance in NLOS can be tackled~by classifying UWB
links to identify whether a ranging exchange took place in LOS or not.
This allows to discard ranging samples taken in
NLOS~\cite{stahlke20nlos}, refine anchor \mbox{selection},~\eg by
picking only LOS anchors~\cite{gallacher2023InSight}, and inform error
\mbox{correction}~\cite{Ember_ImprovingTheCorrection_10767096}. 
Classification typically employs either
\begin{inparaenum}
\item \textit{link \mbox{indicators}}, commonly derived from radio
  diagnostics, or
\item \textit{machine learning} models extracting features from the
  CIR.
\end{inparaenum}

\boldpar{Link indicators (LI).}  Radio manufacturers provide means to
evaluate the state of a wireless link~\cite{dw1000:metrics} to assess
whether it is LOS or NLOS.  Despite their practicality, these LI have
received limited attention. 
In contrast to ML-based approaches, which require substantial
computational resources and large labeled datasets, LI are lightweight
and do not require prior training. They can be derived directly from
radio reception diagnostics or by analyzing only a small CIR
portion.

The LI used with the popular DW1000 radio and considered in this work are the following.

\italicpar{Power difference (\pd).}  This LI compares the received
signal strength (\rss) and the first path power level (\fppl). Given $\pd\,{=}\,\rss\,{-}\,\fppl$, 
Qorvo recommends to consider the link as LOS if $\pd\,{<}\,6$\,dB,~NLOS if $\pd\,{>}\,10$\,dB, and uncertain if
$6\,{\leq}\,\pd\,{\leq}\,10$\,dB~\cite{dw1000-manual}. 
\pd has been commonly used to discern LOS from~NLOS~\cite{chenY_10.1145/3657639, le22hnlos, gururaj2017RssFpplDetection} and as feature
of ML classifiers~\cite{stocker2021PerformanceofSupport, Shalihan_NLOSRanging_2022_9926650, chenC_10422747}. 

\italicpar{Confidence level (\cl).}  This LI is calculated from three
parameters derived from the radio diagnostics and the
CIR~\cite{dw1000:metrics}:
\begin{inparaenum}
\item the \emph{likelihood of undetected early path}~(\luep), computed
  by analyzing 16 CIR samples before the FP,
\item the \emph{probability of NLOS}~(\prnlos), evaluated based on the
  time interval between the detected FP and the CIR peak with the highest amplitude, and 
\item the \emph{saturation of the CIR accumulator} (\mc).
\end{inparaenum}
If~$\luep\,{>}\,0$, the FP may have been missed with the current noise
threshold, indicating NLOS: in this case, \cl is set to 0\,\%, the
minimum.  If $\luep\,{=}\,0$, \prnlos is considered; a high value
implies that the receiver is subject to strong and late reflections,
thus possibly NLOS.  If $\prnlos\,{=}\,0$, \cl is set to 100\%, the
maximum. Otherwise, \mc is inspected; if it indicates saturation, \cl
is set to 100\%, otherwise \cl is set to \mbox{$(100\,{-}\,\prnlos)$\%}.
A~link is then considered as LOS if $\cl\,{=}\,100$\%, NLOS if
$\cl\,{=}\,0$\%, and uncertain otherwise. 
To date, \cl has found relatively little adoption in the literature, either directly~\cite{le22hnlos} or as a feature in ML
classifiers~\cite{Li_2024_NLOSOcclusionRecognition}. 

\italicpar{Modified confidence level (\clnl).}
Le~et~al.~\cite{le22hnlos} have observed that \cl may mistakenly
classify LOS as WNLOS\footnote{The same WNLOS condition we focus on in
  this paper is called \emph{human-induced NLOS (HNLOS)}
  in~\cite{le22hnlos}.} when \luep is triggered by
noise. 
Therefore, they proposed a new variant, \clnl, which considers only
\prnlos and \mc and omits the check on \luep; this also makes \clnl
less computationally expensive than \cl, as no CIR samples need to be
extracted. We consider this variant since it is directly relevant to
our goals.

\boldpar{Machine learning (ML).}  The NLOS classification literature
is dominated by approaches based on ML and using features derived from
radio diagnostics or the CIR, \eg FP amplitude, kurtosis, rise time,
and the above~LI. Two categories exist, determined by how features are
extracted.

\italicpar{Manual feature engineering.}  Many works use engineered
features as inputs to train models like random forests, decision
trees, \mbox{$k$-nearest} neighbors, and support vector machines
(SVM)~\cite{marano2010NLOSIdentificationMit,
  arsuaga2024NovelClassificationMethod, VanNguyen15ML,
  barralNLOSIdentificationMitigation2019,
  stocker2021PerformanceofSupport, Sang_2020_IdentificationOfNLOS}.
Although 30+~features have been proposed, recent studies show
that using only a small subset (\eg the root mean square delay
spread, signal-to-noise ratio, and CIR energy) can substantially
reduce the size of SVM models while maintaining high classification
accuracy~\cite{ember23feature, Zeng_2019_UWBNLOSIdentification,
  Wang_2025_ImprovedUWBBased}.  However, running these models
on-device can take relatively long, even with very few input
features. For instance, SVM inference with just 3 features
requires 
15.9\,ms on the \mdek tag used in this work; feature computation adds
further delay in the order of ms, and increases with the number of CIR
samples processed, typically ${>}100$
for good classification accuracy~\cite{gallacher2023InSight}.

\italicpar{Automatic feature extraction.} Several ML classifiers
rely~instead~on \emph{automatic} feature extraction to identify
structures often not evident to humans.  Deep neural networks~(DNNs),
in particular convolutional neural networks~(CNNs), are widely used
for this purpose~\cite{zeng_8631718, angarano2021REMNet,
  jiang_9108193, flueratoru_9662532, Pei03072024, chenC_10422747,
  Han_2025_RobustUWBPositioning, Wei_2022_NLOSIdentification,
  Hwang_2025_UltrawidebandNonLineofSightClassification,
  Deng_2023_UWBNLOSIdentificationAndMitigation}. 
Although these approaches are computationally-heavy and typically
require powerful edge devices, a few recent works have proposed
DNN-based NLOS classifiers 
that can run on resource-constrained UWB
tags~\cite{angarano2021REMNet, gallacher2023InSight}.  
Angarano~et~al.~\cite{angarano2021REMNet} proposed \rem, a small and
efficient CNN with variable size that automatically extracts features
from CIR samples. 
Although originally designed for error correction, \rem has been
adapted by Gallacher et al.~\cite{gallacher2023InSight} to run NLOS
classification with only 8.4\,kB of RAM and 8.8\,kB of flash
memory. 
The same work also proposed a classifier based on \xgb~\cite{gallacher2023InSight}, a
gradient-boosted decision tree model that builds an ensemble of many
small decision trees, each correcting the errors of the previous ones,
to produce a strong overall predictor.  \xgb is an energy-efficient
alternative to CNNs, as it requires only 4\,B of RAM and 6.7\,kB of
flash memory.  Interestingly, both \rem and \xgb retain their classification accuracy
also when using only few CIR samples~\cite{gallacher2023InSight}, significantly reducing the SPI
overhead compared to retrieving the full CIR---with consequent benefits in terms of shorter runtime and energy efficiency.  
In~this~work, we focus on these
two state-of-the-art models (\ref{sec:classifiers}) because of their
small memory footprint and existing open-source implementations
allowing \emph{full on-device execution} on off-the-shelf \mdek UWB tags.

\subsection{WNLOS Classification} 
\label{subsec:relwork_nlos_vs_hnlos_classification}
\noindent 
All aforementioned classifiers, both LI-based and ML-based, are designed to distinguish between LOS and \textit{generic NLOS conditions}. 
To the best of our knowledge, apart from \clnl~\cite{le22hnlos}, no classifier has been specifically designed to detect WNLOS.

In general, studies addressing the detection of NLOS conditions induced by the human body can be divided into two categories:
\begin{inparaenum}
\item studies evaluating NLOS classifiers across different types of obstructions located \textit{between} two communicating UWB devices, with the human body considered as one of the possible obstacles; and
\item studies specifically considering the WNLOS scenario, in which a
  UWB device is worn on the body, typically on the chest or abdomen. 
\end{inparaenum}

The first category typically evaluates the accuracy of generic NLOS classifiers using the human body alongside other types of obstructions, \eg walls, metal plates, and glass windows. 
The reported classification accuracies are generally \perval{{\approx}90} and vary only slightly across obstructing materials, with differences of \perval{1{-}2} among materials within the same study~\cite{Li_2024_NLOSOcclusionRecognition, Wei_2022_NLOSIdentification}. 
Importantly, none of these studies identifies a classifier that is specifically suited to body-induced NLOS conditions. Moreover, their evaluation primarily focuses on classification accuracy, without jointly considering other aspects that are relevant for practical on-device deployment, such as energy consumption and runtime~(\ref{sec:intro}). 
For example, the multiple-input learning neural network proposed by Wei et al.~\cite{Wei_2022_NLOSIdentification}, which uses both the raw CIR and a time-frequency diagram of the CIR as inputs, is designed for offline execution on a laptop rather than on a resource-constrained UWB tag. 

The second category of works directly investigates WNLOS conditions,
evaluating the accuracy of different classifiers when a UWB tag is
worn on the body.  However, these works focus exclusively on either
ML\footnote{These ML approaches follow the architecture of generic
  NLOS classifiers and do not specifically target WNLOS
  conditions.}~\cite{Bharadwaj_2022_ExperimentalInvestigation,
  Ferreira_2021_FeatureSelection, Mirama_2023_NLOSDetection,
  chenC_10422747, Han_2025_RobustUWBPositioning}, or
LI~\cite{le22hnlos}, and thus do not offer a direct comparison across
these two main classification approaches
(\ref{subsec:relwork_hnlos_classification}).  Moreover, on-device
execution and the associated resource constraints are generally not
treated as primary design requirements.  This leaves open, among
others, the question of which approach provides the best trade-off
between detection performance
and practical deployability for WNLOS detection on resource-constrained UWB devices.

%% file: content/research_questions.tex
\section{Research Questions}
\label{sec:rq}

\noindent
Despite the large body of works studying the impact of human body occlusion on UWB ranging and proposing NLOS classification methods~(\ref{sec:bg}), several questions remain open. We outline them next and provide concrete answers in~\ref{sec:los-hnlos_class} and \ref{sec:recommendations}.

\vspace{+0.50mm} 
\qone
\vspace{+0.25mm}

\noindent
Only a few works compare the accuracy of NLOS classifiers, yet focusing
only on \emph{either} LI or ML approaches~(\ref{subsec:relwork_nlos_vs_hnlos_classification}). 
Le~et~al.~\cite{le22hnlos} study
the suitability of LI approaches in detecting WNLOS,
suggesting they can be a viable alternative to ML methods but offering
no direct comparison.  Other works~\cite{stocker23nlos,
  gallacher2023InSight} quantitatively compare the accuracy of ML
classifiers based on manual \vs automatic feature extraction,
demonstrating the superiority of the latter; however, they do not
include LI in their
analysis. 
The absence of a \emph{direct} comparison between LI and ML
classifiers in the same setting hinders a clear understanding
of their strengths and weaknesses. 
Our experimental results quantitatively answer these questions.

\vspace{+0.50mm}
\qtwo
\vspace{+0.25mm}

\noindent
To the best of our knowledge, no prior work has investigated these
aspects. We explore several strategies for exploiting in synergy the
complementary strengths of LI and ML, resulting in 16~new
\emph{hybrid} classifiers (\ref{subsec:approach_hybrid}) whose
performance we evaluate experimentally with our methodology.

\vspace{+0.50mm}
\qthree
\vspace{+0.25mm}

\noindent
Nearly all prior studies lack equipment capable of accurately tracking
the ground-truth heading of the body \wrt the link status during movement. 
As a result, the link conditions are commonly studied either in
full LOS \vs full body occlusion, as in Fig.\,\ref{fig:hnlos_impact},
or in fixed body orientations (\eg multiples of $90^{\circ}$).  In
this work, we leverage an OptiTrack motion capture system as
in~\cite{chenC_10422747, yogesh2024UWBDistanceEstimationa} to track
$360^{\circ}$ body rotations, enabling us to study how these affect the classifier performance (\ref{subsec:opt}).

\vspace{+0.50mm}
\qfour
\vspace{+0.25mm}

\noindent
One of the primary objectives of classification is to enable the
discarding of ranging samples obtained in (W)NLOS conditions, as these
typically incur large errors.  Existing works commonly assume that
high classification accuracy implies high ranging accuracy for samples
classified as LOS---an assumption we show to be incorrect
(\ref{sec:los-hnlos_class}) by characterizing, for each classifier,
the ranging error of LOS-classified samples, which include both true
LOS and misclassified WNLOS samples. 

\vspace{+0.50mm}
\qfive
\vspace{+0.25mm}

\noindent
Most works tackling NLOS offload classification to \mbox{powerful} edge infrastructure and omit an analysis of associated overheads.  Still,
for body-worn UWB tags performing classification entirely
on-device---the focus of this work---quantifying the system
responsiveness and battery lifetime is crucial. 
Equally important is relating these costs to classification accuracy since,
as shown in our analysis (\ref{sec:los-hnlos_class}), the small improvements offered by a
classifier may come at order-of-magnitude higher overhead.

\vspace{+0.50mm}
\qsix
\vspace{+0.25mm}

\noindent 
Existing studies often neglect the role of radio settings in
classification~\cite{Otim_2019_ImpactofBodyWearable,
  Kim_2023_UWBPositioning, Ferreira_2021_FeatureSelection,
  Mirama_2023_NLOSDetection, sung23position,
  Han_2025_RobustUWBPositioning, Wei_2022_NLOSIdentification,
  chenC_10422747, Tian_2019_HumanBodyShadowing} by focusing on fixed PHY parameters or even omitting their details\footref{fn:problem}. Our experiments
are based on three~channels and three~radio optimizations (\ref{subsec:parameter_space}), enabling
quantitative comparison of their performance across classifiers.

\vspace{+0.50mm}
\qseven
\vspace{+0.25mm}

\noindent
Existing research largely focuses on maximizing classification
accuracy and thus provides little guidance when selecting a classifier
that best reconciles multiple performance requirements, as commonly
required by real-world applications (\ref{sec:intro}). 
In this work, we present \libra, a recommender tool (\ref{sec:recommendations}) that fills this gap. 

%% file: content/exp_setup.tex
\section{Experimental Methodology and Datasets}
\label{sec:exp_settings}

\noindent
We answer these questions experimentally, by adopting a rigorous
methodology that analyzes the impact of human body occlusion on
off-the-shelf UWB chest-worn tags caused by both controlled and free
movements~(\ref{subsec:exp_setup}) across a large parameter space
including multiple environments, test points, subjects, and radio
configurations~(\ref{subsec:parameter_space}).  An \opt motion capture
system is used to acquire \mbox{mm-accurate} ground-truth data and
record the impact of human body rotation over
360$^{\circ}$~(\ref{subsec:opt}), which allows us to accurately label
LOS/WNLOS conditions~(\ref{subsec:data_labeling}).  This extensive
experimental effort produced over 1.5~million ranging estimates,
yielding two high-quality datasets~(\ref{subsec:datasets}) to analyze
the performance of the various classifiers.

\subsection{Hardware and Experimental Setup} 
\label{subsec:exp_setup}

\begin{figure}[!t]
    \centering
    \includegraphics[width=0.50\columnwidth]{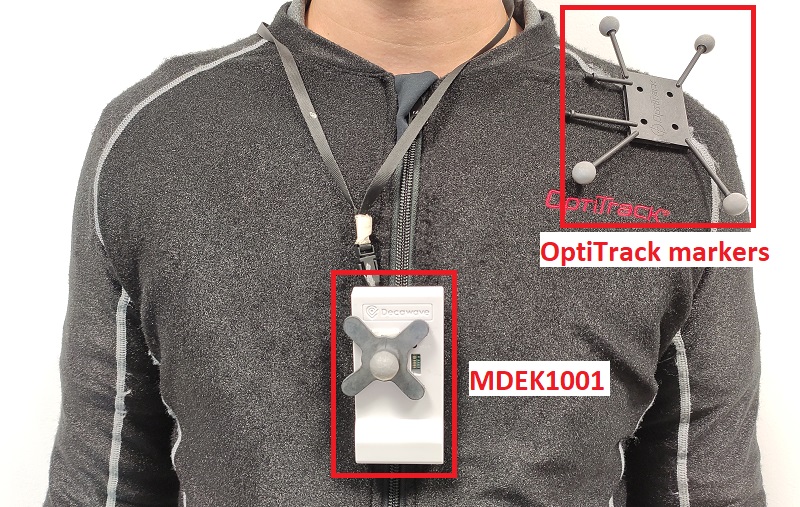}
    \caption{\mdek board (white box) worn on the chest of a person. 
    Nine spherical reflective markers (silver-colored) are used by the \opt system to track the position and orientation of the person. They are mounted on the \mdek board, as well as on the person shoulder and back (not visible).}
    \label{fig:tag_postition}
\end{figure}

\noindent  
  Our setup consists of five Qorvo \mdek UWB development
  kits~\cite{qorvo_mdek1001}, each embedding a DW1000
  radio~\cite{dw1000-datasheet} and an nRF52832~MCU with 64\,kB RAM
  and 512\,kB flash memory. Four of the boards, 
  mounted on a wooden pole at 1.35\,m above ground, serve as
  \mbox{\textit{initiators}} of periodic SS-TWR exchanges with another
  \textit{responder} board, attached to a lanyard and worn as a tag on
  a person's chest (Fig.\,\ref{fig:tag_postition}). 
  We schedule communications so that every ranging exchange occurs in
  isolation \wrt the
  others and ensure absence of external RF interference, \eg from nearby \mbox{Wi-Fi 6E/7}~devices~\cite{brunner22uwbwifi}. 
For each successful ranging, initiators log the estimated distance,
radio diagnostics,~and 
128~CIR samples, starting at 16 before the estimated FP, as for \cl
(\ref{subsec:relwork_hnlos_classification}).

We perform two types of experiments, yielding different datasets.  The
first one (\orient) involves \textit{controlled} body rotation over
360$^{\circ}$ to systematically generate occlusions. The person
wearing the responder stands at a predefined position $P$ and slowly
and continuously rotates around the vertical axis until 100~ranging
samples are collected for each 15$^{\circ}$~angular step per
initiator, yielding 9600 ranging samples per
position. 
The person then moves to another position and the same procedure
repeats.  This controlled setup systematically sweeps all possible
link conditions, from clear LOS to full body occlusion, with a
granularity unmatched in the literature. 

In contrast, the second dataset (\random) captures
\emph{unconstrained} motion and slightly different signal propagation
conditions. The subject wearing the responder tag moves freely along
arbitrary paths within a predefined area until we obtain
3500~successful ranging samples per initiator, yielding a dataset with
a much larger spatial diversity of positions than in \orient.
Initiators are static also in \random, but they are exposed to
different multipath effects yielding different ranging errors, due to
the mobility of the responder tag and to the fact that their position
is changed \wrt\ \orient, enabling the evaluation of classifiers under
different CIR dynamics.

The two datasets are not directly comparable in quantitative terms
(\ref{subsec:datasets}), due to their different setup and goals;
however, their value lies in their complementarity. \orient provides a
highly structured dataset ideal to train ML and hybrid classifiers
(\ref{subsec:training_setup}) across all possible occlusions and
perform in-depth analyses without the bias induced by arbitrary
patterns of movement. \random, used only for testing, offers instead a
less structured yet more realistic mix of distances and occlusions
along with mobility. 
Together, they allow us to offer two perspectives on the complexity of
WNLOS classification.

\begin{table}[!t]
\newcommand{\parcol}{3cm} 
    \centering
    \caption{Parameter space investigated across different experiments.}
    \label{tab:exp_st}
    \vspace{-3.00mm}
    \setlength\tabcolsep{2pt}
    \renewcommand{\arraystretch}{0.95}
    \begin{tabular}{|c|M{\parcol}|M{\parcol}|}
        \hline
        \textbf{Parameter} & \textbf{\orient} & \textbf{\random} \\
        \hline
        Tag positions & 4, fixed & many, variable\\
        \# of subjects & 2 & 3 \\
        Environments & \multicolumn{2}{c|}{\office, \wh} \\ 
        UWB channels & \multicolumn{2}{c|}{\chtwo, \chfour, \chfive} \\
        Radio optimizations & \multicolumn{2}{c|}{\clos, \closlong, \cnlos} \\ 
        \hline
    \end{tabular}%
    \renewcommand{\arraystretch}{1.00} 
\end{table}

\begin{figure}[b!]
\centering
\begin{minipage}[b]{0.62\columnwidth}
    \centering
    \subfloat[\office]{%
        \includegraphics[width=0.475\columnwidth]{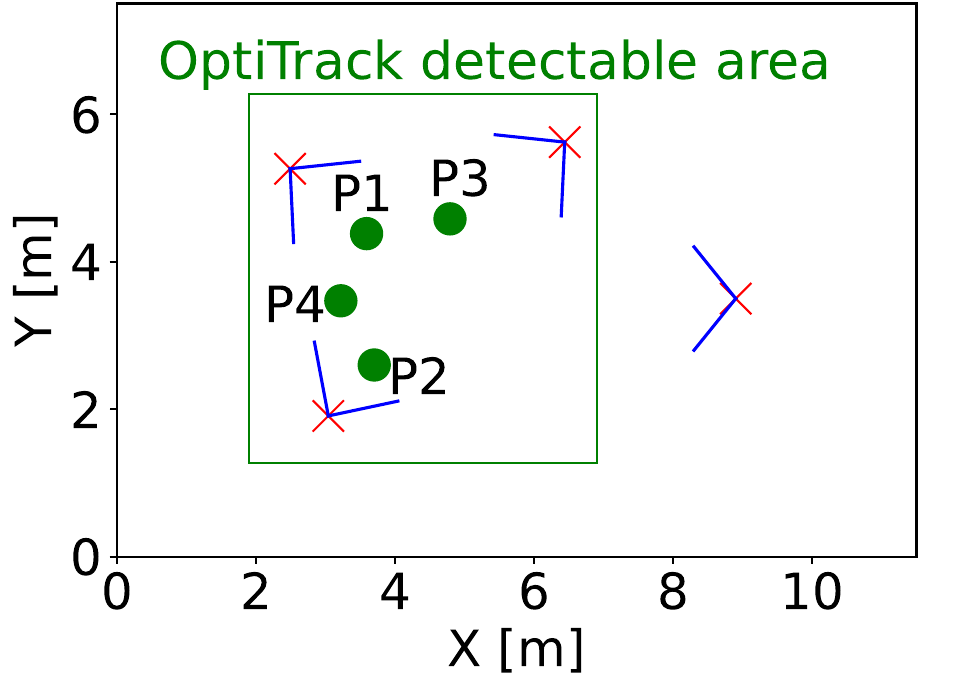}
    }
    \hfill
    \subfloat[\wh]{%
        \includegraphics[width=0.475\columnwidth]{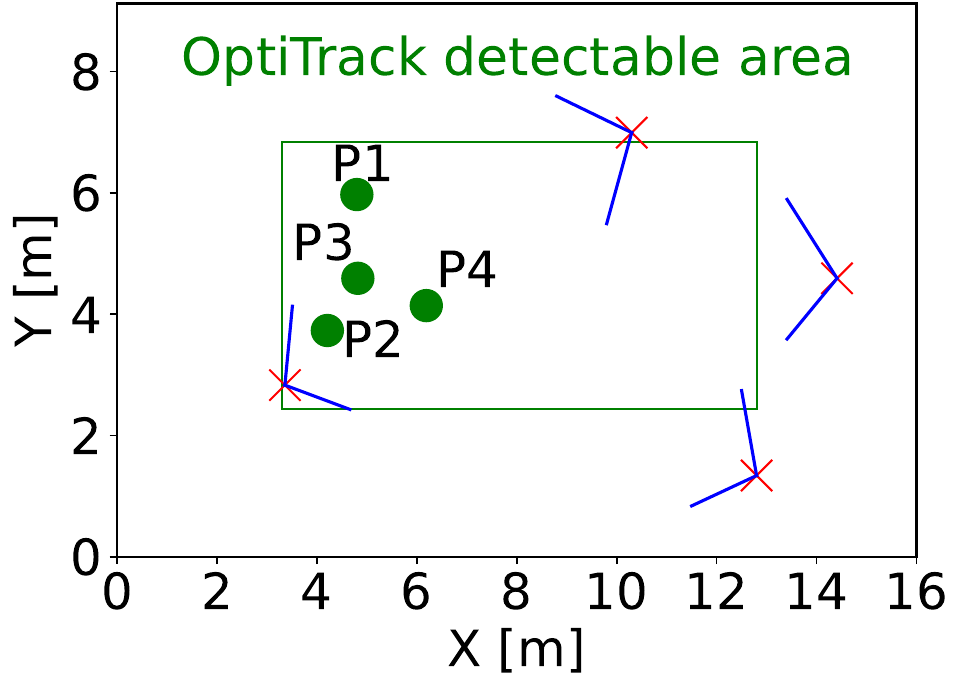}
    }
    \vspace{-3.50mm}
    \caption{Setup used to collect the \orient dataset in two different indoor spaces. Red crosses and green dots mark the fixed positions of initiators and the~4 tag positions (\texttt{P1}--\texttt{P4}) we use, respectively. The chart boundaries match the size of the wall perimeter.}
    \label{fig:orient_floor_map}
\end{minipage}
\hfill
\begin{minipage}[b]{0.2975\columnwidth}
    \centering
    \includegraphics[width=1.0\columnwidth]{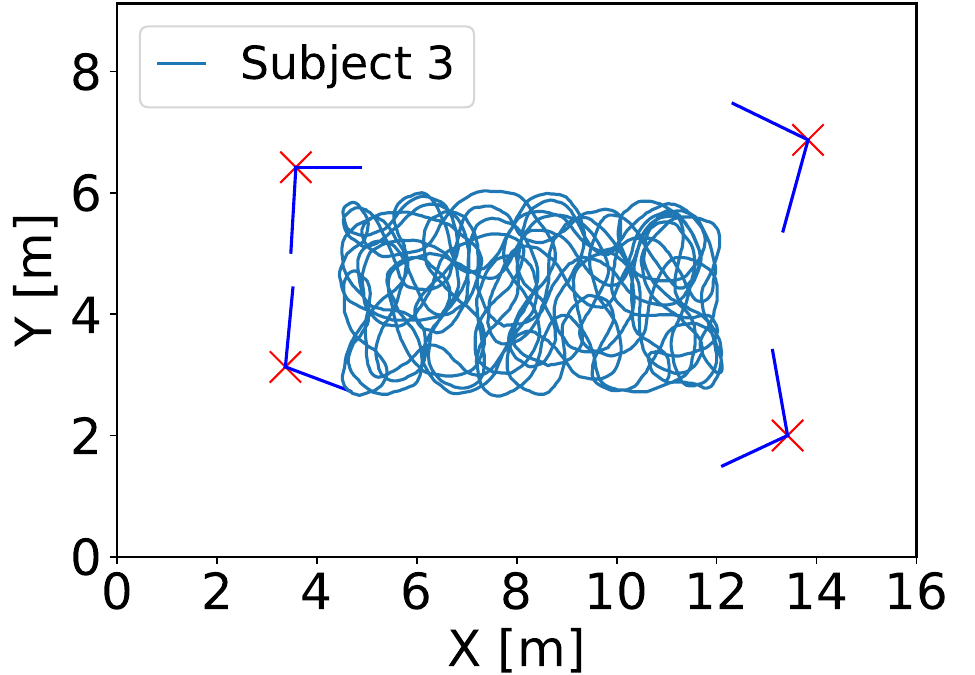}
    \vspace{-6.50mm}
    \caption{A trace from the \random dataset in \wh. The subject wearing the tag moves freely along arbitrary paths within a predefined area. }
    \Description{}
    \label{fig:traces_random}
\end{minipage}
\end{figure}

\subsection{Parameter Space}
\label{subsec:parameter_space}
\noindent
We describe next all the parameters considered in our experiments, also summarized in Table~\ref{tab:exp_st}.

\boldpar{Environments.} We conduct experiments in two indoor spaces
with distinct characteristics. \office is a 7.5\,$\times$\,11.5\,m$^2$
laboratory deployed at our premises.  The limited size and low ceiling
(3.3\,m) of the room make the environment prone to multipath effects.
\wh is a 9.1\,$\times$\,16\,$\times$\,6\,m$^3$ cage used for drone
flight tests, located within a larger warehouse facility.  It allows
testing at longer distances \wrt \office and is arguably subject to
different multipath effects, thus increasing generality.  Both
environments are equipped with an \opt motion capture system
(\ref{subsec:opt}) covering an area of 5\,$\times$\,5\,m$^2$ and
5\,$\times$\,8\,m$^2$, respectively, marked in green in
Fig.\,\ref{fig:orient_floor_map}.

\boldpar{Initiator positions.}  In \orient, we explore $4$~initiator
positions per environment (Fig.\,\ref{fig:orient_floor_map}, red
crosses), chosen to maximize the set of initiator-responder distances
under study while ensuring an even distribution~of~short~and~long~links. 
We also make sure to place the responder within the $\pm45\degree$ field of view of each initiator antenna to achieve a relatively uniform antenna gain, as suggested by the radiation pattern provided by the manufacturer~\cite{dwm1001-datasheet}. 
Instead, in \random, the person wearing the tag moves freely, yielding 
many 
responder positions per experiment
(Fig.\,\ref{fig:traces_random}).

\begin{table}[!t]
    \centering
    \caption{Body characteristics of the persons wearing the tag.}
    \label{tab:body}
    \vspace{-3.00mm}
    \setlength\tabcolsep{4pt}
    \renewcommand{\arraystretch}{0.94}
    \begin{tabular}{|c|c|c|c|c|c|}
        \hline
        \textbf{ID} &\textbf{Dataset} & \textbf{Gender} & \textbf{Height} & \textbf{Weight} & \textbf{BMI}\\
        \hline
         1 &\multirow{2}{*}{\orient} & M & 1.84~m & 100~kg & 29.5 \\
         2 &                         & M & 1.74~m & 70~kg & 23.1  \\
       \hline
         3 &\multirow{3}{*}{\random} & F & 1.54~m & 52~kg & 21.9 \\
         4 &   & M & 1.69~m & 55~kg & 20.2 \\
         5 & & M & 1.90~m & 105~kg & 29.1 \\
        \hline
    \end{tabular}%
    \renewcommand{\arraystretch}{1.00} 
\end{table}

\boldpar{Human subjects.}  Earlier studies have shown that the
severity of WNLOS obstruction is proportional to the build of the
person carrying the UWB tag~\cite{Kumpuniemi14bodysize}.  
Therefore, we carry out experiments with five subjects, different in \orient and \random and with different gender and body features (Table~\ref{tab:body}), who expressed their informed consent to the experiments.  

\boldpar{UWB channels.}  We perform measurements in channel~2,~4,
and~5.  \chfive is recommended by the
manufacturer, and the only one supported also by newer generations of UWB radios, \eg the
DW3000~\cite{dw3000-datasheet}. \chtwo and \chfour are popular choices
due to their central frequency lower than \chfive
($3993.6$~\vs~$6489.6$\,MHz) yielding lower path loss and, in the case
of \chfour, larger nominal bandwidth ($1331.2$~\vs~$499.2$\,MHz),
which enables longer communication range and higher ranging accuracy.

\boldpar{Radio optimizations.}  We consider three configurations
(\ref{sec:sec:uwb_bg}).  \clos, the DW1000 default intended for LOS
conditions, combines the highest noise multiplier with a 128-symbol
\plen.  In contrast, \cnlos tackles NLOS by using the lowest noise
multiplier and a 1024-symbol \plen. 
Finally, \closlong~\cite{le22hnlos} is an intermediate configuration
using a 1024-symbol \plen as in \cnlos, to maximize the
signal-to-noise ratio (SNR) in NLOS, and the highest noise multiplier
as in \clos, to reduce the likelihood of mistaking noise as FP.

\fakepar Concerning specifically \cnlos, our experiments clearly
evidenced that it works well in full NLOS conditions while, in full
LOS, the lowest noise multiplier can easily mistake noise as the
signal FP.  This causes \cnlos to \emph{under}estimate distance in
LOS, to the point of frequently returning impossible \textit{negative}
distances.  This phenomenon was already reported in~\cite{le22hnlos},
but we observed it with a much greater impact.  In some settings, we
could not collect 100~samples per $15^{\circ}$ angular step in LOS;
sometimes, not~even~a~\mbox{single} valid measurement was returned.
This is an important finding per se, which leads us to \textit{advise
  against the use of \cnlos} with chest-worn UWB tags. Consequently, hereafter
we exclude this optimization from our analysis.

\subsection{Acquiring Accurate Ground Truth}
\label{subsec:opt}

\noindent
A cornerstone of our analysis is the use of an \opt motion capture system~\cite{optitrack}, enabling us to establish ground truth with mm-level spatial accuracy and fine temporal resolution (240\,Hz) while tracking UWB tag movements in real time. 

\boldpar{Tracking UWB devices.} 
\opt relies on cameras emitting infra-red~(IR) light into the capture area and track reflections from
passive \textit{markers}.  These are spherical, \mbox{IR-reflective} \mbox{objects}
that, once attached to a target, enable precise tracking of its position and 
orientation.  The data captured by each
camera is sent to the \opt software, Motive~\cite{motive} in our case, which
processes this info to reconstruct the 3D position of each
marker, provided it is detected by at least three cameras.

In theory, a single marker attached to the responder tag should be
enough for \opt to track its movements. 
In practice, as the tag was worn on a person's chest, we observed that the human body often obstructed the view of multiple cameras, causing the \opt system to frequently lose track of the marker and leaving wide gaps in the ground-truth traces.  
To counter this, we slightly reduced the \opt detectable area (Fig.\,\ref{fig:orient_floor_map}) \wrt the perimeter of the
cameras, and used nine~markers: one on the UWB tag, four~on the person's left
shoulder and four~on the back (Fig.\,\ref{fig:tag_postition}). 
This ensured that the cameras in front and behind the person wearing the tag could always illuminate a subset of the markers while the person was moving. 
Then, we exploited Motive to reconstruct a rigid body out of all marker positions, and aligned its centroid with the marker on the responder tag. 
These countermeasures, combined, drastically improved tag tracking, enabling \opt to \emph{always} determine the position of the UWB responder.

As for initiators, they are static; therefore, we simply attached a marker to the device and recorded its position both at the start and end of an experiment, to ensure that no unintentional movement had occurred.

\boldpar{Synchronizing \opt and UWB data.}  To exploit the accurate ground truth provided by \opt, it is crucial to associate each UWB ranging sample with the corresponding positions and relative angles of initiators and responder from \opt. 
This entails:
\begin{inparaenum}
\item defining a common time reference for the \opt and UWB systems,
  and
\item associating each UWB sample with the most
    representative \opt one.
\end{inparaenum}
We accomplish this via a wired, centralized setup where a single PC is
interconnected with all \opt cameras (via Ethernet) and \uwb initiators
(via active USB cables). 
Upon completing a \sstwr round, the initiator immediately dumps the
relevant logs to the PC via SEGGER-RTT~\cite{rtt}, preferred to serial
printing due to its higher communication speed.  There, a custom
Python script precisely timestamps all incoming data from \opt and
initiators.  Finally, once all samples are ordered in time, we
associate each UWB ranging sample to the \opt sample~closest in~time;
this yields accurate temporal ground truth since \opt acquires data at
a much higher rate (240\,Hz) than our UWB setup (12--15\,Hz).

\input{content/labeling}

\begin{table*}[!t]
\footnotesize
\centering
\setlength{\tabcolsep}{4.5pt} 
\caption{Characteristics of \orient and \random datasets. LOS and WNLOS labels are assigned as detailed in \ref{subsec:data_labeling}.}
\label{tab:datasets_characteristics}
\vspace{-3.00mm}
\renewcommand{\arraystretch}{1.04} 
\begin{tabular}{|c|c|c|ccc|cccc|cccc|}
 \hline
\multicolumn{2}{|c|}{ } & & \multicolumn{3}{c|}{\rule{0pt}{2ex}\textbf{LOS and WNLOS}} & \multicolumn{4}{c|}{\textbf{LOS only}} & \multicolumn{4}{c|}{\textbf{WNLOS only}}\\
 \multicolumn{2}{|c|}{\textbf{Radio settings}} & \multicolumn{1}{c|}{\textbf{Dataset}} & distance & \multicolumn{2}{c|}{error} & ratio & distance & \multicolumn{2}{c|}{error} & ratio & distance & \multicolumn{2}{c|}{error}\\
 
\multicolumn{2}{|c|}{ } & & mean & \pn & \pnn &  & mean & \pn & \pnn & & mean & \pn & \pnn\\
\multicolumn{2}{|c|}{ } & & (m) & (m) & (m) & (\%) & (m) & (m) & (m) & (\%) & (m) & (m) & (m)\\
\hline \rule{0pt}{2ex}
\multirow[c]{6}{*}{\clos} & \multirow[c]{2}{*}{\chtwo} & \orient & 4.88 & 1.57 & 9.47  & 58.3 & 4.84 & 0.17 & 0.34 & 41.7 & 4.94 & 4.91 & 10.54 \\
& & \random & 4.32 & 0.53 & 5.91  & 57.9 & 4.29 & 0.21 & 0.45 & 42.1 & 4.36 & 1.40 & 8.85 \\
\cline{2-14} \rule{0pt}{2ex}

& \multirow[c]{2}{*}{\chfour} & \orient & 4.88 & 1.05 & 7.01  & 58.3 & 4.84 & 0.15 & 0.35 & 41.7 & 4.94 & 3.92 & 9.37 \\
& & \random & 4.39 & 0.38 & 3.92  & 57.4 & 4.36 & 0.18 & 0.42 & 42.1 & 4.42 & 0.71 & 6.07 \\
\cline{2-14}\rule{0pt}{2ex}

& \multirow[c]{2}{*}{\chfive} & \orient & 4.88 & 6.16 & 11.85 & 50.0   & 4.83 & 0.11 & 0.19 & 50.0 & 4.93 & 9.13 & 12.64 \\
& & \random & 4.33 & 3.10 & 11.31  & 48.8 & 4.28 & 0.21 & 0.44 & 51.2 & 4.39 & 6.23 & 12.37 \\

\hline \rule{0pt}{2ex}
\multirow[c]{6}{*}{\closlong} & \multirow[c]{2}{*}{\chtwo} & \orient & 4.88 & 0.46 & 5.14  & 58.3 & 4.83 & 0.17 & 0.44 & 41.7 & 4.94 & 1.17 & 8.15 \\
& & \random & 4.38 & 0.37 & 1.13  & 57.9 & 4.34 & 0.22 & 0.44 & 42.1 & 4.43 & 0.60 & 1.63 \\

\cline{2-14}\rule{0pt}{2ex}
& \multirow[c]{2}{*}{\chfour} & \orient & 4.87 & 0.33 & 2.84  & 58.3 & 4.82 & 0.12 & 0.26 & 41.7 & 4.95 & 0.59 & 4.45 \\
& & \random & 4.38 & 0.32 & 0.81  & 57.7 & 4.35 & 0.19 & 0.39 & 42.3 & 4.42 & 0.44 & 1.24 \\

\cline{2-14}\rule{0pt}{2ex}
& \multirow[c]{2}{*}{\chfive} & \orient & 4.87 & 3.48 & 9.52  & 58.3 & 4.82 & 0.12 & 0.24 & 41.7 & 4.95 & 5.63 & 10.32 \\
& & \random & 4.33 & 0.45 & 7.47  & 58.0   & 4.31 & 0.25 & 0.45 & 42.0 & 4.36 & 2.64 & 8.86 \\

\hline
\end{tabular}
\renewcommand{\arraystretch}{1.00}
\end{table*}

\subsection{Datasets}
\label{subsec:datasets}
\noindent 
The combination of parameters we explore in this work
(Table~\ref{tab:exp_st}) corresponds to 96~experiments in \orient,
which resulted in a total of 921,600 ranging samples; in \random, we
performed 36~experiments yielding 126,000 ranging
samples. Table~\ref{tab:datasets_characteristics} summarizes the
salient characteristics of these two datasets\footnote{Our public
  datasets also contain \cnlos data, which we do not exploit in our
  analysis due to ranging samples collected in LOS being unusable
  (\ref{subsec:exp_setup}).  Accounting for those, our datasets
  encompass 144 experiments in \orient and 54 experiments in \random, for a total
  of $\geq1.5$\,million rangings.} by reporting the ratio (\%) of ranging
samples collected in LOS and WNLOS conditions based on our labeling
procedure (\ref{subsec:data_labeling}), the average ground truth
distance between initiator and responder, and the ranging error in
each condition in terms of $90^{th}$ and $99^{th}$ percentiles (\pn,
\pnn).

As expected, WNLOS conditions result in large ranging errors, even
${>}12$\,m, while errors in LOS are typically ${<}$0.5\,m, regardless
of the radio configuration.  \chfive is clearly the worst-performing
channel, with a \pnn error in WNLOS significantly higher than in other
channels, up to $7\times$ when using \closlong in \mobile.  In
contrast, \chfour provides the most accurate ranging estimates when
human occlusion is present.  These observations confirm the importance
of considering different radio settings in our experiments.

\fakepar We also observe that performance varies largely across radio
optimizations, as \closlong---thanks to its longer
\plen---significantly outperforms \clos in WNLOS.  For instance, in
\random, \closlong achieves sub-meter errors, with a \pn of $0.60$\,m
and $0.44$\,m and a \pnn of $1.63$\,m and $1.24$\,m in \chtwo and
\chfour, respectively.  Under the same conditions, \clos exhibits a
\pn of $1.40$\,m and $0.71$\,m and a \pnn of $8.85$\,m and
$6.07$\,m. 
However, \closlong requires $3\times$ to $4\times$ more energy and
runtime than \clos, as discussed in \ref{subsec:energy_runtime}.

%% file: content/labeling.tex
\subsection{Data Labeling}
\label{subsec:data_labeling}

\noindent
One of our goals is to compare the accuracy of classifiers in detecting WNLOS
conditions caused by the human body carrying the responder tag, which
may partially or fully obstruct the link with the \sstwr initiator. To perform our analysis, we
need to associate each ranging sample in our datasets with a
\emph{label}, either LOS or WNLOS, depending on the relative
ground-truth angle~\rangl between initiator and responder, as
accurately determined by \opt~(\ref{subsec:opt}).  

\begin{wrapfigure}{r}{0.36\columnwidth}
\begin{center}
\includegraphics[width=0.385\columnwidth]{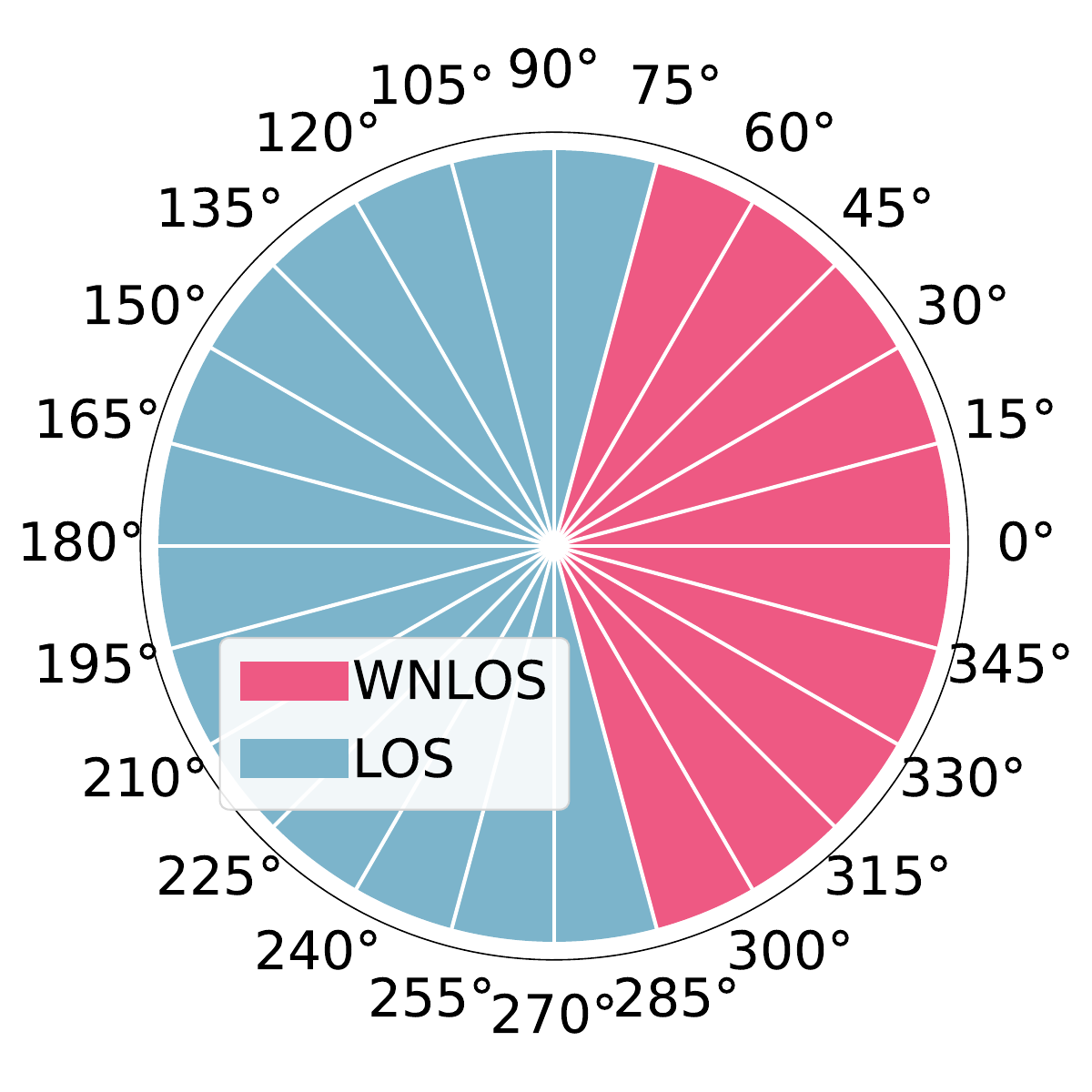}
\end{center} 
\caption{Labeling of LOS \vs WNLOS based on the ground-truth angle
  $\theta$ between the initiator and the user tag; a threshold
  \mbox{$\nlosth\,{=}\,75\degree$} is shown.}
\label{fig:polarplot}
\end{wrapfigure} 
In this respect, we observe that, since the tag is placed on the center
of the chest, this introduces a geometric symmetry: link conditions
for angles \rangl and \mbox{($360^{\circ}{-}\,\rangl$)} are equivalent,
since the body generally obstructs the link in the same manner on
either side (Fig.~\ref{fig:polarplot}).
  
Hereafter, we hence simplify the analysis by folding, \ie merging
the samples for the angles in $(180\degree, 360\degree)$ onto
$[0\degree, 180\degree]$ and report the latter, simplifying our
presentation. Based on this, we consider $\rangl\,{=}\,0\degree$ as
perfect WNLOS and \mbox{$\rangl=180\degree$} as perfect
LOS. Still, a crucial decision is how to determine the angle
threshold \nlosth marking the change between the two conditions.

\begin{figure}[!t]
	\includegraphics[width=0.72\columnwidth]{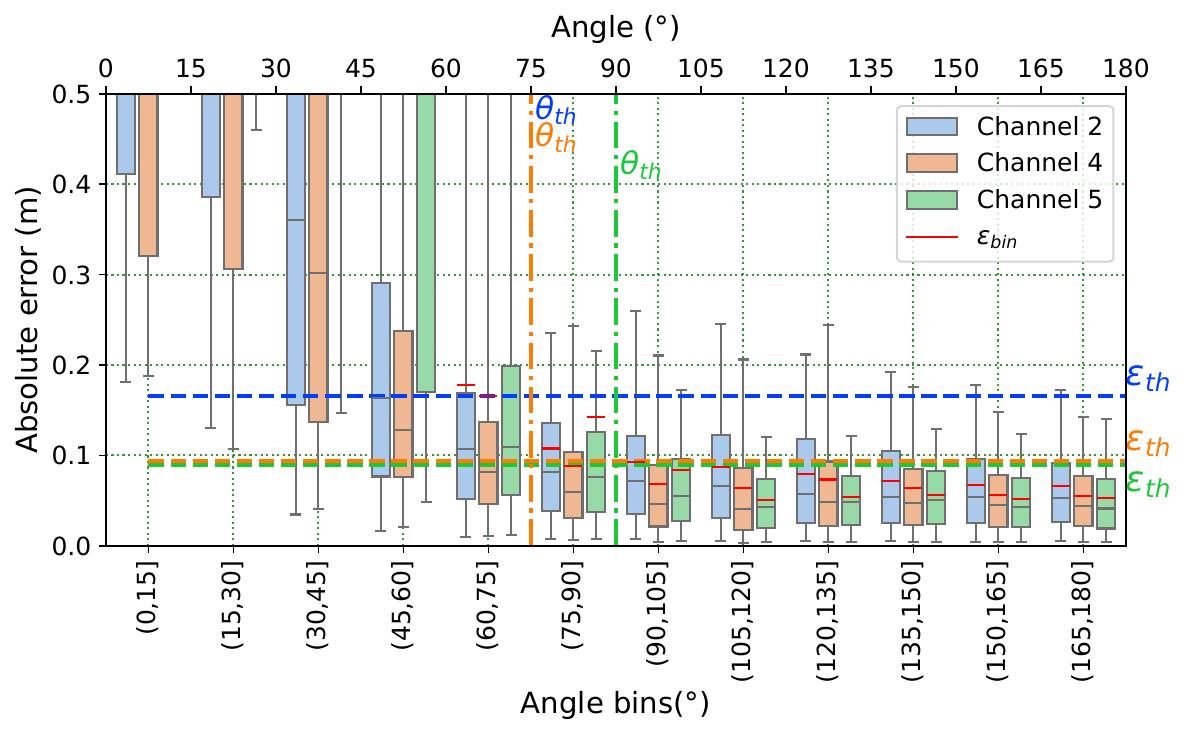}
	\caption{Absolute ranging error per angle bin for \chtwo
          (blue), \chfour (orange), and \chfive (green) when using
          \clos as radio optimization.  The dashed lines show the
          error threshold \errth, allowing visual comparison against
          each bin's mean absolute error \errbin (short red
          line). Some angle bins 
          have $\errbin\,{>}\,0.5$\,m and are cut off in the figure
          for readability.  The dash-dot lines depict the resulting
          angle threshold \nlosth. The color of \errth and \nlosth is
          the same of the corresponding channel.  }
	\label{fig:error_over_angles}
\end{figure}

\boldpar{Does a single angle threshold exist?} 
This problem proved harder than expected. 
In principle, prior work already considered a similar problem. 
However, the reported values for \nlosth are inconsistent, \eg   $15\degree$~\cite{cao20206FitAPartProtocolPhysical}, $67.5\degree$~\cite{Mirama_2023_NLOSDetection} and $90\degree$~\cite{geng2013modeling}. 
Moreover, 
our datasets show quantitatively that the impact of WNLOS on the UWB ranging error for a given \rangl varies depending on the combination of channel and  radio optimization used; Fig.\,\ref{fig:error_over_angles} provides an example showing, for the \clos radio optimization, the impact of the channel alone. 
Therefore, a single, one-size-fits-all value for \nlosth does not exist.

\boldpar{Determining configuration-specific thresholds.} We
select the value of \nlosth for each radio configuration based on the
absolute error distribution of ranging samples in our \pose dataset. \nlosth is set equal to the value \rangl at which the ranging error exhibits a significant increase, according to the following procedure. 
We partition the dataset samples into $15$\degree angle bins and use
the two closest to perfect LOS (\ie from $150\degree$ to $180\degree$) as
the \emph{baseline} for the absolute ranging error, for which we
compute the mean~$\mu$ and standard deviation~$\sigma$. 
We then define an error threshold $\errth\,{=}\,\mu\,{+}\,n\sigma$,
$n\,{\in}\,\mathbb{N}$, 
and iteratively compare it against the mean absolute error \errbin of
each individual $15$\degree angle bin in $(0\degree, 150\degree]$. 
Finally, we set \nlosth to the lower extreme of the \emph{first} bin
for which $\errbin\,{>}\,\errth$, thus attributing the increase in
error to the onset of body obstruction.

We empirically determined that $n$\,=\,2 for \chtwo and $n$\,=\,1 for
\chfour and \chfive yield the sharpest discontinuity in error,
regardless of the radio optimization used; as a sanity check,
  we later confirmed that these are also the values for which pure
  radio link indicators (\lithr, \ref{subsec:approach_li}) yield the
  best classification accuracy. 
With these values of $n$, we
determine the angle threshold as $\nlosth\,{=}\,75\degree$ for all
combinations of channel and optimization except for \chfive with
\clos, for which
$\nlosth\,{=}\,90\degree$. Fig.~\ref{fig:error_over_angles} shows the
resulting $\errth$ and \nlosth for each channel when using \clos.

%% file: content/classifiers.tex
\section{Classifiers}
\label{sec:classifiers}

\noindent
Our goal is to investigate the performance of several 
classifiers executing directly on off-the-shelf UWB tags. In this
section, we provide details for the 21~classifiers we considered,
whose characteristics are concisely summarized in
Table~\ref{tab:classifiers}. We begin with state-of-the-art
LI~(\ref{subsec:approach_li}) and ML
classifiers~(\ref{subsec:approach_ml}), followed by the 16~hybrid ones
we designed~(\ref{subsec:approach_hybrid}) exploiting them \emph{in
  synergy} in an attempt to outperform either approach alone.
Moreover, we illustrate the process adopted for training and testing
the ML models~(\ref{subsec:training_setup}) used in both ML and hybrid
classifiers.

\input{content/classifiers_table}

\subsection{LI-based Classifiers}
\label{subsec:approach_li}

\noindent
We quantitatively evaluate the performance of the three link
indicators \pd, \cl, \clnl, which are readily available from the
DW1000 radio and require little processing
(\ref{subsec:relwork_hnlos_classification}).  However, the
manufacturer only states how to interpret \emph{high-confidence}
values corresponding to LOS or WNLOS; intermediate ones denote
uncertainty, without further information helping to derive a
classification label. Moreover, \pd values are expressed in~dB, while
\cl and \clnl ones as percentages.

We address these issues as follows.  First, we \mbox{normalize} all
indicators to return a value in $[0,1]$, where the two extremes denote
high-confidence LOS and WNLOS, respectively. The values of \cl and
\clnl are already in $[0,1]$, but we use the opposite convention \wrt
Qorvo~\cite{dw1000-manual} to match the definition of ML-based
(\ref{subsec:approach_ml}) and hybrid (\ref{subsec:approach_hybrid})
classifiers. 
For \pd,~we~map values ${<}\,6$\,dB to high-confidence LOS, values
${>}\,10$\,dB to \mbox{high-confidence} WNLOS, and normalize those in
$6\,{\leq}\,\pd\,{\leq}\,10$\,dB to the interval $(0,1)$, in line with
the manufacturer
recommendations~\cite{dw1000:metrics}. 
Second, in both cases we determine the label of low-confidence values
based on a threshold $\rho$; normalized values ${\geq}\,\rho$ are
considered WNLOS, and LOS otherwise.  The simplest threshold selection
is to split in half the interval of low-confidence values, \ie
$\rho\,{=}\,0.5$. We refer to the corresponding classifiers as \pdthr,
\clthr, and \clnlthr, to distinguish them from  hybrid versions
(\ref{subsec:approach_hybrid}) where $\rho$ is determined via ML. In
the following, we refer collectively to them as \lithr.

\subsection{ML-based Classifiers}
\label{subsec:approach_ml}

In line with our goals (\ref{sec:intro}), we restrict our study to
classifiers that can execute entirely \emph{on-device} on commercially
available UWB modules.  As discussed
in~\ref{subsec:relwork_hnlos_classification}, SVM are not generally
applicable. Moreover, many existing approaches based on DNNs target edge
computing platforms and therefore exceed the computational
capabilities of resource-constrained UWB tags, such as the \mdek used
in this work employing an nRF52832 with only 64\,kB of RAM. An
exception is \rem~\cite{angarano2021REMNet}, a lightweight CNN
designed to extract features directly from the CIR, and adapted for
NLOS classification on embedded platforms~\cite{gallacher2023InSight}.
The classifier's low memory footprint of 8.4\,kB of RAM and 8.8\,kB of
flash memory enables deployment on highly resource-constrained UWB
devices.  Accordingly, we adopt \rem as one of the ML-based
classifiers considered in this work, using its publicly available
open-source implementation~\cite{gallacher2023InSight}. 
At the same time, tree-based ensemble methods, such as Random
Forest~\cite{breiman2001RandomForests} and \xgb~\cite{tianqi_xgboost},
provide a more suitable alternative due to their low inference
complexity and compact model representation.  Since \xgb represents a
more recent and highly-optimised gradient-boosting approach with
efficient implementations, 
we focus on this method rather than evaluating both ensemble variants.

Specifically, as in~\cite{gallacher2023InSight}, we~use~an \xgb version based on
the~default settings of the \textit{dmlc/xgboost} library
(v2.1.0), and a \rem model adapted for classification trained in
\textit{TensorFlow} (v2.9.1).  We also employ the original
hyper-parameters\footnote{ \xgb: \emph{subsample}\,=\,0.6,
  \emph{objective}\,=\,binary:logistic, $\eta$\,=\,0.1,
  $\gamma$\,=\,0.5, \emph{n\_estimators}\,=\,10,
  \emph{max\_depth}\,=\,3, \emph{CIR window}$=$20, \emph{CIR samples
    before FP}$=$2.  \rem: \emph{\#filters}\,=\,8,
  \emph{\#parameters}\,=\,659, \emph{\#residual\_blocks}\,=\,1;
  \emph{CIR window}$=$20 (25 in~\cite{gallacher2023InSight}),
  \emph{CIR samples before FP}$=$15 (20 in~\cite{gallacher2023InSight}).},
with the exception of a shorter CIR in \rem due to our datasets
containing fewer samples before the FP.

\subsection{Hybrid Classifiers}
\label{subsec:approach_hybrid}

\noindent
One of our main contributions is the definition and evaluation~of
novel hybrid classifiers exploiting in synergy the complementary
strengths of LI and ML approaches in ways currently unexplored in the
literature. The former run significantly faster and with less
computing overhead. The latter are more general and possibly more
accurate, as they can ``see'' richer patterns that are hard to
explicitly model using predefined indicators. 
We propose four groups of hybrid approaches, each corresponding to a
different combination strategy. 

\boldpar{LI with \xgb-inferred thresholds.} 
One potential limitation of LI classifiers is that the fixed threshold
of $\rho\,{=}\,0.5$, albeit simple, is arbitrary and potentially
sub-optimal. Our first hybrid approach hence exploits ML to
determine the value of $\rho$ that captures the best ``split'' between
the normalized uncertain LI values, used as input
features. We use \xgb due to its ability to
operate on discrete data like LI; in contrast, \rem uses convolutional layers to detect spatial patterns of images or time series data, and  it is thus unsuitable. 
During our training phase (\ref{subsec:training_setup}), 
\xgb is applied separately for each combination of LI, channel, and
radio optimization, yielding different values of $\rho$ that are used
in the LI classifiers instead of the fixed value of $0.5$. We refer
to these classifiers as \pdxgb, \clxgb, \clnlxgb and, collectively, as
\lixgb. Importantly, they do not incur energy or runtime overhead \wrt
\lithr ones; however, they do require an initial training phase based
on a dataset.

\boldpar{\xgb trained with LI.}  Our second hybrid approach is the
dual of the former, as it is based on \xgb trained with the \emph{raw}
values of \cl, \clnl, and \pd as input features.  Its model results in
a decision tree with only $2$--$3$ value comparisons, depending on the
subset of LI considered, yielding faster and more energy-efficient
classification than the default \xgb trained on CIR~samples.  We
explored various alternatives: using all three raw indicators as
features, as well as all possible individual pairs. However, to keep
the discussion of results focused, hereafter we consider only the variant with
\pd and \clnl, denoted as \xgbclnlpd. We select it for two reasons:
\begin{inparaenum}
\item across all 12~combinations of channels, optimizations, and
  scenarios it either yields the best classification accuracy or comes
  very close to it (\perval{-0.20} on average), and
\item it incurs the least energy and runtime overhead since, unlike
  the other three variants, it does not rely on \cl and therefore does
  not require reading the CIR.
\end{inparaenum}

\boldpar{Conditional classifiers.} Our third approach uses a ML
classifier only \emph{after} a LI one and \emph{conditional} to its
output.  If the LI classifier returns a high-confidence output
($0$~or~$1$), this is used directly and classification completes with
minimum runtime and energy costs. Otherwise, a ML classifier is
triggered to classify the uncertain link state. The overall
performance clearly depends on the fraction of high-confidence values;
if high, performance approaches the one of the LI classifier used. We
denote conditional classifiers with the keyword \ifsym\ qualified with
the two classifiers used, as in \ifelsepdxgb. No LI threshold is
indicated, as we only consider high-confidence values (0 or 1) which
are the same for both \lithr and \lixgb, whose threshold comes into
play only for the uncertain values in between.

\boldpar{Consensus classifiers.} Finally, our last approach combines
one LI and one ML~classifier by requiring that a link is classified as
LOS only if \emph{both} classifiers agree on this output value;
otherwise the link is considered in WNLOS. This conservative strategy
is expected to minimize the likelihood that WNLOS links are
misclassified as LOS, causing large ranging errors. 
We denote this group of classifiers
with the symbol \andsym\ qualified with the two classifiers used, as
in \pdandxgb. 
In principle, a dual approach requiring consensus on WNLOS could be
used, offering a way to filter out the corresponding samples. However,
we verified that the performance of this dual approach is
significantly lower, both in terms of accuracy and ranging error of
LOS-classified samples. Therefore, to avoid unnecessarily bloating
the presentation of our results, we do not consider this option
further.

\subsection{Training Procedure for ML Models}
\label{subsec:training_setup}

\noindent
In \orient, each ML model, to be used alone or as part of a hybrid
classifier, is trained and tested separately for each combination of
channel and radio
optimization. 
We perform \textit{\kfold} cross-validation to ensure that models are
as independent as possible from the training and testing data. \kfold
splits data into $k$ subsets, training the model on $k{-}1$ folds and
testing on the remaining one; in our case, $k\,{=}\,5$, \ie each fold
contains $20\%$ of the data. However, in our setup
(\ref{subsec:exp_setup}), each initiator collects multiple samples per
responder position. This may result in similar measurements in both
the training and test data, leading to a positively-biased
classification.  To avoid this, we ensure that samples~from~the same
responder position, initiator-responder pair, human subject, and
environment are \emph{always} kept within the same \kfold split. 
Additionally, for each \kfold split, 20\% of the training data are
reserved as a validation set, used for early stopping of the training,
to reduce overfitting~\cite{Prechelt1998,Goodfellow2016}. Training
terminates when the validation loss does not improve for
10~consecutive training steps; the model parameters corresponding to
the lowest validation loss are retained for evaluation on the test
fold.

In contrast, \mobile is used only as test data for models
trained over the whole \pose dataset, \ie without \kfold but following
the same validation procedure described above, again separately for
each combination of channel and radio optimization.

%% file: content/classifiers_table.tex
\begin{table}[!t]
    \centering
    \small
    \caption{All 21 classifiers considered in this work.}
    \label{tab:classifiers}
    \vspace{-3.25mm}    
    \begingroup
    \begin{tabular}{|c|M{2.9cm}|P{8.90cm}|}
      \hline
      \textbf{Type} & \textbf{Short name} & \multicolumn{1}{c|}{\textbf{Description}}\\
      \hline
      LI   & \pdthr, \clthr, \clnlthr & LI whose value is normalized in $[0,1]$; values\,$\geq$\,$\rho$\,=\,0.5 denote WNLOS, else LOS. \\
      \hline
      ML & \xgb, \rem & ML approaches in~\cite{gallacher2023InSight} and~\cite{angarano2021REMNet}, adapted to run on-device. They take as input a portion of the CIR around the first path.\\
      \hline
      \multirow{8}{10mm}{Hybrid} & \pdxgb, \clxgb, \clnlxgb & Same as \class{\{\cdot\}\qual{0.5}}, except this time the threshold is determined via \xgb. \\
      \cline{2-3}
                    & \xgbclnlpd & \xgb trained by using \pd and \clnl as features instead of the CIR.\\ 
      \cline{2-3}
                    & \mbox{\ifelsepdxgb, \ifelsepdrem,} \mbox{\ifelseclxgb, \ifelseclrem,} 
                      \mbox{\ifelseclnlxgb, \ifelseclnlrem} &                                                      \emph{Conditional classifiers}, composed of a LI- and ML-based classifier. If the former yields a high confidence value based on the manufacturer's definition (\ref{subsec:relwork_hnlos_classification}), this is used; else, the ML classifier is used.\\
      \cline{2-3}
                    & \pdandxgb, \pdandrem, \clandxgb, \clandrem,  \clnlandxgb, \clnlandrem  & \emph{Consensus classifiers}, in which a link is classified as LOS only if \emph{both} classifiers yield this output; otherwise, it is WNLOS. \\
      \hline
    \end{tabular}
    \endgroup
  \end{table}


%% file: content/eval.tex
\section{Experimental Results}
\label{sec:los-hnlos_class}

\noindent
We analyze the performance of the classifiers in
Table~\ref{tab:classifiers} along the dimensions: 
\begin{inparaenum} \item classification accuracy, \item ranging error in LOS-classified samples, \item runtime, and \item energy consumption. \end{inparaenum}

%% file: content/accuracy.tex
\subsection{Classification Accuracy} 
\label{subsec:accuracy_eval}

\boldpar{Metric.} 
Since LOS and WNLOS samples are unevenly distributed in our datasets
(\ref{subsec:datasets}), we consider the \emph{balanced accuracy}, 
which gives equal weight to the two classes and is defined as the average of sensitivity and specificity, \ie 
$\frac{1}{2}\left(\frac{TP}{TP+FN}{+}\frac{TN}{TN+FP}\right)$. 
We consider samples correctly classified as WNLOS as true positives (TP) and samples correctly classified as LOS as true negatives (TN). Consequently, false positives (FP) are LOS samples misclassified as WNLOS; false negatives (FN) are WNLOS samples misclassified as LOS. 

\begin{figure}[!p]
    \centering
    \vspace{-10.00mm}
    \makebox[\columnwidth][c]{%
    \includegraphics[width=1.25\columnwidth]{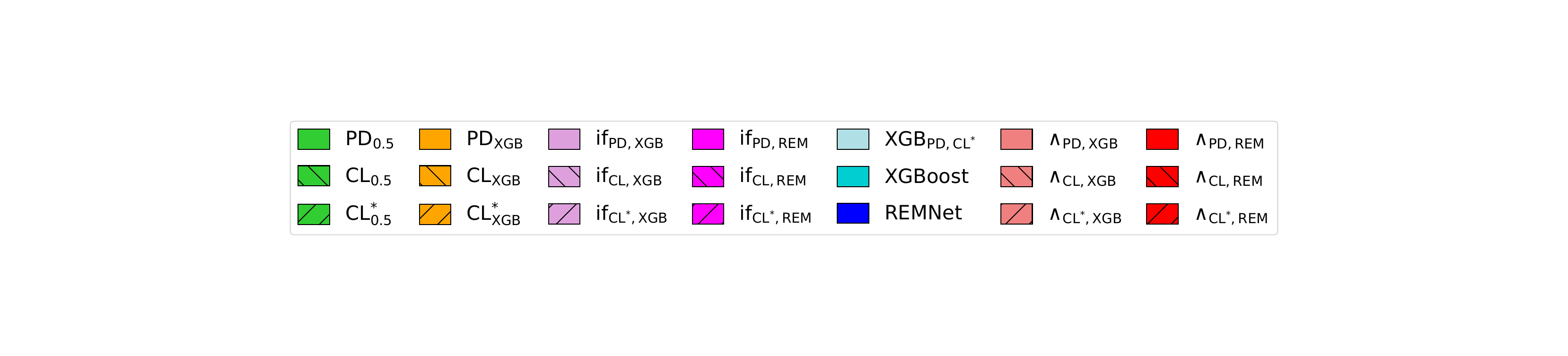}}\\
    \vspace{-14.00mm}%
    \subfloat[Balanced accuracy (\chtwo)]{%
        \includegraphics[width=0.90\columnwidth]{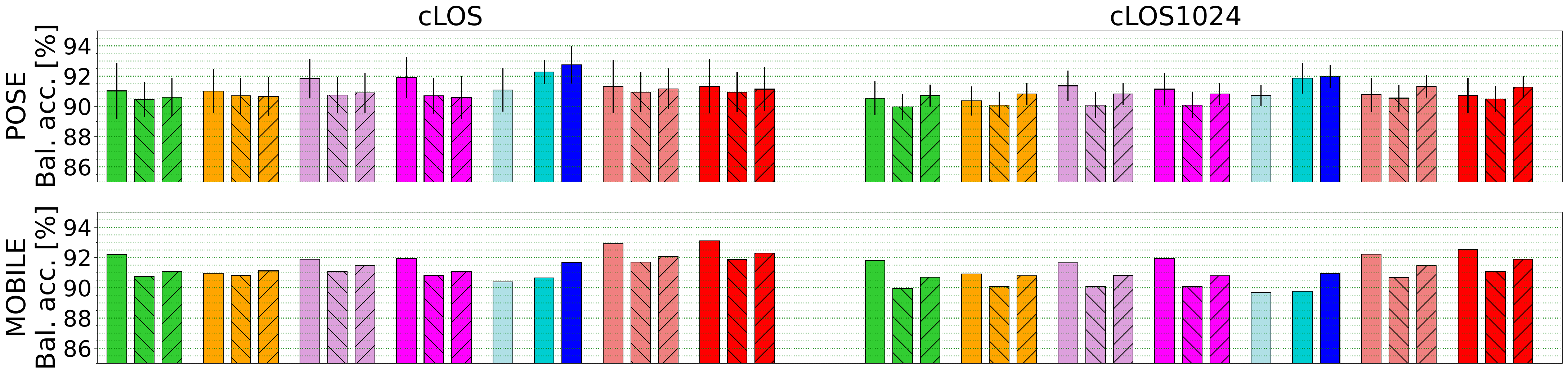}
        \label{fig:five:ch2}}\\
        \vspace{-0.15cm}%
    \subfloat[Balanced accuracy (\chfour)]{%
        \includegraphics[width=0.90\columnwidth]{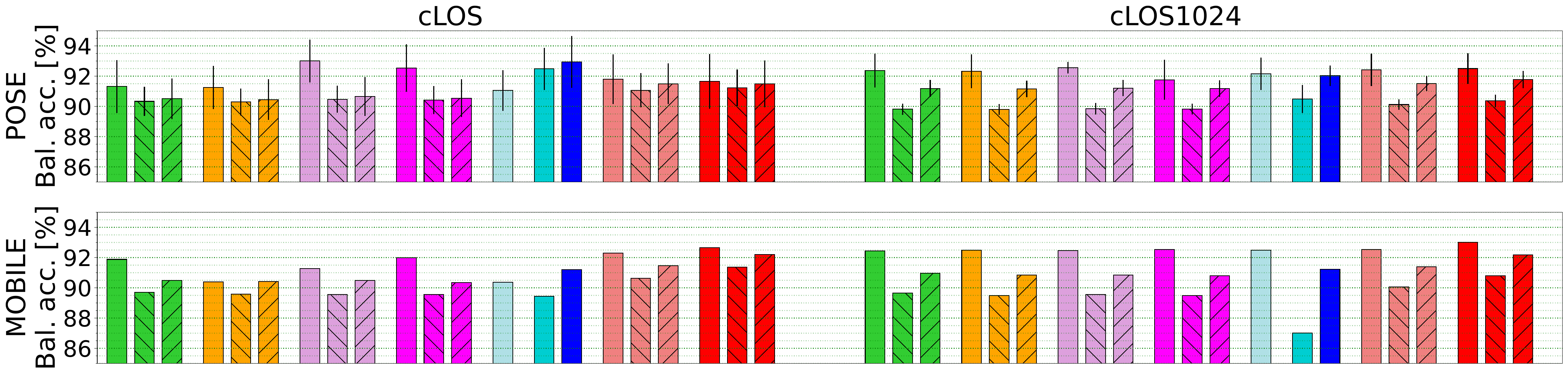}
        \label{fig:five:ch4}}\\
        \vspace{-0.15cm}%
    \subfloat[Balanced accuracy (\chfive)]{%
        \includegraphics[width=0.90\columnwidth]{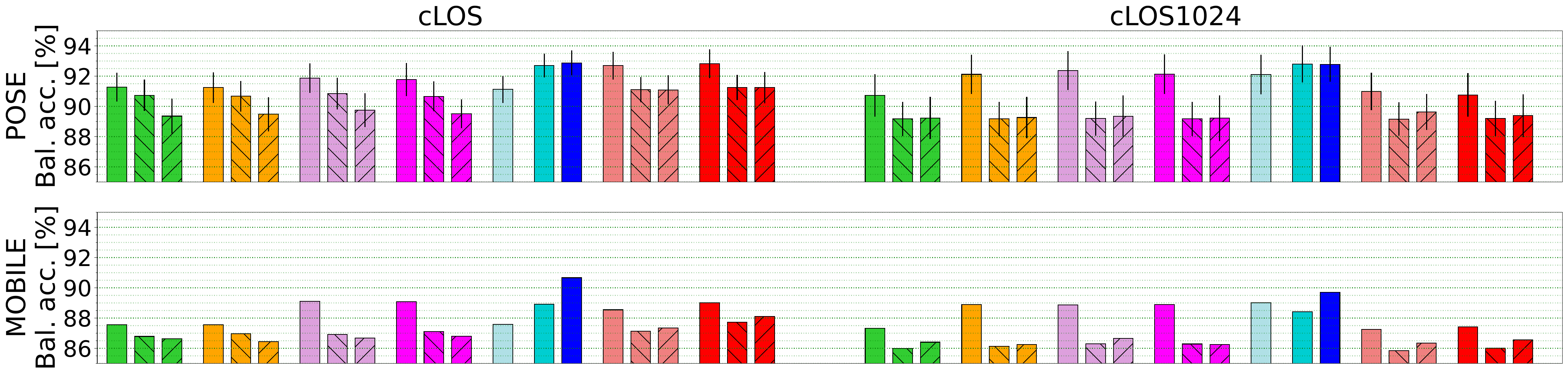}
        \label{fig:five:ch5}}\\
        \vspace{-0.15cm}%
        \hspace{1mm}
    \subfloat[Energy]{%
        \includegraphics[width=0.89\columnwidth]{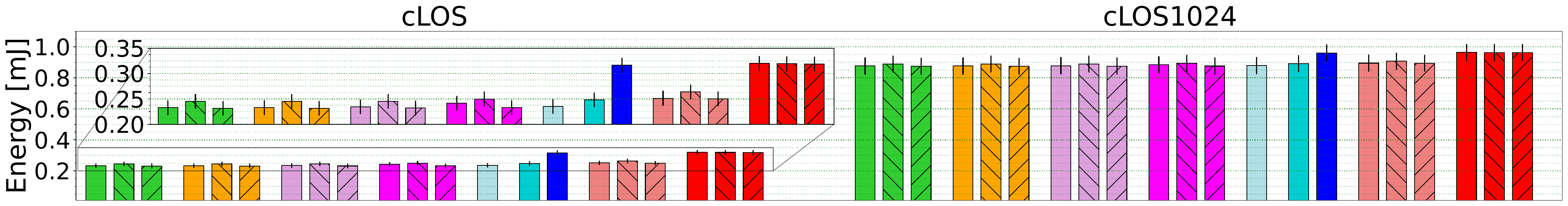}
        \label{fig:five:energy}}\\
        \vspace{-0.15cm}%
        \hspace{2.5mm}
    \subfloat[Runtime]{%
        \includegraphics[width=0.88\columnwidth]{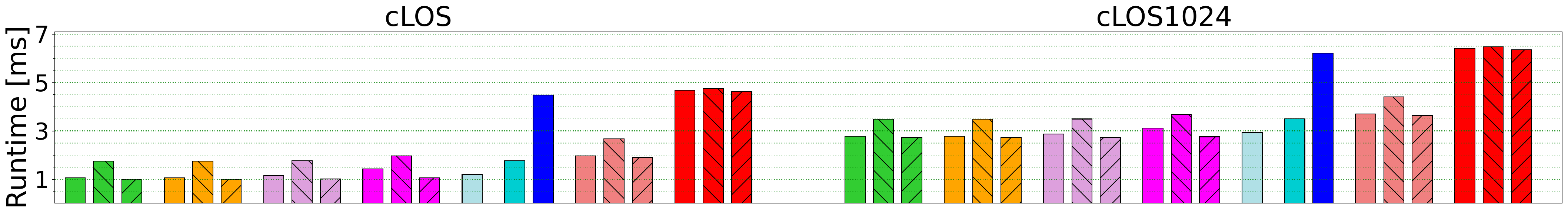}
        \label{fig:five:runtime}} \\
        \vspace{-0.15cm}%
    \vspace{-2.00mm}
    \caption{Balanced classification accuracy per channel in \orient
      and \random experiments, along with overall energy consumption
      and runtime, of all LI, ML-based, and hybrid classifiers considered in
      this study, separated based on the radio optimization used (\clos vs. \closlong).
      Error bars denote the standard deviation of accuracy between
      \mbox{\kfold{s}} in the training dataset (\orient), and the
      min/max energy consumption between channels in the
      energy plot. Energy and runtime figures, discussed
      in~\ref{subsec:energy_runtime}, include the SS-TWR exchange
      and are based on the classification procedure performed with \orient.}
    \Description{}
    \label{fig:five_subfigures}
\end{figure}

\boldpar{General considerations.}
Figs.\,\ref{fig:five:ch2}--\ref{fig:five:ch5} show the balanced
accuracy of all classifiers in both our datasets, across channels and radio optimizations. 
Overall, all classifiers, \ie LI, ML, and their
hybrid variants (\ref{sec:classifiers}), exhibit good classification
accuracy: up to $93.1$\,\%, and always ${>}89.1$\,\% in \orient and
${>}85.8$\,\% in \random regardless of radio settings. 
Interestingly, the lowest classification accuracy is obtained in the
combination\footnote{To improve readability, hereafter we use the
  notation \conf{$A$}{$B$} to denote a combination of two among
  channel, radio optimization, or dataset.}  \conf{\chfive}{\mobile},
with a decrease by up to $4.4$\% for some classifiers. This can be
explained by a combination of two factors. First, as already mentioned,
\chfive is the \textit{channel most affected by WNLOS} in comparison
with \chtwofour (Table\,\ref{tab:datasets_characteristics}). The
reason is that its higher center frequency leads to increased signal attenuation
induced by body occlusion\footnote{Our measurements in \pose
    show a drop of the first-path signal-to-noise ratio (SNR) in
    \chfive of \mbox{-1.45\,dB} compared to \chtwo and
    \mbox{-2.13\,dB} compared to \chfour. These drops accentuate in
    \mobile: \mbox{$-2.70$\,dB} and \mbox{$-3.72$\,dB}, respectively.} 
and more pronounced multipath effects. Second, the different setup of
\mobile, notably based on unconstrained motion of subjects, affects
the ranging error in two opposite ways, across radio channels and
optimizations: the \pnn in LOS is generally significantly higher \wrt\
\pose and, on the contrary, the \pnn in WNLOS is lower. When taken
separately, the impact of these two factors is less marked: in \pose,
whose setup is significantly more controlled and uniform, the overall
difference in classification accuracy between \chtwofour and \chfive
is not dramatic and, dually, neither is between \pose and \mobile in
\chtwofour. However, the \emph{combination} of \emph{both}
factors---more challenging channel, more challenging setup---in
\conf{\chfive}{\mobile} significantly increases the complexity of the
classification task, inducing a more significant drop in its
accuracy.

\boldpar{Comparing classifiers.} Although all classifiers perform
well, they yield different accuracy depending on radio channel,
optimization, and dataset, but also their mode of operation.

\italicpar{Pure LI.} Among \lithr classifiers, based purely on link
indicators, \pdthr outperforms (up to~$+2.8$\,\% accuracy) the other two across
all channels and optimizations except for \conf{\chtwo}{\closlong} where \clnlthr is marginally more accurate ($+0.18$\,\%). \pdthr always yields accuracies
$\geq$\,$90.5$\,\%, except in \conf{\chfive}{\random}, where
nonetheless it remains the best performing LI. 

As for the other two LI
classifiers, \clthr has similar or worse accuracy than its faster and
more energy efficient variant \clnlthr across all channels and
optimizations except for \conf{\chfive}{\clos}, where the difference
is $1.4$\,\%. This generally confirms the
observation that \luep is not the decisive factor for correct WNLOS
classification~\cite{le22hnlos}.

\italicpar{Pure ML.}  
In the \pose dataset, \xgb or \rem are
consistently among the most accurate classifiers across all channels
and optimizations, except for \conf{\chfour}{\closlong}, with \rem
typically outperforming \xgb.  In contrast, in
\conf{\chtwofour}{\mobile} both ML classifiers are among the least
accurate (\eg \xgb is the second best classifier in
\conf{\chtwo}{\pose}, but the second worst in \conf{\chtwo}{\mobile}),
and exhibit a decrease in accuracy up to $3.5$\,\% compared to \pose.
This drop stems from the ML classifiers being trained on \pose, where
user motion is scripted; in contrast, \mobile represents a markedly
different scenario due to the unconstrained motion and different
position of the fixed initiators.  This confirms that ML approaches
have limited robustness to unseen conditions and may benefit from
training on data from the target environment~\cite{stocker23nlos}.
Yet, this is not always the case: \rem is always outperforming \xgb in
\mobile, and represents the best overall choice in
\conf{\chfive}{\mobile}, with its $90.7$\,\% accuracy in \clos and
$89.7$\,\% in \closlong.

\italicpar{Hybrid.} 
Figs.\,\ref{fig:five:ch2}--\ref{fig:five:ch5} also
show whether hybrid approaches improve the accuracy of pure LI and ML
classifiers, although the answer varies depending on the dataset,
channel, and radio configuration.

\lixgb, the most lightweight hybrid technique where ML is used only offline to determine the best threshold $\rho$, generally yields negligible differences in accuracy \wrt \lithr. 
\clxgb and \clnlxgb match the performance of their
fixed-threshold counterparts across all channels and radio
optimizations. 
The same applies to \pdxgb, except for two cases: 
\begin{inparaenum}
\item in \conf{\chtwofour}{\mobile}, where \pdxgb performs up to $1.5$\,\% worse than \pdthr in \clos; and 
\item in \conf{\chfive}{\closlong}, where \pdxgb performs up to $1.6$\,\% better than \pdthr. 
\end{inparaenum}
Our datasets provide no conclusive explanation for these two exceptions. 

Also the dual approach \xgbclnlpd, using \xgb with LI as features, does not yield improvements in accuracy, as it performs almost always worse than the original \xgb. 
In the few cases
when it performs better, \ie with \closlong in \conf{\chfive}{\mobile}
and \chfour in both datasets, its accuracy closely matches the one of \pdxgb,
which however uses \xgb only offline. 

Instead, improvements are enabled by both conditional and consensus
classifiers combining LI and ML.  Interestingly, the
  accuracy of \emph{both} types of classifiers is only marginally
  affected by the choice between \xgb or \rem, even in the cases where
  the difference in accuracy between the two pure ML classifiers is
  relatively high, \eg up to $4.2$\,\% in \conf{\chfour}{\closlong}.
  This is because the ML component is applied only in a few
  low-confidence cases in conditional approaches, and is tightly
  coupled with the outcome of the LI component in consensus
  approaches. 
Another common observation is that both hybrid
approaches nearly always perform best in the combinations involving
\pd, mirroring our observation for LI classifiers.

As for the relative performance of conditional \vs consensus
classifiers, it depends on the settings at hand. 
For example, when considering \orient and \chtwofour, 
we observe that the conditional classifiers based on
\pd yield slightly higher accuracy than their consensus counterpart in \clos, 
matching ML classifiers while executing them only when the LI
component yields low confidence.  Those based on \cl and \clnl,
instead, yield the same or slightly worse accuracy than consensus
ones.  When using \closlong, instead, consensus classifiers yield
similar or slightly better performance than conditional ones. However,
in all these cases the differences are limited, ${<}\,1.2$\,\%.  In
contrast, when considering \mobile and \chtwofour, consensus
classifiers not only are \emph{always} more accurate (up to
\perval{+1.9}) than conditional ones, but the \pd-based variants are
achieving the \emph{best accuracy overall} across the combinations of
channel and radio optimization.  Consensus classifiers are thus
better, in \chtwofour, at tolerating the more challenging \mobile
environment via the more conservative approach required by the
agreement between LI and ML classifiers. 
However, once again the tradeoffs are slightly different in
\chfive. Overall, the superiority of both conditional and consensus
classifiers based on \pd \vs those based on \cl and \clnl is
confirmed. However, unlike in \chtwofour, \pd-based consensus
classifiers are one of the best choices only in \conf{\pose}{\clos},
where they match the performance of pure ML classifiers, unlike
conditional ones. Instead, in the other cases in \chfive, conditional
classifiers perform similarly (\conf{\mobile}{\clos}) or significantly
better (\closlong) than consensus ones. In any case, the only
classifier that consistently yields the best accuracy in \chfive
across radio optimizations and datasets remains \rem, as already
observed.

\begin{figure}[!t]
    \centering
    \includegraphics[width=0.61\columnwidth]{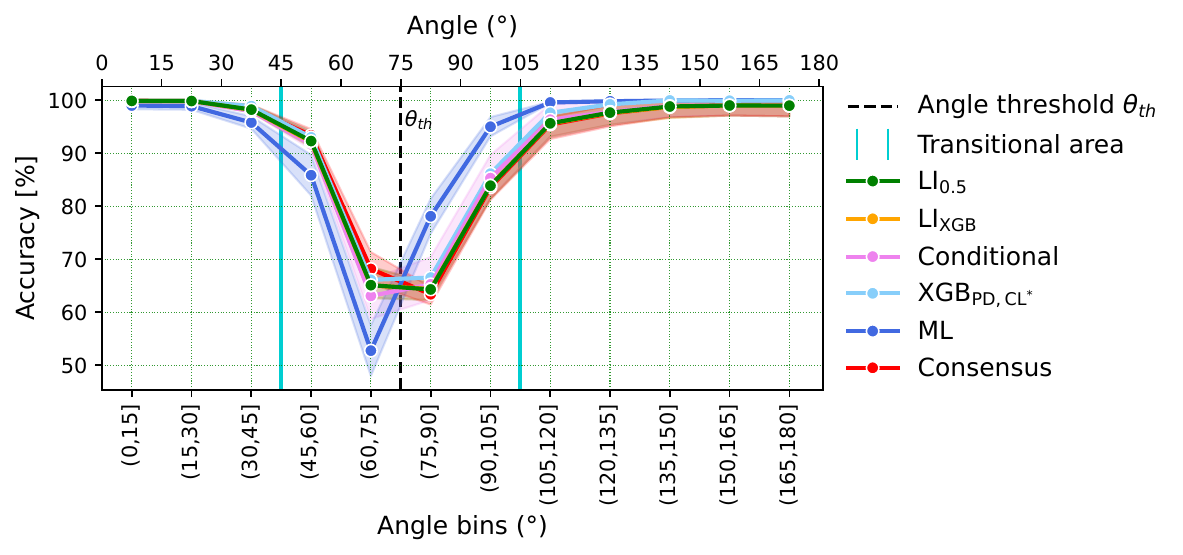}
    \caption{Accuracy per $15\degree$ angle bin for
      \conf{\chfour}{\closlong} in \orient. Solid lines show the mean accuracy, areas show the min/max percentile interval. Angles are folded so that $0\degree/180\degree$ denote WNLOS/LOS (\ref{subsec:data_labeling}). 
    }
    \Description{}
    \label{fig:transitional_area}
\end{figure}

\boldpar{Impact of body occlusion degree.} 
Using the accurate angular information provided by \opt, we analyze how the accuracy of classifiers changes in function of the fraction of the body causing occlusion. 
For illustration, we focus on \conf{\chfour}{\closlong} in \pose and, to facilitate the analysis, we aggregate the results from the 6 groups of classifiers we consider (\ref{sec:classifiers}). 
Fig.\,\ref{fig:transitional_area} shows the results as a function of the angle \rangl between
initiator and responder; other combinations of channel, radio
optimization, and dataset exhibit similar trends.

The chart shows that accuracy degrades as the body
rotation approaches the angle threshold, $\nlosth\,{=}\,75$\degree in this
case, which marks the switch between LOS and WNLOS conditions in our
labeling. As expected, ML approaches are the most affected,
as they are trained on labeled samples, and incur the worst
accuracy of ${\approx}53$\% right before the label changes from WNLOS to LOS. As
for the other classifiers, they all exhibit similar trends, yet
different from ML. 
The drop in accuracy is largely concentrated in a
$\pm$\,30\degree interval around \nlosth that, due to its nature,
we call \emph{\trans} area. In this area, the human body
causes random attenuation and blocking of the UWB signals, leading to
inconsistent signal propagation and reception. Instead, outside of this area, 
the link conditions are more clearly LOS or WNLOS, and the accuracy of
all techniques approaches 100\%.

\begin{figure}[!t]
    \centering
    \includegraphics[width=\columnwidth]{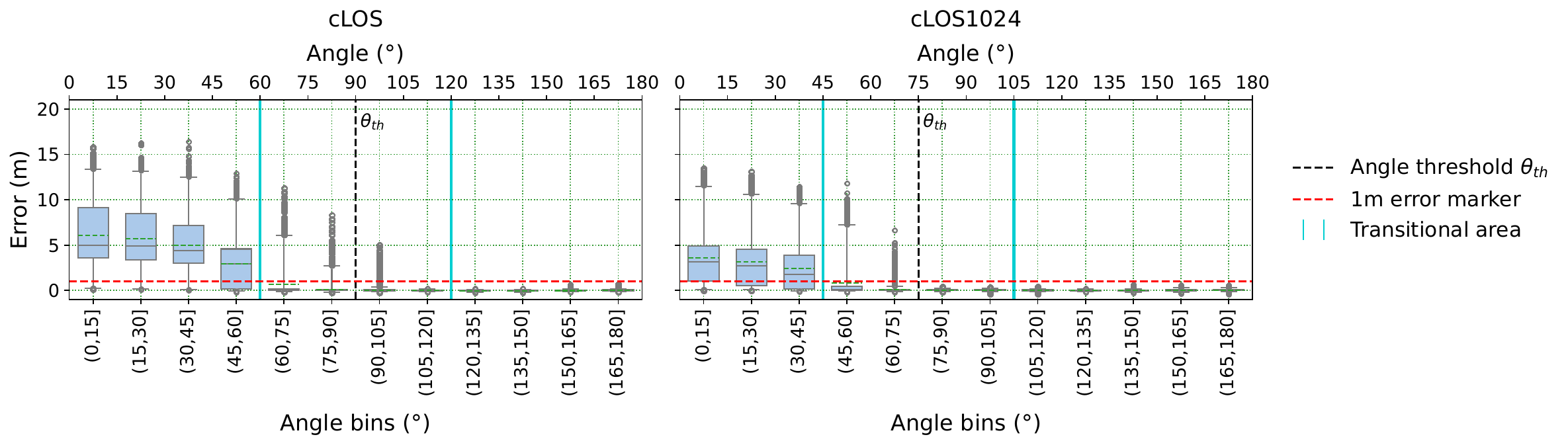}
    \caption{Ranging error per 15\degree angle bin for \conf{\chfive}{\clos} and \conf{\chfive}{\closlong} in \pose. Whiskers are the $1^\mathit{st}$ and $99^\mathit{th}$ percentiles of the error. 
    The transitional area is marked with two parallel lines $\pm30\degree$ around \nlosth.} 
    \label{fig:transitional_area_error}
\end{figure}

\fakepar 
Given that the transitional area exhibits a degradation of
classification accuracy, a natural question is whether a similar
degradation occurs for ranging accuracy.
Fig.\,\ref{fig:transitional_area_error} answers this question by
showing the ranging error for \conf{\chfive}{\pose} for both radio
optimizations, as a function of the angle~\rangl between initiator and
responder. We only show one channel for brevity, and choose \chfive as
it incurs the highest \pnn error
(Table~\ref{tab:datasets_characteristics}). Still, the errors in the
transitional area remain small, on average ${<}0.18$~m, across
channels and optimizations.

These considerations allow us to identify an important, practically
relevant, and desirable property of the classifiers we considered:
\begin{quote}
  \emph{When WNLOS detection matters the most, \ie in prominent
    occlusions causing on average very large ranging errors
    (Figs.~\ref{fig:hnlos_impact},~\ref{fig:transitional_area_error}), all classifiers yield near-perfect
    accuracy (Fig.~\ref{fig:transitional_area}). In the
    transitional area where their accuracy is lowest, the impact of a
    misclassification is reduced since WNLOS-induced average ranging errors
    are small (Fig.~\ref{fig:transitional_area_error}}).
\end{quote}

In the next section, we explore ranging errors in significantly more detail.

%% file: content/error.tex
\newcommand{\ccell}{\cellcolor{green!25}}
\begin{table*}
\centering 
\caption{Analytics of LOS-classified samples per channel,
  optimization, and dataset.\\ \vspace{-3.75mm}} {\footnotesize From
  left to right: \% of LOS-classified errors $<$\,0.5\,m and
  $>$\,1\,m, \pnn error of LOS-classified rangings, and \% of
  LOS-classified rangings. 
  For each channel, we sort classifiers by the smallest \pnn in the \orient dataset for \clos. 
  \gtlabel rows represent the ranging error when LOS \vs
  WNLOS is determined by the ground-truth labeling based on 
  \nlosth (\ref{subsec:data_labeling}).} 
\label{tab:error_analytics}
\vspace{+0.75mm}
\scriptsize
\renewcommand{\arraystretch}{0.60} 
\begin{tabularx}{\textwidth}{|c|c|XXXX|XXXX|XXXX|XXXX|}
\hline \rule{0pt}{2ex}
\multirow[c]{5}{*}{\textbf{Channel}} & \multirow[c]{5}{*}{\textbf{Classifier}} & \multicolumn{8}{c|}{\textbf{\clos}} & \multicolumn{8}{c|}{\textbf{\closlong}} \\
 &  & \multicolumn{4}{c|}{\textbf{\orient}} & \multicolumn{4}{c|}{\textbf{\random}} & \multicolumn{4}{c|}{\textbf{\orient}} & \multicolumn{4}{c|}{\textbf{\random}}\\
 & & \multicolumn{3}{c}{error} & LOS &  \multicolumn{3}{c}{error} & LOS &  \multicolumn{3}{c}{error} & LOS &  \multicolumn{3}{c}{error} & LOS \\
 &  & $<$0.5m & $>$1m & \pnn &  & $<$0.5m & $>$1m  & \pnn &  & $<$0.5m & $>$1m & \pnn & & $<$0.5m & $>$1m  & \pnn &  \\
  &  & (\%) & (\%) & (m) & (\%) & (\%) & (\%) & (m) & (\%) & (\%) & (\%) & (m) & (\%)  & (\%) & (\%) & (m) & (\%) \\
\hline \rule{0pt}{2ex}
\multirow[c]{26}{*}{\chtwo} & \clandrem & 100.0 & 0.0 & 0.29 & 53.9 & 99.4 & 0.0 & 0.47 & 57.9 & 99.9 & 0.0 & 0.38 & 54.5 & 99.6 & 0.0 & 0.44 & 57.7 \\
 & \clandxgb & 100.0 & 0.0 & 0.29 & 54.2 & 99.3 & 0.0 & 0.47 & 58.2 & 99.9 & 0.0 & 0.38 & 54.5 & 99.5 & 0.0 & 0.44 & 58.5 \\
 & \clthr & 99.9 & 0.0 & 0.30 & 55.3 & 99.1 & 0.0 & 0.49 & 59.6 & 99.9 & 0.0 & 0.38 & 55.4 & 99.5 & 0.0 & 0.45 & 59.4 \\
 & \ifelseclxgb & 99.9 & 0.0 & 0.30 & 55.6 & 99.1 & 0.0 & 0.49 & 59.8 & 99.9 & 0.0 & 0.38 & 55.6 & 99.5 & 0.0 & 0.45 & 59.8 \\
 & \ifelseclrem & 99.9 & 0.0 & 0.30 & 55.7 & 99.1 & 0.0 & 0.49 & 60.0 & 99.9 & 0.0 & 0.38 & 55.6 & 99.5 & 0.0 & 0.45 & 59.8 \\
 & \clxgb & 99.9 & 0.0 & 0.30 & 55.7 & 99.1 & 0.0 & 0.49 & 60.0 & 99.9 & 0.0 & 0.38 & 55.6 & 99.5 & 0.0 & 0.45 & 59.8 \\
 & \clnlandxgb & 99.7 & 0.2 & 0.31 & 54.8 & 99.2 & 0.1 & 0.48 & 58.9 & 99.9 & 0.0 & 0.38 & 55.6 & 99.5 & 0.0 & 0.45 & 59.6 \\
 & \clnlandrem & 99.7 & 0.3 & 0.31 & 54.6 & 99.3 & 0.0 & 0.47 & 58.5 & 99.9 & 0.0 & 0.38 & 55.5 & 99.6 & 0.0 & 0.44 & 58.8 \\
 & \pdandxgb & 99.6 & 0.3 & 0.31 & 54.7 & 99.2 & 0.1 & 0.48 & 58.8 & 99.9 & 0.0 & 0.37 & 54.4 & 99.5 & 0.0 & 0.44 & 58.8 \\
 & \pdandrem & 99.6 & 0.3 & 0.31 & 54.6 & 99.4 & 0.0 & 0.47 & 58.5 & 99.9 & 0.0 & 0.37 & 54.3 & 99.6 & 0.0 & 0.44 & 58.0 \\
 & \clnlthr & 99.6 & 0.3 & 0.32 & 56.0 & 99.0 & 0.2 & 0.50 & 60.3 & 99.9 & 0.0 & 0.38 & 56.5 & 99.4 & 0.0 & 0.46 & 60.5 \\
 & \ifelseclnlxgb & 99.6 & 0.4 & 0.32 & 56.2 & 99.0 & 0.2 & 0.50 & 60.4 & 99.8 & 0.0 & 0.38 & 56.7 & 99.4 & 0.0 & 0.46 & 61.0 \\
 & \pdthr & 99.6 & 0.4 & 0.32 & 55.6 & 99.0 & 0.2 & 0.50 & 59.9 & 99.9 & 0.0 & 0.37 & 54.9 & 99.5 & 0.0 & 0.45 & 59.4 \\
 & \ifelseclnlrem & 99.5 & 0.5 & 0.33 & 56.1 & 98.9 & 0.2 & 0.51 & 60.8 & 99.8 & 0.0 & 0.38 & 56.7 & 99.4 & 0.0 & 0.46 & 61.0 \\
 & \clnlxgb & 99.4 & 0.5 & 0.33 & 56.3 & 98.9 & 0.2 & 0.51 & 60.7 & 99.8 & 0.0 & 0.38 & 56.6 & 99.4 & 0.0 & 0.46 & 61.0 \\
 & \rem & 99.6 & 0.4 & 0.33 & 59.0 & 99.3 & 0.0 & 0.47 & 62.5 & 99.7 & 0.0 & 0.41 & 59.7 & 99.5 & 0.0 & 0.45 & 63.1 \\
 & \ccell \gtlabel & \ccell 99.7 & \ccell 0.1 & \ccell 0.34 & \ccell 58.3 & \ccell 99.6 & \ccell 0.0 & \ccell 0.45 & \ccell 57.9 & \ccell 99.4 & \ccell 0.0 & \ccell 0.44 & \ccell 58.3 & \ccell 99.6 & \ccell 0.0 & \ccell 0.44 & \ccell 57.9 \\
 & \ifelsepdrem & 99.5 & 0.5 & 0.34 & 58.2 & 99.1 & 0.2 & 0.49 & 61.8 & 99.8 & 0.0 & 0.38 & 56.9 & 99.5 & 0.0 & 0.45 & 61.0 \\
 & \ifelsepdxgb & 99.4 & 0.5 & 0.34 & 58.4 & 99.1 & 0.2 & 0.49 & 61.5 & 99.8 & 0.0 & 0.38 & 56.5 & 99.5 & 0.0 & 0.45 & 61.0 \\
 & \pdxgb & 99.4 & 0.5 & 0.34 & 58.9 & 98.8 & 0.2 & 0.52 & 62.6 & 99.8 & 0.0 & 0.38 & 57.2 & 99.4 & 0.0 & 0.46 & 62.1 \\
 & \xgbclnlpd & 99.4 & 0.5 & 0.35 & 58.7 & 98.7 & 0.2 & 0.53 & 63.5 & 99.8 & 0.0 & 0.38 & 57.5 & 99.3 & 0.0 & 0.47 & 64.4 \\
 & \xgb & 99.2 & 0.7 & 0.39 & 60.3 & 99.0 & 0.3 & 0.50 & 63.8 & 99.6 & 0.1 & 0.42 & 60.5 & 99.4 & 0.0 & 0.47 & 64.5 \\
\hline \rule{0pt}{2ex}
\multirow[c]{28}{*}{\chfour} & \clandrem & 99.9 & 0.0 & 0.32 & 54.7 & 99.8 & 0.0 & 0.42 & 58.6 & 100.0 & 0.0 & 0.25 & 55.2 & 99.9 & 0.0 & 0.40 & 58.5 \\
 & \clandxgb & 99.9 & 0.0 & 0.32 & 55.0 & 99.8 & 0.0 & 0.43 & 59.7 & 100.0 & 0.0 & 0.25 & 55.4 & 99.9 & 0.0 & 0.40 & 59.4 \\
 & \ifelseclxgb & 99.8 & 0.0 & 0.33 & 56.6 & 99.8 & 0.0 & 0.43 & 61.2 & 100.0 & 0.0 & 0.25 & 55.9 & 99.8 & 0.0 & 0.40 & 60.1 \\
 & \clxgb & 99.8 & 0.0 & 0.34 & 56.5 & 99.8 & 0.0 & 0.43 & 61.2 & 100.0 & 0.0 & 0.25 & 55.8 & 99.8 & 0.0 & 0.40 & 60.2 \\
 & \clthr & 99.8 & 0.0 & 0.34 & 56.5 & 99.8 & 0.0 & 0.43 & 61.0 & 100.0 & 0.0 & 0.25 & 55.8 & 99.8 & 0.0 & 0.40 & 59.9 \\
 & \ifelseclrem & 99.8 & 0.1 & 0.34 & 56.7 & 99.8 & 0.0 & 0.43 & 61.2 & 100.0 & 0.0 & 0.25 & 55.9 & 99.8 & 0.0 & 0.40 & 60.2 \\
 & \pdandxgb & 99.6 & 0.3 & 0.34 & 56.3 & 99.8 & 0.0 & 0.42 & 60.9 & 100.0 & 0.0 & 0.24 & 56.9 & 99.9 & 0.0 & 0.40 & 60.3 \\
 & \ccell \gtlabel & \ccell 99.7 & \ccell 0.0 & \ccell 0.35 & \ccell 58.3 & \ccell 99.8 & \ccell 0.0 & \ccell 0.42 & \ccell 57.4 & \ccell 100.0 & \ccell 0.0 & \ccell 0.26 & \ccell 58.3 & \ccell 99.9 & \ccell 0.0 & \ccell 0.39 & \ccell 57.7 \\
 & \clnlandxgb & 99.5 & 0.3 & 0.36 & 56.1 & 99.8 & 0.0 & 0.43 & 60.8 & 100.0 & 0.0 & 0.25 & 57.4 & 99.8 & 0.0 & 0.40 & 61.3 \\
 & \pdandrem & 99.3 & 0.6 & 0.38 & 56.2 & 99.8 & 0.0 & 0.42 & 60.1 & 100.0 & 0.0 & 0.24 & 56.6 & 99.9 & 0.0 & 0.40 & 59.6 \\
 & \pdthr & 99.3 & 0.6 & 0.39 & 57.5 & 99.7 & 0.1 & 0.43 & 61.8 & 100.0 & 0.0 & 0.24 & 57.1 & 99.9 & 0.0 & 0.40 & 60.6 \\
 & \clnlandrem & 99.3 & 0.6 & 0.40 & 56.0 & 99.8 & 0.0 & 0.42 & 59.7 & 100.0 & 0.0 & 0.25 & 57.2 & 99.9 & 0.0 & 0.40 & 60.4 \\
 & \ifelsepdxgb & 99.3 & 0.6 & 0.40 & 59.0 & 99.7 & 0.1 & 0.43 & 63.1 & 100.0 & 0.0 & 0.25 & 57.9 & 99.9 & 0.0 & 0.40 & 61.2 \\
 & \ifelseclnlxgb & 99.1 & 0.6 & 0.43 & 57.9 & 99.7 & 0.1 & 0.43 & 62.3 & 100.0 & 0.0 & 0.25 & 57.8 & 99.8 & 0.0 & 0.40 & 62.0 \\
 & \pdxgb & 99.1 & 0.7 & 0.43 & 60.9 & 99.7 & 0.1 & 0.43 & 64.2 & 100.0 & 0.0 & 0.24 & 57.2 & 99.9 & 0.0 & 0.40 & 60.4 \\
 & \ifelsepdrem & 99.1 & 0.7 & 0.43 & 59.2 & 99.7 & 0.1 & 0.43 & 62.1 & 100.0 & 0.0 & 0.25 & 59.4 & 99.9 & 0.0 & 0.40 & 60.7 \\
 & \xgbclnlpd & 99.1 & 0.7 & 0.43 & 60.0 & 99.7 & 0.1 & 0.43 & 64.2 & 100.0 & 0.0 & 0.24 & 57.7 & 99.9 & 0.0 & 0.40 & 60.4 \\
 & \rem & 99.1 & 0.7 & 0.44 & 60.1 & 99.8 & 0.0 & 0.42 & 62.8 & 100.0 & 0.0 & 0.26 & 60.4 & 99.8 & 0.0 & 0.40 & 63.3 \\
 & \clnlxgb & 99.1 & 0.7 & 0.45 & 57.9 & 99.7 & 0.1 & 0.43 & 62.3 & 100.0 & 0.0 & 0.25 & 57.8 & 99.8 & 0.0 & 0.40 & 62.0 \\
 & \xgb & 99.1 & 0.5 & 0.45 & 61.2 & 99.7 & 0.1 & 0.43 & 65.1 & 100.0 & 0.0 & 0.29 & 63.0 & 99.7 & 0.0 & 0.41 & 67.6 \\
 & \clnlthr & 99.1 & 0.7 & 0.45 & 57.9 & 99.7 & 0.1 & 0.43 & 62.2 & 100.0 & 0.0 & 0.25 & 57.8 & 99.8 & 0.0 & 0.40 & 61.9 \\
 & \ifelseclnlrem & 99.1 & 0.7 & 0.46 & 57.9 & 99.7 & 0.1 & 0.43 & 62.4 & 100.0 & 0.0 & 0.25 & 57.9 & 99.8 & 0.0 & 0.40 & 62.1 \\
\hline \rule{0pt}{2ex}
\multirow[c]{28}{*}{\chfive} & \clandxgb & 99.9 & 0.1 & 0.18 & 46.8 & 99.8 & 0.0 & 0.42 & 46.5 & 100.0 & 0.0 & 0.20 & 49.3 & 99.8 & 0.0 & 0.44 & 48.6 \\
 & \clandrem & 99.9 & 0.1 & 0.18 & 46.9 & 99.8 & 0.0 & 0.42 & 46.8 & 100.0 & 0.0 & 0.21 & 49.3 & 99.7 & 0.0 & 0.44 & 48.6 \\
 & \ccell \gtlabel & \ccell 99.9 & \ccell 0.1 & \ccell 0.19 & \ccell 50.0 & \ccell 99.6 & \ccell 0.0 & \ccell 0.44 & \ccell 48.8 & \ccell 100.0 & \ccell 0.0 & \ccell 0.24 & \ccell 58.3 & \ccell 99.6 & \ccell 0.0 & \ccell 0.45 & \ccell 58.0 \\
 & \ifelseclxgb & 99.7 & 0.3 & 0.19 & 48.9 & 99.7 & 0.0 & 0.43 & 49.2 & 99.9 & 0.1 & 0.21 & 49.5 & 99.7 & 0.0 & 0.44 & 49.3 \\
 & \clthr & 99.7 & 0.3 & 0.19 & 48.9 & 99.7 & 0.1 & 0.43 & 49.1 & 99.9 & 0.0 & 0.21 & 49.4 & 99.7 & 0.0 & 0.44 & 49.0 \\
 & \clxgb & 99.6 & 0.4 & 0.19 & 48.9 & 99.7 & 0.1 & 0.43 & 49.3 & 99.9 & 0.1 & 0.21 & 49.5 & 99.7 & 0.0 & 0.44 & 49.1 \\
 & \ifelseclrem & 99.4 & 0.6 & 0.20 & 49.1 & 99.7 & 0.1 & 0.43 & 49.4 & 99.9 & 0.1 & 0.21 & 49.5 & 99.7 & 0.0 & 0.45 & 49.4 \\
 & \clnlandrem & 99.0 & 1.0 & 0.26 & 47.8 & 99.7 & 0.1 & 0.43 & 47.3 & 99.1 & 0.9 & 0.33 & 50.6 & 99.7 & 0.0 & 0.45 & 49.4 \\
 & \pdandrem & 99.0 & 1.0 & 0.26 & 49.7 & 99.7 & 0.1 & 0.43 & 48.3 & 99.2 & 0.8 & 0.31 & 52.1 & 99.7 & 0.0 & 0.44 & 49.9 \\
 & \rem & 99.1 & 0.9 & 0.26 & 51.0 & 99.7 & 0.1 & 0.43 & 52.1 & 99.0 & 1.0 & 0.45 & 59.9 & 99.6 & 0.0 & 0.45 & 62.3 \\
 & \clnlandxgb & 99.0 & 1.0 & 0.29 & 47.6 & 99.4 & 0.4 & 0.45 & 47.1 & 99.6 & 0.3 & 0.22 & 50.3 & 99.7 & 0.1 & 0.44 & 49.4 \\
 & \pdandxgb & 99.0 & 1.0 & 2.60 & 49.6 & 99.4 & 0.4 & 0.45 & 48.4 & 99.6 & 0.4 & 0.22 & 51.8 & 99.7 & 0.1 & 0.44 & 49.9 \\
 & \xgb & 98.8 & 1.2 & 2.76 & 51.6 & 99.3 & 0.5 & 0.45 & 51.5 & 99.1 & 0.9 & 0.35 & 60.3 & 99.5 & 0.3 & 0.46 & 61.8 \\
 & \ifelsepdxgb & 97.4 & 2.6 & 4.94 & 52.8 & 98.8 & 0.9 & 0.58 & 52.5 & 99.1 & 0.9 & 0.42 & 55.5 & 99.5 & 0.2 & 0.45 & 55.2 \\
 & \ifelsepdrem & 96.9 & 3.1 & 5.17 & 52.6 & 98.9 & 0.8 & 0.54 & 51.8 & 99.0 & 1.0 & 0.44 & 56.0 & 99.5 & 0.2 & 0.46 & 55.3 \\
 & \ifelseclnlxgb & 96.8 & 3.2 & 6.04 & 50.7 & 98.7 & 1.0 & 3.29 & 49.9 & 99.1 & 0.9 & 0.34 & 50.7 & 99.5 & 0.2 & 0.45 & 50.1 \\
 & \ifelseclnlrem & 96.7 & 3.3 & 6.08 & 50.4 & 98.7 & 1.0 & 3.22 & 50.0 & 98.9 & 1.1 & 2.68 & 50.8 & 99.5 & 0.2 & 0.45 & 49.6 \\
 & \pdxgb & 96.8 & 3.2 & 6.20 & 52.4 & 98.6 & 1.1 & 3.71 & 51.1 & 99.0 & 1.0 & 0.44 & 55.9 & 99.5 & 0.2 & 0.46 & 55.3 \\
 & \clnlxgb & 96.6 & 3.4 & 6.29 & 50.7 & 98.6 & 1.1 & 3.58 & 49.8 & 99.0 & 1.0 & 0.44 & 50.7 & 99.5 & 0.2 & 0.45 & 49.6 \\
 & \xgbclnlpd & 96.7 & 3.3 & 6.36 & 52.1 & 98.6 & 1.1 & 3.70 & 51.0 & 98.9 & 1.1 & 2.60 & 56.1 & 99.5 & 0.2 & 0.46 & 55.6 \\
 & \pdthr & 96.7 & 3.3 & 6.45 & 52.9 & 98.6 & 1.1 & 3.71 & 51.1 & 99.1 & 0.9 & 0.34 & 52.2 & 99.5 & 0.2 & 0.45 & 50.2 \\
 & \clnlthr & 96.2 & 3.8 & 6.51 & 51.2 & 98.5 & 1.2 & 3.84 & 50.2 & 99.0 & 1.0 & 2.29 & 50.8 & 99.5 & 0.2 & 0.46 & 49.8 \\
\cline{1-18}
\end{tabularx}
\renewcommand{\arraystretch}{1.00} 
\vspace{-3.50mm}
\end{table*}

\subsection{Ranging Error of LOS-classified Samples}
\label{subsec:LOS_error}

\noindent Many applications rely exclusively on ranging samples
classified as LOS, with the intent to achieve accurate distance
estimates by avoiding the large errors associated with WNLOS
conditions~(\ref{sec:intro}). 
However, some WNLOS samples may be misclassified as LOS, yielding a high error. 
We hence identify which classifiers yield the lowest ranging
error in LOS-classified samples. 
Table~\ref{tab:error_analytics} summarizes our findings
across all classifiers, radio configurations, and datasets.

\boldpar{Metrics.} Our primary metric for ranging errors is the
$99^\mathit{th}$ percentile (\pnn), as this is a direct indicator of
how long is the tail of the error distribution. Given its importance,
we order the 21~classifiers in Table~\ref{tab:error_analytics} based
on the \pnn of errors in \conf{\clos}{\pose} to provide structure and
improve readability across the many combinations explored. Moreover,
to offer additional details about the error distribution, we also
report the percentage of low and high errors, respectively defined as
$<$\,0.5\,m and $>$\,1\,m.
Another important metric, also reported in
Table~\ref{tab:error_analytics}, is the fraction of LOS-classified
samples over the total, highlighting whether low errors come at the
cost of many samples being discarded. To provide a baseline to interpret
this metric as well as ranging errors, Table~\ref{tab:error_analytics}
includes also the corresponding values obtained from our ground-truth labeling (\ref{subsec:data_labeling}), in the \gtlabel rows, which are the same reported in the column ``LOS only'' of Table~\ref{tab:datasets_characteristics}.

\boldpar{Ranging error.} In~\ref{subsec:accuracy_eval}, we observed
that all \pd-based classifiers, pure and hybrid, generally show the
highest balanced accuracy.  However, this does not reflect in the
lowest ranging error for LOS-classified samples, which is also
strongly affected by the channel and radio optimization used.

In \conf{\chtwo}{\clos}, the spread of \pnn values among classifiers
is relatively small. Remarkably, \emph{all} \cl-based classifiers are
the best performers, with \pnn as low as \cmval{29} in \pose, even
lower than \gtlabel; moreover, they are the \emph{only} classifiers
for which all LOS samples are ${<}\,1$\,m. However, the \pnn of the remaining classifiers is within \cmval{8} of each other, with the exception of pure ML ones whose worst \pnn (\xgb) is nonetheless only \cmval{39}. In \mobile, the \pnn is slightly
higher, but differences are even smaller, with 17~out of~21 classifiers within \cmval{3} of each other. 
Switching to the \closlong optimization brings several benefits:
except for \xgb in \pose, all classifiers yield \emph{all} LOS samples
with sub-meter error, and the spread of \pnn values is reduced.

In \conf{\chfour}{\clos}, the accuracy is similar to
\conf{\chtwo}{\clos}; the \ifelseclnlrem classifier yields the largest
\pnnval{46} in \pose, and \cl-based classifiers, this time along with
\pdandxgb, are the best ones, within \cmval{2} of each other. As for
the other combinations of radio optimization and dataset, the error spread is
always within \cmval{5} and nearly absent in \mobile, coherently with
Table~\ref{tab:datasets_characteristics}. 
Overall, ranging accuracy is often slightly better than in \chtwo,
generally closer to \gtlabel, and even higher with \closlong, due to
the combined effect of larger bandwidth and longer \plen
(\ref{subsec:parameter_space}).  In particular, in
\conf{\closlong}{\pose}, the \pnn for all classifiers is within
\cmval{24\text{--}29}; moreover, \emph{all} classifiers yield samples
whose error is ${<}$\,0.5\,m.

In contrast, in \chfive many of the classifiers incur significantly
worse accuracy, a consequence of higher attenuation and
  stronger multipath we already commented upon. In \clos, \pdthr and
\clnlthr exhibit the largest \pnn and fraction of samples with high
error, which reaches respectively $6.51$\,m and $3.8$\,\% for \clnlthr
in \pose, whereas in the other two channels these LI classifiers show
intermediate accuracy.  On the other hand, \cl-based classifiers yield
again the best accuracy. 
In \pose, \emph{all} these classifiers, plus \clnlandrem, \pdandrem,
\rem, and \clnlandxgb, are the \emph{only} ones yielding sub-meter
\pnn; in particular, both consensus classifiers involving \cl achieve
\pnnval{18}, the lowest across all combinations in
Table~\ref{tab:error_analytics}. In \mobile, the minimum \pnnval{42}
is higher than in \pose, but the maximum $\pnn\,{=}\,3.84$\,m is
nearly half; moreover, the gap of \cl-based classifiers with the
others is smaller, with half of the classifiers yielding a \pnn of
$42$--\cmval{45}.

On the other hand, the worse ranging accuracy in \chfive is addressed by \closlong, whose beneficial impact is significantly
higher than in the other channels. In \pose, only \ifelseclnlrem, \xgbclnlpd,
and again \clnlthr have $\pnn\,{\geq}\,2.29$\,m; \emph{all} the others
have $\pnn\,{\leq}\,0.45$\,m, with $\pnn\,{\leq}\,0.21$\,m for all
\cl-based classifiers. Finally, in \conf{\closlong}{\mobile}, \pnn
values are all within \cmval{2}, with values in line with the same
configuration in the other two channels; however, \cl-based
classifiers, as well as \clnlandrem, \pdandrem, and \rem are the only ones without errors above one meter.

\boldpar{Ranging error \vs number of LOS samples.} Many applications use only LOS samples,
expected to yield high ranging accuracy; WNLOS ones are instead
discarded and possibly trigger a new ranging exchange with a
different, more favorable radio setting. 
It is hence important to assess not only the ranging
error, but also the \emph{fraction} of LOS-classified samples over the total. 
To analyze this aspect, we compare the fraction of LOS-classified
samples of each classifier against the one from our ground-truth
\gtlabel (\ref{subsec:data_labeling}) and take the corresponding
ranging error as a baseline.

First and foremost, we observe that \emph{all} \cl-based classifiers,
and often also consensus ones, are the most conservative across
\emph{all} combinations of channel, optimization, and dataset.  This
is coherent with the fact that \cl-based classifiers generally yield
the lowest errors, and that consensus ones, by construction, decide
LOS only if their LI and ML constituents agree. These types of
classifiers often yield even \emph{lower} ranging errors than \gtlabel
because, 
in comparison, they also identify as LOS \emph{fewer} and
\emph{better} samples. The question, however, is whether the loss in
potential LOS samples is worth the higher ranging accuracy.

The answer to this question greatly depends on the combination at
hand. As discussed earlier, the ranging error of LOS-classified
samples has significant variations across classifiers only in some
combinations: all \chfive ones except for \conf{\closlong}{\mobile}
and, to a lesser extent, \conf{\pose}{\clos} in \chtwo and \chfour. In
these, the application developer faces the trade-off between maximizing
ranging accuracy or the number of LOS samples obtained. The tension between
the two is particularly evident in the three \chfive combinations
above; for instance, in \conf{\pose}{\closlong} \xgb incurs a ranging error
up to \cmval{15} higher than the best classifiers, but retains up
to $11$\,\% more LOS samples. Similar considerations hold for
several classifiers, in all five combinations mentioned above.

Instead, in all the other combinations, the developer is free to
choose the classifier that best suits application requirements. Using
\cl-based and consensus ones is no longer beneficial, as this brings less
LOS-classified samples without any relevant benefit in reduced ranging
error. In contrast, classifiers that \emph{over}estimate LOS samples
may be more useful, as they provide more data to the application
without incurring any error penalty.  

%% file: content/runtime_energy.tex
\vspace{-0.50mm}
\subsection{Runtime and Energy Consumption}
\label{subsec:energy_runtime}
\vspace{-0.25mm}
\noindent
In applications operating on-device, the choice of classifier
depends not only on its accuracy and on the quality and number of 
LOS-classified samples but also, crucially, on the
associated runtime~and energy overheads, both typically neglected in
the literature~(\ref{subsec:relwork_hnlos_classification}). 
Runtime has a direct impact on the update rate sustainable by the device; energy consumption affects the tag's battery lifetime. 

\boldpar{Metrics.} 
Runtime and energy consumption are determined not
only by the operation of the classifier, but also by the transmission
and reception of the packets required by the \sstwr exchange. The two
aspects are interrelated because, as we have discussed in earlier
sections, the performance of classifiers strongly depends on the
channel and radio optimization used, and the latter strongly affects
runtime and energy consumption (\ref{subsec:parameter_space}). To
analyze in-depth the trade-offs at stake, we analyze first the overhead
of classifiers \emph{alone} and then the \emph{overall} one including \sstwr. 

We characterize the corresponding metrics in different ways. For the
\sstwr exchange, we estimate them via the overhead calculator provided
by Qorvo\footnote{Unfortunately, while preparing and finalizing this manuscript, the calculator spreadsheet became no longer available online. However, it can be obtained from the authors upon request.} 
for the DW1000 radio we use in this work, for each combination of
channel and radio optimization. Instead, we directly measure the
overhead of classifiers alone via the Nordic Power Profiler
Kit~II~\cite{nordic-ppk2}. To ensure accurate results, we timestamp
the toggling of~GPIO pins when collecting CIR data, pre-processing it,
and performing the actual classification. 
  As our experiments run on a MCU entirely dedicated to them,
  execution times are highly deterministic and run-to-run variations
  are negligible.  We therefore measure the average current
  consumption over the corresponding time intervals identified through
  GPIO timestamps.   Since the \mdek board prevents direct
measurement of current consumption of the nRF52832 SoC, we use for our
measurements a development
kit\footnote{\url{https://nordicsemi.com/Products/Development-hardware/nRF52840-DK}}
with the same CPU and hosting a UWB shield with the DW1000.

\boldpar{Overhead of classifiers alone.} Among the
21~classifiers~\mbox{under} study, it is sufficient to measure the
overhead only for pure~LI (\ref{subsec:approach_li}) and
pure~ML~ones~(\ref{subsec:approach_ml}), as they determine~also~the
overhead of hybrid variants.  \lixgb classifiers have the same
overhead of \lithr ones, as the ML component is executed only offline
to determine the appropriate threshold.  The overhead of \xgbclnlpd is
the sum of \pdthr and \clnlthr; the contribution of the \xgb model is
negligible, as it embeds only $2$--$3$ tree levels
(\ref{subsec:approach_hybrid}).  Similarly, the overhead of consensus
classifiers is the sum of their LI and ML components.  Finally, the
overhead of conditional approaches cannot be predetermined, as it
depends on the specific UWB signal; if the LI component is highly
confident ($0$ or $1$) only its overhead applies, otherwise the one of
the ML component must be added. For context, this latter case happens
only for \perval{4\text{--}5} of the samples in \random. 

\begin{table}[!t]
    \centering
    \caption{Measurements of energy consumption and runtime for
      \emph{pure} LI and ML classifiers.}
    \label{tab:energy_runtime}
    \vspace{-3.00mm}
   \setlength\tabcolsep{4pt}
    \begin{tabular}{|c|c|c|c||c|c|}
      \cline{2-6}
      \multicolumn{1}{c|}{} & \clnlthr & \pdthr & \clthr  & \xgb & \rem \\
      \hline\rule{0pt}{2ex}
      \textbf{Energy ($\mu$J)} &  2.29 & 3.27 &15.68 & 18.60 & 87.36 \\
      \textbf{Runtime ($\mu$s)} & 140 & 200 & 900 & 917 & 3629\\\hline
    \end{tabular}
\end{table}

Table~\ref{tab:energy_runtime} shows the overhead of pure LI and ML classifiers, from which the others can be derived. 
Overall, \textit{\clnl incurs the least overhead}, closely followed by \pdthr, which requires $1.4\times$ more energy and
runtime. 
Instead, \cl incurs $6.8\times$ energy and $8.2\times$
runtime than \clnl due to the need of acquiring a CIR portion. 
This brings the overhead of \cl close to \xgb, which is only \perval{2} slower and \perval{19} more energy-hungry. 
At the other extreme, \rem is by far the classifier with the highest overhead at $4.7\times$ energy and $4\times$ runtime than \xgb, and $38\times$ and $26\times$ than \clnl, respectively.

\boldpar{Total classification overhead.}
Fig.~\ref{fig:five:energy}--\ref{fig:five:runtime} show the energy
and runtime of each classifier including the \sstwr exchange, 
based on the \orient dataset and averaged on all channels. 

In terms of \textit{energy}, the difference among classifiers is
dwarfed by the difference in radio optimization, as \closlong incurs
$3.8\times$ higher consumption than \clos \footnote{In \clos,
  \chtwofour consume $224.5\,\mu J$ \vs $239.1\,\mu J$ of \chfive. In
  \closlong, \chtwofour consume $854\,\mu J$ \vs $909.6\,\mu J$ of
  \chfive.}. 
However, within the less expensive \clos optimization, the difference
between \rem and the other approaches is still significant, \ie
\perval{+37.5} higher than \clnlthr.  The variations within each
classifier, shown by the error bars denoting the min/max values in
the zoomed-in box, are to be ascribed to the different channels;
\chfive shows the highest energy consumption, roughly $6.5\%$ higher than \chtwofour. 

As for \textit{runtime}, the difference among classifiers is more marked. The
choice of radio optimization is still key, given that a \sstwr
exchange lasts $2588~\mu$s in \closlong \vs $856~\mu$s in \clos, \ie
$\approx$\,$3\times$ longer.  However, within the same optimization,
the choice of classifier may have an even higher impact. \rem alone
and the consensus classifiers exploiting it are $\approx$\,$4.5\times$
slower than \clnlthr and \pdthr in \clos; unlike with energy, the
runtime of \rem and its variants in \clos is actually \emph{higher}
than all the other classifiers in \closlong. Differences are more
marked also among classifiers belonging to the same group. For
instance, in \clos, among pure LI classifiers, \clthr ($1756~\mu$s) is $\approx$\,$1.8\times$ slower than \clnlthr ($996~\mu$s) and $\approx$\,$1.7\times$ slower than \pdthr ($1056~\mu$s) while, within pure ML classifiers, \rem ($4485~\mu$s)
is $\approx$\,$2.5\times$ slower than \xgb ($1773.4\,\mu$s).  Similar trends exist in \closlong,
although with slightly smaller differences.

%% file: content/putting_together.tex
\subsection{Summary of Findings}
\label{subsec:rec_tool}

\noindent
The results presented
in~\ref{subsec:accuracy_eval}--\ref{subsec:energy_runtime} contain a
large number of findings, concisely summarized
in~Table~\ref{tab:energy_runtime_restructured} 
as answers to the questions we posed in~\ref{sec:rq}. 
These findings clearly show that
no classifier or approach has performance superior to the others in
all the metrics we considered and across all radio configurations;
worse, a classifier performing very well along one dimension may be
impractical in others. Our detailed and quantitative analysis
contributes to shed light on these complex interrelationships and
trade-offs. However, navigating through all the findings and trade-offs
we reported can be overly complex when the goal is to identify the
best compromise satisfying a given set of application requirements.
We address this problem next, with our final contribution.

\input{content/final_table_answers}

%% file: content/final_table_answers.tex
\newcommand{\qcell}{\cellcolor{CornflowerBlue!45}}

\newcommand{\finding}[2]{\qcell #1\\\hline\begin{minipage}[t]{\linewidth}\vspace{0.25pt}\begin{compactitem}[$\circ$]#2\end{compactitem}\vspace{5pt}\end{minipage}\\\hline}

\begin{table}[p]
    \centering
    \footnotesize
    \renewcommand{\arraystretch}{1.00} 
    \caption{Answers to the research questions in~\ref{sec:rq}.}
    \label{tab:energy_runtime_restructured}
    \vspace{-3.25mm}
    \begin{tabularx}{\columnwidth}{|X|}
      \hline
      \finding{\qone}{
      \vspace{-1.00mm}
      \item Both LI and ML classifiers exhibit high accuracy up to \perval{93.1}, always \perval{{>}89.1} in \orient and \perval{{>}85.8} in \random. 
      %
    \item Their accuracy is comparable; relative differences
      depend on the classifiers compared but are always
      \perval{{<}5.4}.
    \item Among LI classifiers, \pdthr generally has the highest
      accuracy, and \clnlthr generally outperforms \clthr.
      \item Among ML classifiers, \rem \emph{always} outperforms
      \xgb or performs similarly to it; moreover, it is the \emph{best} choice across all classifiers in \chfive. 
    \item ML classifiers generally
        outperform 
        LI ones in \orient, \ie the dataset on which they are
        trained. However, their accuracy decreases in \mobile, where
        LI are often preferable, confirming that ML~classifiers are
        negatively affected by unseen environments. 
      }
           \finding{\qtwo}{
      \vspace{-1.00mm}
    \item Yes, combining LI and ML approaches can improve
      accuracy. Especially in \conf{\chtwofour}{\random}, consensus
      classifiers involving \pd and either ML achieve a higher
      accuracy 
      than using any of the pure LI or ML classifiers in isolation.
      \item \xgb-inferred thresholds for LI (\lixgb) 
      and training \xgb on LI 
      (\xgbclnlpd) generally have negligible impact on accuracy. 
    \item Conditional classifiers do not achieve higher accuracy, but
      help bridging the gap between pure LI and ML classifiers in
      \orient at a fraction of the runtime and energy overhead of ML ones.
      } 
      \finding{\qthree}{     
      \vspace{-1.00mm}
      \item Yes: \emph{all} classifiers experience the lowest accuracy ($53$--$70$\%) within a $\pm$\,30\degree transitional area around the \nlosth angle chosen for labeling LOS/WNLOS. 
      \item However, ranging errors in this area are small, ${<}18$~cm on average, therefore misclassifications have a reduced impact. Instead, 
      outside this area, when WNLOS detection matters the most as prominent occlusions cause very large ranging errors, all classifiers
      yield \perval{{\approx}100} accuracy.
      }
      \finding{\qfour}{
      \vspace{-1.00mm}
        \item The answer strongly depends not only on the classifier but
          also on the radio channel and optimization chosen.
      \item \emph{All} \cl-based classifiers, both pure and hybrid,
        \emph{always} yield the lowest \pnn, as low as $18$\,cm.
      \item \cl-based classifiers, and most consensus ones, often
        classify the least amount of samples as LOS, potentially
        leading to unnecessary discarding.
      \item Overall, the ranging error of LOS samples shows
        significant differences across classifiers only in some
        combinations of channel and optimization, mostly involving \chfive.
      \item \conf{\chfour}{\closlong} is the best configuration,
        yielding errors ${<}\,1$\,m for \emph{all} classifiers.
      \item The highest errors occur in \chfive, especially in \clos;
        \closlong is effective in reducing them significantly for all
      classifiers.
      } 
    \finding{\qfive}{ \vspace{-1.00mm}
      \item Both costs vary dramatically across classifiers and radio configurations. 
      \item The difference in energy among  classifiers is  small compared to the cost of the \sstwr exchange, except for \rem. 
      %
      \item Runtime differences are instead significant, up to
          $4.5\times$, even when taking into account \sstwr.
      \item Classifiers based on \clnl and \pd are the fastest and most energy-efficient. 
      \item \closlong incurs $3.8\times$ higher energy and $3\times$ higher runtime than \clos.
      \item \chtwo and \chfour need the least energy; \chfive the highest (\perval{+6.5}).
      %
      %
      }
      \finding{\qsix}{
      \vspace{-1.00mm}
      \item Both channel and radio optimization strongly affect the classifiers' accuracy, ranging error of LOS-classified samples, runtime, and energy, as already hinted at by some of the answers above.
      \item They also affect the \emph{relative} performance among
        classifiers. 
        For instance, in \orient, \xgb is among the 4~most accurate
        classifiers in \conf{\chfour}{\clos}, all within 0.5\,\%
        difference,~but~is~outperformed by 14~classifiers in
        \conf{\chfour}{\closlong}.
      \item \chfive experiences the highest attenuation; when
          combined with the \mobile scenario, this causes an accuracy
          degradation up to $-4.4$\% w.r.t. \orient across all
          classifiers.
      %
      %
      %
    \item \closlong does not improve classification accuracy
      significantly compared to \clos, but it helps to lower the error
      of LOS-classified samples, albeit at the price of significantly higher runtime and energy overhead.
    \item The use of \cnlos is discouraged, as it often underestimates
      distances in LOS and yields impossible negative values.  }
    \finding{\qseven}{ \vspace{-1.00mm}
      \item It is impossible to say without considering application requirements. 
      \item Our recommender tool identifies the most suitable classifier based on the relative weight of application requirements.}
    \end{tabularx}
    \renewcommand{\arraystretch}{1.00}
  \end{table}


%% file: content/recommendations.tex
\section{\rectool: Application-aware Classifier Selection}
\label{sec:recommendations}

\noindent
We facilitate the process of determining which classifiers are best
suited for the application at hand with \rectool, our Python-based
recommender tool. 

\begin{figure}[!t]
    \centering
    \includegraphics[width=\columnwidth]{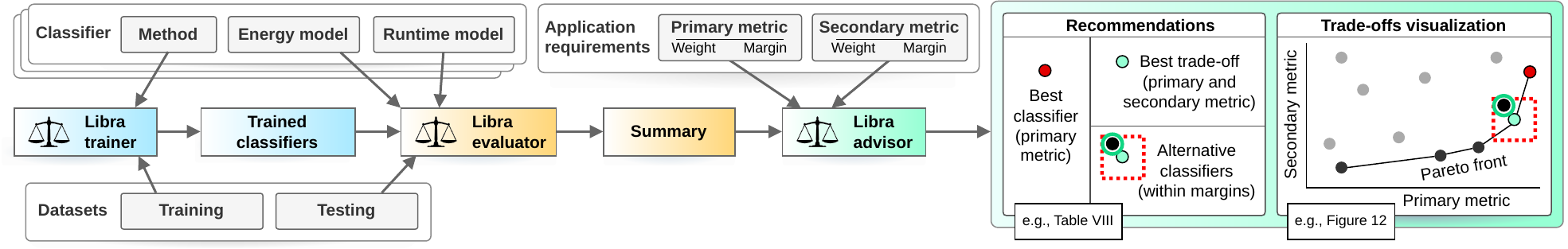}
    \caption{Overview of \rectool's expected input data (including the application requirements and their weights) and outputs (in graphical and tabular form).}
    \label{fig:rectool}
\end{figure}

\libra expects the following inputs from the user
(Fig.\,\ref{fig:rectool}). The \emph{datasets} capture the UWB ranging
performance in the environment and radio settings of interest. Each
entry contains the ranging estimate obtained from a single \sstwr,
labeled as LOS or WNLOS, along with channel and radio
optimization used as well as the recorded radio diagnostics and CIR
portion necessary to compute the outcome of classifiers. In general,
both a \emph{training} and \emph{testing} dataset is provided, as in
our case where we used \pose for the former and \mobile for the
latter. However, a single dataset can be used for both purposes, \eg
to verify the absence of overfitting via \kfold, as we did for
\pose. In this case, supported by \libra but not shown in the figure,
the information necessary to derive appropriate \kfold splits
(\ref{subsec:training_setup}) must also be supplied.  All datasets in
this work are publicly released\footref{fn:links} with \rectool to provide users with an
example of how to generate their own datasets; moreover, they can be
used by readers to explore in more detail the trade-offs reported in
this work.

Next, the user must select the \emph{classifiers} to be analyzed~by
\libra. All classifiers in this work are available;
new,~custom ones can be easily integrated. Classifiers are executed by the \libra
\emph{trainer} with the training dataset as input, generating
\emph{automatically} the trained version of each selected classifier,
which could then be reused for different testing datasets. Then, the \libra
\emph{evaluator} takes as input these trained classifiers, along with
the runtime and energy consumption for each classifier based on our
measurements (\ref{subsec:energy_runtime}) or ad-hoc ones, and
generates a \textit{summary} of their performance. 

This summary is directly available to the user; however, the large
amount of data in it makes it very difficult to parse and interpret to
distill actionable insights. This is instead the goal of the \libra
\emph{advisor}, which takes as input the summary along with the
\emph{application requirements} guiding the identification of the
best-suited classifiers, based on the \emph{performance metrics} we
considered, \ie classification accuracy, ranging error of
LOS-classified samples, runtime, and energy consumption. The user must
select two of these metrics and specify their relative importance and
weight; for instance, specifying \recwg{0.8}{0.2} for classification
accuracy and energy consumption sets the former as \emph{primary}
metric with weight \perval{80} and the latter as \emph{secondary}
metric with~weight~\perval{20}.~Moreover,~as the performance of
classifiers can be very similar, the user can also control how
many classifiers must be considered good candidates by specifying a \emph{margin} for each metric; classifiers whose performance is within
the given margin from the best one are reported along with it. The
corresponding classifiers striking the best trade-offs based on this
user-specified information are determined by the \libra advisor via a
multi-objective decomposition framework (\ref{subsec:moo}). Finally,
the resulting recommendations are shown both in \textit{graphical and
  tabular form} (\ref{subsec:output}), thereby providing users with
intuitive and immediately actionable information, complemented by all quantitative data.

\subsection{Determining the Classifiers with the Best Trade-off}
\label{subsec:moo}

\rectool uses a multi-objective decomposition approach to identify the
classifiers that best satisfy the trade-offs imposed by the
application requirements provided as input. 

\rectool first finds the
Pareto-optimal points, forming the Pareto front, \ie the set of
classifiers for which no alternative can improve classification
accuracy without also increasing energy consumption, and vice versa.
Then, it performs min-max scaling on both objectives\footnote{Finding
  trade-offs among more than two metrics is possible, but at the cost
  of greatly increasing the complexity of finding Pareto-optimal
  solutions~\cite{ALLMENDINGER2022105857} and of hampering the
  intuitive understanding of why a classifier is chosen.} and uses an
augmented achievement scalarization
function~(AASF)~\cite{WIERZBICKI1982391} to transform the
multi-objective problem into a single solution respecting the given
relative weights.

\fakepar By default, the AASF finds the best classifier with the
provided weight distribution. However, \rectool shows classifiers
performing within a user-specified margin, defined as a percentage
over the range of values after min-max scaling. 
The margin must be determined empirically: if it is set too small, no classifiers other than the best are shown and valid alternatives could be missed; if it is too large, \libra might present the user with too many ``best'' classifiers, defeating its very purpose.

\subsection{\rectool in Action} 
\label{subsec:output}

\noindent
We now offer an illustration of the use and value of \libra, focusing
on classification accuracy and energy consumption. We use the \mobile
dataset, for which we empirically determined that the values
\mbox{5\%}, \mbox{3\%}, \mbox{1\%}, and \mbox{1\%} as margins for
accuracy, energy, runtime, and \pnn error, respectively, yield a
reasonable number of alternative classifiers. 

\begin{figure}[!t]
    \centering
    \includegraphics[width=0.72\columnwidth]{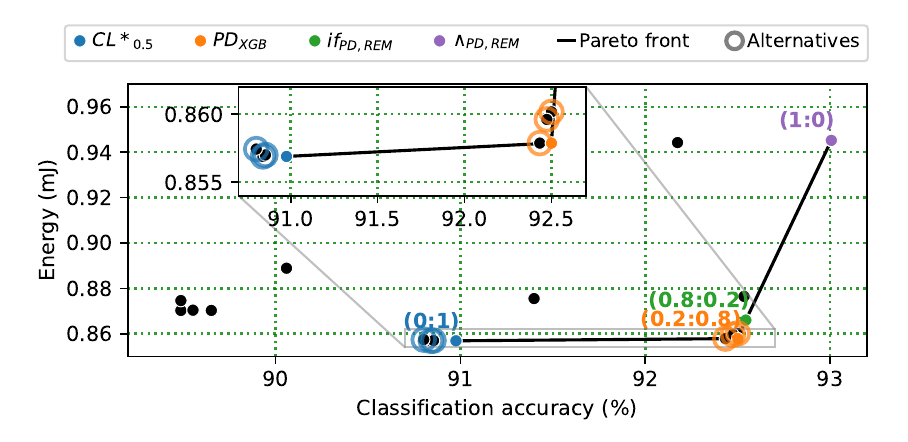}
    \vspace{-2mm}
    \caption{\libra: graphical output. The example shows the
        trade-offs in \random for \conf{\chfour}{\closlong}, when considering
        classification accuracy and energy consumption as metrics. The solid black curve is the Pareto front showing the best classifiers
        that maximize classification accuracy and minimize energy
        consumption. \rectool finds the best trade-off to the
        multi-objective problem based on the chosen weight
        distributions, annotated in the graph, and also circles any
            close-by alternatives within a given margin.}
    \label{fig:pareto_example}
\end{figure}

\boldpar{Navigating the trade-offs graphically.}
Fig.\,\ref{fig:pareto_example} shows the graphical output of \rectool for \chfour and \closlong, when considering classification accuracy and energy consumption as metrics and \random as dataset. Each dot
represents the performance of a single classifier, while the curve
shows the Pareto front, containing the optimal classifiers (in color)
for the specified weights. Crucially, these recommendations
\emph{change} depending on the weight: for instance, with
\recwg{1}{0}, \ie only classification accuracy matters, \rectool
recommends \pdandrem, while with 
\recwg{0.2}{0.8}, \ie greater
weight to energy consumption, the best trade-off is instead
\pdxgb. However, in this latter case, three other classifiers are within our specified margins, marked with a colored circle; their difference in performance \wrt the best classifier is reported in the textual output, described next.

\input{content/new_libra_table}

\boldpar{Navigating the trade-offs textually.} The graphical
representation is immediate and intuitive, but a finer-grained
quantitative analysis including \emph{all} metrics is often desirable,
\eg to verify that the values of those not selected is still
reasonable.  To this end, \rectool also provides output in tabular
form by reporting for each metric:
\begin{inparaenum}
\item the \emph{absolute best} classifier\footnote{In some cases,
    there may be multiple best classifiers. For example, when
    considering energy consumption, \clnlthr and \clnlxgb perform
    equally.} for the primary metric considered in isolation, \ie as if weights \recwg{1}{0} were specified, and
\item the classifiers striking the best trade-off between primary and secondary metric, based on user-specified weights and margins.
\end{inparaenum}
The former serves as a baseline of the best possible performance
achievable with the primary metric of interest; the others offer a
view of the trade-offs at stake in terms of the difference in
performance when finding the best compromise with a secondary metric.

As an example, Tables~\ref{tab:rectable_abs_accen}--\ref{tab:rectable_abs_accen:two} 
show the output of \libra for each combination of channel and radio
optimization, using weights \recwg{0.8}{0.2}. When classification
accuracy is the only metric to optimize for (top of
Table~\ref{tab:rectable_abs_accen}), the best classifier is
determined only by the radio settings: \pdandrem in \chtwo and
\chfour, and \rem in \chfive. However, when a secondary metric is
considered with \perval{20} weight, only \rem remains the best across
all combinations of classification accuracy and the other three
metrics, but only in \conf{\chfive}{\clos}; in the other settings,
other classifiers are sometimes better---and with significant
benefits. For instance, in \conf{\chtwo,\chfour}{\clos} the use of
\pdandxgb decreases energy by \perval{-21.81} and runtime by
\perval{-57.88}, with only a limited absolute decrease in accuracy,
\perval{-0.20} and \perval{-0.34} depending on the channel. 
Similar recommendations hold for \conf{\chfive}{\closlong}, where
nonetheless different classifiers can be exploited to strike different
compromises within the assigned margins, with \xgbclnlpd notably
yielding the smallest absolute decrease in accuracy (\perval{-0.76}) and
a very high relative gain in runtime (\perval{-52.90}).

When focusing on energy consumption as the primary metric, the set of
best candidates changes significantly. Across all radio settings
shown, the absolute best when considering this metric alone are
\clnlthr and \clnlxgb; their energy is exactly the same, given that
they both operate based on a threshold, which for \clnlxgb is computed
offline via \xgb. However, when considering \emph{any} of the other
metrics as secondary, these classifiers are \emph{never} the best
trade-off. For instance, when considering also accuracy in
\conf{\chfour}{\closlong}, all of \pdthr, \xgbclnlpd, \ifelsepdxgb
(circled in orange in Fig.\,\ref{fig:pareto_example}), and \pdxgb
(solid orange) yield an absolute improvement of
\perval{1.60\text{--}1.83} compared to \clnlthr and \clnlxgb---a
significant gain given the small differences in accuracy among
classifiers (\ref{subsec:accuracy_eval}), that comes with only a
negligible relative penalty of \perval{0.11\text{--}0.38} in energy
consumption. In \conf{\chfive}{\closlong} differences are even larger,
with \xgbclnlpd improving accuracy by \perval{3.02} and~\perval{3.22}
in absolute compared to \clnlthr and \clnlxgb, with just a
\perval{0.36} penalty in energy. Similar considerations hold when
considering energy in conjunction with the other secondary metric; for
instance, choosing \ifelsepdxgb reduces the \pnn by $3.26$~m
($-84.84$\%) \wrt the two classifiers based on \clnl, with a penalty
in energy of only $3~\mu$J ($+1.31$\%).

Table~\ref{tab:rectable_abs_accen:two} also shows the best classifiers
when selecting runtime or \pnn as the primary metric.  For runtime,
\clnlthr and \clnlxgb represent again the best choice across all
channels and radio configurations. However, when considering also
secondary requirements, choosing \ifelsepdxgb enables a marked
increase in accuracy, up to $+3.10$\% \wrt \clnlxgb, and obtain the
same high reduction in the \pnn of LOS-classified errors already
commented for energy, while only requiring an additional $1.6$~ms of
runtime. 
When choosing \pnn as primary metric, the best configuration is always
given by consensus classifiers involving \cl, \rem, or
  both. 
  However, when considering secondary requirements, the choice of
  hybrid strategy and of LI and ML components matters. For instance,
  in \conf{\chtwo}{\clos}, swapping \cl for \pd by using \pdandrem
  does not affect \pnn but yields an absolute \perval{1.35} increase
  in accuracy. Similarly, in \conf{\chfour}{\closlong}, using the
  conditional classifier \ifelsepdrem of \pdandrem yields a high
  reduction in runtime (\perval{-51.28}) with only a negligible
  increase of $4$~mm (\perval{0.09}) in \pnn.

Overall, specific values aside, Tables~\ref{tab:rectable_abs_accen}--\ref{tab:rectable_abs_accen:two} concretely
and quantitatively reiterate that a one-size-fits-all classifier does not exist. 
On the contrary, the identification of the best choice requires a multi-dimensional analysis tackling the interplay of radio configuration, performance requirements, and the intrinsic
characteristics of each classifier. 
\libra removes the burden of this complex analysis via direct automated support, providing developers with actionable information based on their own datasets and goals. 

%% file: content/new_libra_table.tex
\begin{table}
\centering
\footnotesize
\caption{\libra: tabular output (1/2). \vspace*{-3.75mm}}
Example of the tabular output of \libra for different primary
requirements (classification accuracy and energy consumption) when
using the \mobile dataset. For each combination of radio settings, the
third column contains the ``absolute best'' classifier maximizing the
corresponding (primary) metric when considered alone, \ie weight
\recwg{1}{0}. The remaining columns show the best classifiers with
weight \recwg{0.8}{0.2}, \ie \perval{80} for the same primary metric
and \perval{20} for each of the remaining three, along with the
corresponding variation \wrt the absolute best. When the latter is
achieved by multiple classifiers (as with energy consumption and
runtime), the variations are shown \wrt \emph{each} of
them. Variations are shown as absolute values for all metrics
  and in percentage for all except accuracy, to avoid confusion among
  percentages.  \vspace*{1.5mm}
\vspace*{1.5mm}
\label{tab:rectable_abs_accen}
\scriptsize
\setlength\tabcolsep{1.2pt}
\renewcommand{\arraystretch}{2} 
\begin{tabularx}{\textwidth}{|c|c|c|c|c|c|c|c|c|c|c|c|}
\cline{1-12}
  \multicolumn{2}{|c|}{\multirow{3}{*}{\textbf{Radio settings}}} & \multirow{2}{*}{\textbf{Accuracy \perval{100}}} & \multicolumn{9}{c|}{\textbf{Classification accuracy \perval{80} and \perval{20} for the other metric}}\\
\cline{4-12} 
\multicolumn{2}{|c|}{}  & & \multicolumn{3}{c|}{\textbf{Energy consumption}} & \multicolumn{3}{c|}{\textbf{Runtime}} & \multicolumn{3}{c|}{\textbf{\pnn error (LOS)}} \\
\cline{3-12} 
\multicolumn{2}{|c|}{} & Classifier & Classifier & \da & \de & Classifier & \da & \dr & Classifier & \da & \dpnn\\
\cline{1-12} 
\multirow[c]{2}{*}{$\vcenter{\hbox{\chtwo}}$} & $\vcenter{\hbox{\clos}}$ & $\vcenter{\hbox{\shortstack{\pdandrem\\(93.10\%)}}}$ & $\vcenter{\hbox{\pdandxgb}}$ & $\vcenter{\hbox{-0.20}}$ & $\vcenter{\hbox{\shortstack{-69\\(-21.82\%)}}}$ & $\vcenter{\hbox{\pdandxgb}}$ & $\vcenter{\hbox{-0.20}}$ & $\vcenter{\hbox{\shortstack{-2712\\(-57.88\%)}}}$ & $\vcenter{\hbox{\pdandrem}}$ & $\vcenter{\hbox{-}}$ & $\vcenter{\hbox{-}}$ \\
\cline{2-12}
 & $\vcenter{\hbox{\closlong}}$ & $\vcenter{\hbox{\shortstack{\pdandrem\\(92.54\%)}}}$ & $\vcenter{\hbox{\pdandxgb}}$ & $\vcenter{\hbox{-0.34}}$ & $\vcenter{\hbox{\shortstack{-69\\(-7.27\%)}}}$ & $\vcenter{\hbox{\pdandxgb}}$ & $\vcenter{\hbox{-0.34}}$ & $\vcenter{\hbox{\shortstack{-2712\\(-42.26\%)}}}$ & $\vcenter{\hbox{\pdandrem}}$ & $\vcenter{\hbox{-}}$ & $\vcenter{\hbox{-}}$ \\
\cline{1-12}
\multirow[c]{2}{*}{$\vcenter{\hbox{\chfour}}$} & $\vcenter{\hbox{\clos}}$ & $\vcenter{\hbox{\shortstack{\pdandrem\\(92.65\%)}}}$ & $\vcenter{\hbox{\pdandxgb}}$ & $\vcenter{\hbox{-0.37}}$ & $\vcenter{\hbox{\shortstack{-69\\(-21.82\%)}}}$ & $\vcenter{\hbox{\pdandxgb}}$ & $\vcenter{\hbox{-0.37}}$ & $\vcenter{\hbox{\shortstack{-2712\\(-57.88\%)}}}$ & $\vcenter{\hbox{\pdandrem}}$ & $\vcenter{\hbox{-}}$ & $\vcenter{\hbox{-}}$ \\
\cline{2-12}
 & $\vcenter{\hbox{\closlong}}$ & $\vcenter{\hbox{\shortstack{\pdandrem\\(93.01\%)}}}$ & $\vcenter{\hbox{\ifelsepdrem}}$ & $\vcenter{\hbox{-0.50}}$ & $\vcenter{\hbox{\shortstack{-79\\(-8.38\%)}}}$ & $\vcenter{\hbox{\ifelsepdrem}}$ & $\vcenter{\hbox{-0.50}}$ & $\vcenter{\hbox{\shortstack{-3291\\(-51.28\%)}}}$ & $\vcenter{\hbox{\pdandrem}}$ & $\vcenter{\hbox{-}}$ & $\vcenter{\hbox{-}}$ \\
\cline{1-12}
\multirow[c]{2}{*}{$\vcenter{\hbox{\chfive}}$} & $\vcenter{\hbox{\clos}}$ & $\vcenter{\hbox{\shortstack{\rem\\(90.67\%)}}}$ & $\vcenter{\hbox{\rem}}$ & $\vcenter{\hbox{-}}$ & $\vcenter{\hbox{-}}$ & $\vcenter{\hbox{\rem}}$ & $\vcenter{\hbox{-}}$ & $\vcenter{\hbox{-}}$ & $\vcenter{\hbox{\rem}}$ & $\vcenter{\hbox{-}}$ & $\vcenter{\hbox{-}}$ \\
\cline{2-12}
 & \multirow[c]{3}{*}{$\vcenter{\hbox{\closlong}}$} & \multirow[c]{3}{*}{$\vcenter{\hbox{\shortstack{\rem\\(89.70\%)}}}$} & $\vcenter{\hbox{\pdxgb}}$ & $\vcenter{\hbox{-0.89}}$ & $\vcenter{\hbox{\shortstack{-84\\(-8.44\%)}}}$ & $\vcenter{\hbox{\xgbclnlpd}}$ & $\vcenter{\hbox{-0.76}}$ & $\vcenter{\hbox{\shortstack{-3289\\(-52.90\%)}}}$ & \multirow[c]{3}{*}{$\vcenter{\hbox{\rem}}$} & \multirow[c]{3}{*}{$\vcenter{\hbox{-}}$} & \multirow[c]{3}{*}{$\vcenter{\hbox{-}}$} \\
\cline{4-6}\cline{7-9}
 &  &  & $\vcenter{\hbox{\xgbclnlpd}}$ & $\vcenter{\hbox{-0.76}}$ & $\vcenter{\hbox{\shortstack{-82\\(-8.21\%)}}}$ & \multirow{2}{*}{$\vcenter{\hbox{\ifelsepdxgb}}$} & \multirow{2}{*}{$\vcenter{\hbox{-0.92}}$} & \multirow{2}{*}{$\vcenter{\hbox{\shortstack{-3317\\(-53.36\%)}}}$} &  &  &  \\
\cline{4-6}
 &  &  & $\vcenter{\hbox{\ifelsepdxgb}}$ & $\vcenter{\hbox{-0.92}}$ & $\vcenter{\hbox{\shortstack{-82\\(-8.21\%)}}}$ &  &  &  &  &  &  \\
\cline{1-12}
  \multicolumn{2}{|c|}{\multirow{3}{*}{\textbf{Radio settings}}} & \multirow{2}{*}{\textbf{Energy \perval{100}}} & \multicolumn{9}{c|}{\textbf{Energy consumption \perval{80} and \perval{20} for the other metric}}\\
\cline{4-12} 
\multicolumn{2}{|c|}{}  & & \multicolumn{3}{c|}{\textbf{Classification accuracy}} & \multicolumn{3}{c|}{\textbf{Runtime}} & \multicolumn{3}{c|}{\textbf{\pnn error (LOS)}} \\
\cline{3-12} 
\multicolumn{2}{|c|}{} & Classifier & Classifier & \de & \da & Classifier & \de & \dr & Classifier & \de & \dpnn\\
\cline{1-12} 
\multirow[c]{2}{*}{$\vcenter{\hbox{\chtwo}}$} & $\vcenter{\hbox{\clos}}$ & $\vcenter{\hbox{\shortstack{\clnlthr/\clnlxgb\\(227\,µJ)}}}$ & $\vcenter{\hbox{\pdthr}}$ & $\vcenter{\hbox{\shortstack{+1\\(+0.43\%)}}}$ & $\vcenter{\hbox{+1.22 / +1.18}}$ & $\vcenter{\hbox{\ifelseclnlxgb}}$ & $\vcenter{\hbox{\shortstack{+0.3\\(+0.14\%)}}}$ & $\vcenter{\hbox{\shortstack{+16\\(+1.59\%)}}}$ & $\vcenter{\hbox{\ifelsepdxgb}}$ & $\vcenter{\hbox{\shortstack{+3\\(+1.24\%)}}}$ & $\vcenter{\hbox{\shortstack{-1.7 / -2.2\\(-3.27\% / -4.26\%)}}}$ \\
\cline{2-12}
 & $\vcenter{\hbox{\closlong}}$ & $\vcenter{\hbox{\shortstack{\clnlthr/\clnlxgb\\(857\,µJ)}}}$ & $\vcenter{\hbox{\pdthr}}$ & $\vcenter{\hbox{\shortstack{+1\\(+0.11\%)}}}$ & $\vcenter{\hbox{+1.22 / +1.10}}$ & $\vcenter{\hbox{\ifelseclnlxgb}}$ & $\vcenter{\hbox{\shortstack{+0.4\\(+0.04\%)}}}$ & $\vcenter{\hbox{\shortstack{+18\\(+0.64\%)}}}$ & $\vcenter{\hbox{\pdthr}}$ & $\vcenter{\hbox{\shortstack{+1\\(+0.11\%)}}}$ & $\vcenter{\hbox{\shortstack{-0.69 / -0.70\\(-1.51\% / -1.53\%)}}}$ \\
\cline{1-12}
\multirow[c]{4}{*}{$\vcenter{\hbox{\chfour}}$} & \multirow[c]{2}{*}{$\vcenter{\hbox{\clos}}$} & \multirow[c]{2}{*}{$\vcenter{\hbox{\shortstack{\clnlthr/\clnlxgb\\(227\,µJ)}}}$} & \multirow[c]{2}{*}{$\vcenter{\hbox{\pdthr}}$} & \multirow[c]{2}{*}{$\vcenter{\hbox{\shortstack{+1\\(+0.43\%)}}}$} & \multirow[c]{2}{*}{$\vcenter{\hbox{+1.54 / +1.63}}$} & $\vcenter{\hbox{\ifelseclnlxgb}}$ & $\vcenter{\hbox{\shortstack{+0.1\\(+0.04\%)}}}$ & $\vcenter{\hbox{\shortstack{+4\\(+0.44\%)}}}$ & \multirow[c]{2}{*}{$\vcenter{\hbox{\ifelsepdrem}}$} & \multirow[c]{2}{*}{$\vcenter{\hbox{\shortstack{+9\\(+4.09\%)}}}$} & \multirow[c]{2}{*}{$\vcenter{\hbox{\shortstack{-0.6 / -0.7\\(-1.43\% / -1.54\%)}}}$} \\
\cline{7-9}
 &  &  &  &  &  & $\vcenter{\hbox{\ifelseclnlrem}}$ & $\vcenter{\hbox{\shortstack{+0.4\\(+0.18\%)}}}$ & $\vcenter{\hbox{\shortstack{+17\\(+1.73\%)}}}$ &  &  &  \\
\cline{2-12}
 & \multirow[c]{4}{*}{$\vcenter{\hbox{\closlong}}$} & \multirow[c]{4}{*}{$\vcenter{\hbox{\shortstack{\clnlthr/\clnlxgb\\(857\,µJ)}}}$} & $\vcenter{\hbox{\pdthr}}$ & $\vcenter{\hbox{\shortstack{+1\\(+0.11\%)}}}$ & $\vcenter{\hbox{+1.60 / +1.75}}$ & $\vcenter{\hbox{\ifelseclnlxgb}}$ & $\vcenter{\hbox{\shortstack{+0.1\\(+0.01\%)}}}$ & $\vcenter{\hbox{\shortstack{+6\\(+0.21\%)}}}$ & $\vcenter{\hbox{\pdthr}}$ & $\vcenter{\hbox{\shortstack{+1\\(+0.11\%)}}}$ & $\vcenter{\hbox{\shortstack{-0.4 / -0.5\\(-1.09\% / -1.23\%)}}}$ \\
\cline{4-6}\cline{7-9}\cline{10-12}
 &  &  & $\vcenter{\hbox{\pdxgb}}$ & $\vcenter{\hbox{\shortstack{+1\\(+0.11\%)}}}$ & $\vcenter{\hbox{+1.68 / +1.83}}$ & \multirow{3}{*}{$\vcenter{\hbox{\ifelseclnlrem}}$} & \multirow{3}{*}{$\vcenter{\hbox{\shortstack{+1\\(+0.06\%)}}}$} & \multirow{3}{*}{$\vcenter{\hbox{\shortstack{+22\\(+0.82\%)}}}$} & $\vcenter{\hbox{\pdxgb}}$ & $\vcenter{\hbox{\shortstack{+1\\(+0.11\%)}}}$ & $\vcenter{\hbox{\shortstack{-0.4 / -0.5\\(-1.11\% / -1.25\%)}}}$ \\
\cline{4-6}\cline{10-12}
 &  &  & $\vcenter{\hbox{\xgbclnlpd}}$ & $\vcenter{\hbox{\shortstack{+3\\(+0.38\%)}}}$ & $\vcenter{\hbox{+1.68 / +1.83}}$ &  &  &  & \multirow{2}{*}{$\vcenter{\hbox{\xgbclnlpd}}$} & \multirow{2}{*}{$\vcenter{\hbox{\shortstack{+3\\(+0.38\%)}}}$} & \multirow{2}{*}{$\vcenter{\hbox{\shortstack{-0.4 / -0.5\\(-1.11\% / -1.25\%)}}}$} \\
\cline{4-6}
 &  &  & $\vcenter{\hbox{\ifelsepdxgb}}$ & $\vcenter{\hbox{\shortstack{+3\\(+0.32\%)}}}$ & $\vcenter{\hbox{+1.65 / +1.80}}$ &  &  &  &  &  &  \\
\cline{1-12}
\multirow[c]{2}{*}{$\vcenter{\hbox{\chfive}}$} & $\vcenter{\hbox{\clos}}$ & $\vcenter{\hbox{\shortstack{\clnlthr/\clnlxgb\\(241\,µJ)}}}$ & $\vcenter{\hbox{\ifelsepdxgb}}$ & $\vcenter{\hbox{\shortstack{+3\\(+1.31\%)}}}$ & $\vcenter{\hbox{+2.87 / +3.10}}$ & $\vcenter{\hbox{\ifelseclnlxgb}}$ & $\vcenter{\hbox{\shortstack{+0.4\\(+0.17\%)}}}$ & $\vcenter{\hbox{\shortstack{+20\\(+1.98\%)}}}$ & $\vcenter{\hbox{\ifelsepdxgb}}$ & $\vcenter{\hbox{\shortstack{+3\\(+1.31\%)}}}$ & $\vcenter{\hbox{\shortstack{-326 / -300\\(-84.84\% / -83.73\%)}}}$ \\
\cline{2-12}
 & \multirow[c]{3}{*}{$\vcenter{\hbox{\closlong}}$} & \multirow[c]{3}{*}{$\vcenter{\hbox{\shortstack{\clnlthr/\clnlxgb\\(912\,µJ)}}}$} & $\vcenter{\hbox{\pdxgb}}$ & $\vcenter{\hbox{\shortstack{+1\\(+0.11\%)}}}$ & $\vcenter{\hbox{+2.88 / +3.07}}$ & \multirow[c]{3}{*}{$\vcenter{\hbox{\ifelseclnlxgb}}$} & \multirow[c]{3}{*}{$\vcenter{\hbox{\shortstack{+0.3\\(+0.04\%)}}}$} & \multirow[c]{3}{*}{$\vcenter{\hbox{\shortstack{+16\\(+0.60\%)}}}$} & $\vcenter{\hbox{\clthr}}$ & $\vcenter{\hbox{\shortstack{+13\\(+1.47\%)}}}$ & $\vcenter{\hbox{\shortstack{-1.1 / -1.0\\(-2.44\% / -2.14\%)}}}$ \\
\cline{4-6}\cline{10-12}
 &  &  & $\vcenter{\hbox{\xgbclnlpd}}$ & $\vcenter{\hbox{\shortstack{+3\\(+0.36\%)}}}$ & $\vcenter{\hbox{+3.02 / +3.22}}$ &  &  &  & $\vcenter{\hbox{\clxgb}}$ & $\vcenter{\hbox{\shortstack{+13\\(+1.47\%)}}}$ & $\vcenter{\hbox{\shortstack{-1.1 / -1.0\\(-2.43\% / -2.13\%)}}}$ \\
\cline{4-6}\cline{10-12}
 &  &  & $\vcenter{\hbox{\ifelsepdxgb}}$ & $\vcenter{\hbox{\shortstack{+3\\(+0.36\%)}}}$ & $\vcenter{\hbox{+2.85 / +3.04}}$ &  &  &  & $\vcenter{\hbox{\ifelseclxgb}}$ & $\vcenter{\hbox{\shortstack{+14\\(+1.50\%)}}}$ & $\vcenter{\hbox{\shortstack{-1.1 / -1.0\\(-2.44\% / -2.14\%)}}}$ \\
\cline{1-12}
\end{tabularx}
\renewcommand{\arraystretch}{1.00} 
\end{table}

\begin{table}
\centering
\footnotesize
\caption{\libra: tabular output (2/2). \vspace*{-3.75mm}}
\vspace*{1.5mm}
\label{tab:rectable_abs_accen:two}
\scriptsize
\setlength\tabcolsep{0.7pt}
\renewcommand{\arraystretch}{2} 
\begin{tabularx}{\textwidth}{|c|c|c|c|c|c|c|c|c|c|c|c|}
\cline{1-12}
  \multicolumn{2}{|c|}{\multirow{3}{*}{\textbf{Radio settings}}} & \multirow{2}{*}{\textbf{Runtime \perval{100}}} & \multicolumn{9}{c|}{\textbf{Runtime \perval{80} and \perval{20} for the other metric}}\\
\cline{4-12} 
\multicolumn{2}{|c|}{}  & & \multicolumn{3}{c|}{\textbf{Energy consumption}} & \multicolumn{3}{c|}{\textbf{Classification accuracy}} & \multicolumn{3}{c|}{\textbf{\pnn error (LOS)}} \\
\cline{3-12} 
\multicolumn{2}{|c|}{} & Classifier & Classifier & \dr & \de & Classifier & \dr & \da & Classifier & \dr & \dpnn\\
\cline{1-12} 
\multirow[c]{2}{*}{$\vcenter{\hbox{\chtwo}}$} & $\vcenter{\hbox{\clos}}$ & $\vcenter{\hbox{\shortstack{\clnlthr/\clnlxgb\\(996\,µs)}}}$ & $\vcenter{\hbox{\ifelseclnlxgb}}$ & $\vcenter{\hbox{\shortstack{+16\\(+1.59\%)}}}$ & $\vcenter{\hbox{\shortstack{+0.3\\(+0.14\%)}}}$ & $\vcenter{\hbox{\pdthr}}$ & $\vcenter{\hbox{\shortstack{+60\\(+6.02\%)}}}$ & $\vcenter{\hbox{+1.22 / +1.18}}$ & $\vcenter{\hbox{\ifelsepdxgb}}$ & $\vcenter{\hbox{\shortstack{+150\\(+15.06\%)}}}$ & $\vcenter{\hbox{\shortstack{-1.7 / -2.2\\(-3.27\% / -4.26\%)}}}$ \\
\cline{2-12}
 & $\vcenter{\hbox{\closlong}}$ & $\vcenter{\hbox{\shortstack{\clnlthr/\clnlxgb\\(2728\,µs)}}}$ & $\vcenter{\hbox{\ifelseclnlxgb}}$ & $\vcenter{\hbox{\shortstack{+18\\(+0.64\%)}}}$ & $\vcenter{\hbox{\shortstack{+0.4\\(+0.04\%)}}}$ & $\vcenter{\hbox{\pdthr}}$ & $\vcenter{\hbox{\shortstack{+60\\(+2.20\%)}}}$ & $\vcenter{\hbox{+1.22 / +1.10}}$ & $\vcenter{\hbox{\pdthr}}$ & $\vcenter{\hbox{\shortstack{+60\\(+2.20\%)}}}$ & $\vcenter{\hbox{\shortstack{-0.69 / -0.70\\(-1.51\% / -1.53\%)}}}$ \\
\cline{1-12}
\multirow[c]{4}{*}{$\vcenter{\hbox{\chfour}}$} & \multirow[c]{2}{*}{$\vcenter{\hbox{\clos}}$} & \multirow[c]{2}{*}{$\vcenter{\hbox{\shortstack{\clnlthr/\clnlxgb\\(996\,µs)}}}$} & $\vcenter{\hbox{\ifelseclnlxgb}}$ & $\vcenter{\hbox{\shortstack{+4\\(+0.44\%)}}}$ & $\vcenter{\hbox{\shortstack{+0.1\\(+0.04\%)}}}$ & \multirow[c]{2}{*}{$\vcenter{\hbox{\pdthr}}$} & \multirow[c]{2}{*}{$\vcenter{\hbox{\shortstack{+60\\(+6.02\%)}}}$} & \multirow[c]{2}{*}{$\vcenter{\hbox{+1.54 / +1.63}}$} & \multirow[c]{2}{*}{$\vcenter{\hbox{\ifelsepdrem}}$} & \multirow[c]{2}{*}{$\vcenter{\hbox{\shortstack{+405\\(+40.62\%)}}}$} & \multirow[c]{2}{*}{$\vcenter{\hbox{\shortstack{-0.6 / -0.7\\(-1.43\% / -1.54\%)}}}$} \\
\cline{4-6}
 &  &  & $\vcenter{\hbox{\ifelseclnlrem}}$ & $\vcenter{\hbox{\shortstack{+17\\(+1.73\%)}}}$ & $\vcenter{\hbox{\shortstack{+0.4\\(+0.18\%)}}}$ &  &  &  &  &  &  \\
\cline{2-12}
 & \multirow[c]{2}{*}{$\vcenter{\hbox{\closlong}}$} & \multirow[c]{2}{*}{$\vcenter{\hbox{\shortstack{\clnlthr/\clnlxgb\\(2728\,µs)}}}$} & $\vcenter{\hbox{\ifelseclnlxgb}}$ & $\vcenter{\hbox{\shortstack{+6\\(+0.21\%)}}}$ & $\vcenter{\hbox{\shortstack{+0.1\\(+0.01\%)}}}$ & $\vcenter{\hbox{\pdthr}}$ & $\vcenter{\hbox{\shortstack{+60\\(+2.20\%)}}}$ & $\vcenter{\hbox{+1.60 / +1.75}}$ & $\vcenter{\hbox{\pdthr}}$ & $\vcenter{\hbox{\shortstack{+60\\(+2.20\%)}}}$ & $\vcenter{\hbox{\shortstack{-0.4 / -0.5\\(-1.09\% / -1.23\%)}}}$ \\
\cline{4-6}\cline{7-9}\cline{10-12}
 &  &  & $\vcenter{\hbox{\ifelseclnlrem}}$ & $\vcenter{\hbox{\shortstack{+22\\(+0.82\%)}}}$ & $\vcenter{\hbox{\shortstack{+1\\(+0.06\%)}}}$ & $\vcenter{\hbox{\pdxgb}}$ & $\vcenter{\hbox{\shortstack{+60\\(+2.20\%)}}}$ & $\vcenter{\hbox{+1.68 / +1.83}}$ & $\vcenter{\hbox{\pdxgb}}$ & $\vcenter{\hbox{\shortstack{+60\\(+2.20\%)}}}$ & $\vcenter{\hbox{\shortstack{-0.4 / -0.5\\(-1.11\% / -1.25\%)}}}$ \\
\cline{1-12}
\multirow[c]{2}{*}{$\vcenter{\hbox{\chfive}}$} & $\vcenter{\hbox{\clos}}$ & $\vcenter{\hbox{\shortstack{\clnlthr/\clnlxgb\\(996\,µs)}}}$ & $\vcenter{\hbox{\ifelseclnlxgb}}$ & $\vcenter{\hbox{\shortstack{+20\\(+1.98\%)}}}$ & $\vcenter{\hbox{\shortstack{+0.4\\(+0.17\%)}}}$ & $\vcenter{\hbox{\ifelsepdxgb}}$ & $\vcenter{\hbox{\shortstack{+168\\(+16.82\%)}}}$ & $\vcenter{\hbox{+2.87 / +3.10}}$ & $\vcenter{\hbox{\ifelsepdxgb}}$ & $\vcenter{\hbox{\shortstack{+168\\(+16.82\%)}}}$ & $\vcenter{\hbox{\shortstack{-326 / -300\\(-84.84\% / -83.73\%)}}}$ \\
\cline{2-12}
 & $\vcenter{\hbox{\closlong}}$ & $\vcenter{\hbox{\shortstack{\clnlthr/\clnlxgb\\(2728\,µs)}}}$ & $\vcenter{\hbox{\ifelseclnlxgb}}$ & $\vcenter{\hbox{\shortstack{+16\\(+0.60\%)}}}$ & $\vcenter{\hbox{\shortstack{+0.3\\(+0.04\%)}}}$ & $\vcenter{\hbox{\pdxgb}}$ & $\vcenter{\hbox{\shortstack{+60\\(+2.20\%)}}}$ & $\vcenter{\hbox{+2.88 / +3.07}}$ & $\vcenter{\hbox{\pdthr}}$ & $\vcenter{\hbox{\shortstack{+60\\(+2.20\%)}}}$ & $\vcenter{\hbox{\shortstack{-0.4 / -0.3\\(-0.90\% / -0.59\%)}}}$ \\
\cline{1-12}
  \multicolumn{2}{|c|}{\multirow{3}{*}{\textbf{Radio settings}}} & \multirow{2}{*}{\textbf{\pnn error \perval{100}}} & \multicolumn{9}{c|}{\textbf{\pnn error \perval{80} and \perval{20} for the other metric}}\\
\cline{4-12} 
\multicolumn{2}{|c|}{}  & & \multicolumn{3}{c|}{\textbf{Energy consumption}} & \multicolumn{3}{c|}{\textbf{Runtime}} & \multicolumn{3}{c|}{\textbf{Classification accuracy}} \\
\cline{3-12} 
\multicolumn{2}{|c|}{} & Classifier & Classifier & \dpnn & \de & Classifier & \dpnn & \dr & Classifier & \dpnn & \da\\
\cline{1-12} 
\multirow[c]{2}{*}{$\vcenter{\hbox{\chtwo}}$} & $\vcenter{\hbox{\clos}}$ & $\vcenter{\hbox{\shortstack{\clandrem\\(47\,cm)}}}$ & $\vcenter{\hbox{\clandxgb}}$ & $\vcenter{\hbox{\shortstack{+0.3\\(+0.71\%)}}}$ & $\vcenter{\hbox{\shortstack{-56\\(-17.69\%)}}}$ & $\vcenter{\hbox{\clandxgb}}$ & $\vcenter{\hbox{\shortstack{+0.3\\(+0.71\%)}}}$ & $\vcenter{\hbox{\shortstack{-2081\\(-43.77\%)}}}$ & $\vcenter{\hbox{\pdandrem}}$ & $\vcenter{\hbox{\shortstack{<0.001\\(<0.001\%)}}}$ & $\vcenter{\hbox{+1.35}}$ \\
\cline{2-12}
 & $\vcenter{\hbox{\closlong}}$ & $\vcenter{\hbox{\shortstack{\pdandrem\\(44\,cm)}}}$ & $\vcenter{\hbox{\pdandxgb}}$ & $\vcenter{\hbox{\shortstack{+1\\(+1.22\%)}}}$ & $\vcenter{\hbox{\shortstack{-69\\(-7.27\%)}}}$ & $\vcenter{\hbox{\pdandxgb}}$ & $\vcenter{\hbox{\shortstack{+1\\(+1.22\%)}}}$ & $\vcenter{\hbox{\shortstack{-2712\\(-42.26\%)}}}$ & $\vcenter{\hbox{\pdandrem}}$ & $\vcenter{\hbox{-}}$ & $\vcenter{\hbox{-}}$ \\
\cline{1-12}
\multirow[c]{2}{*}{$\vcenter{\hbox{\chfour}}$} & $\vcenter{\hbox{\clos}}$ & $\vcenter{\hbox{\shortstack{\clandrem\\(42\,cm)}}}$ & $\vcenter{\hbox{\clandrem}}$ & $\vcenter{\hbox{-}}$ & $\vcenter{\hbox{-}}$ & $\vcenter{\hbox{\clnlandrem}}$ & $\vcenter{\hbox{\shortstack{+0.1\\(+0.25\%)}}}$ & $\vcenter{\hbox{\shortstack{-129\\(-2.71\%)}}}$ & $\vcenter{\hbox{\pdandrem}}$ & $\vcenter{\hbox{\shortstack{+0.1\\(+0.22\%)}}}$ & $\vcenter{\hbox{+1.40}}$ \\
\cline{2-12}
 & $\vcenter{\hbox{\closlong}}$ & $\vcenter{\hbox{\shortstack{\pdandrem\\(40\,cm)}}}$ & $\vcenter{\hbox{\ifelsepdrem}}$ & $\vcenter{\hbox{\shortstack{+0.04\\(+0.09\%)}}}$ & $\vcenter{\hbox{\shortstack{-79\\(-8.38\%)}}}$ & $\vcenter{\hbox{\ifelsepdrem}}$ & $\vcenter{\hbox{\shortstack{+0.04\\(+0.09\%)}}}$ & $\vcenter{\hbox{\shortstack{-3291\\(-51.28\%)}}}$ & $\vcenter{\hbox{\pdandrem}}$ & $\vcenter{\hbox{-}}$ & $\vcenter{\hbox{-}}$ \\
\cline{1-12}
\multirow[c]{2}{*}{$\vcenter{\hbox{\chfive}}$} & $\vcenter{\hbox{\clos}}$ & $\vcenter{\hbox{\shortstack{\clandxgb\\(42\,cm)}}}$ & $\vcenter{\hbox{\ifelsepdrem}}$ & $\vcenter{\hbox{\shortstack{+12\\(+28.32\%)}}}$ & $\vcenter{\hbox{\shortstack{-21\\(-7.60\%)}}}$ & $\vcenter{\hbox{\ifelsepdrem}}$ & $\vcenter{\hbox{\shortstack{+12\\(+28.32\%)}}}$ & $\vcenter{\hbox{\shortstack{-1192\\(-44.59\%)}}}$ & $\vcenter{\hbox{\rem}}$ & $\vcenter{\hbox{\shortstack{+1\\(+2.26\%)}}}$ & $\vcenter{\hbox{+4.05}}$ \\
\cline{2-12}
 & $\vcenter{\hbox{\closlong}}$ & $\vcenter{\hbox{\shortstack{\clandxgb\\(44\,cm)}}}$ & $\vcenter{\hbox{\clandxgb}}$ & $\vcenter{\hbox{-}}$ & $\vcenter{\hbox{-}}$ & $\vcenter{\hbox{\clandxgb}}$ & $\vcenter{\hbox{-}}$ & $\vcenter{\hbox{-}}$ & $\vcenter{\hbox{\pdandxgb}}$ & $\vcenter{\hbox{\shortstack{+0.3\\(+0.80\%)}}}$ & $\vcenter{\hbox{+1.63}}$ \\
\cline{1-12}
\end{tabularx}
\renewcommand{\arraystretch}{1.00} 
\end{table}


%% file: content/conclusions.tex
\section{Conclusion and Future Work}
\label{sec:conclusion}

UWB tags yield very accurate distance estimation, yet are negatively
affected by NLOS conditions. Among these, we focus on the occlusion
caused by the wearer's body (WNLOS) when the tag is worn on the
chest---a typical placement in many applications that, despite the
strong negative impact on UWB ranging, received little attention in
the literature. Moreover, we focus on the problem of reliably
detecting whether a ranging sample occurred in WNLOS, which
empowers applications with the ability to discard it or treat it
differently. 

Our study fills several gaps in the related literature. First,
existing works, for NLOS in general and WNLOS alike, are dominated by
machine learning approaches. Instead, along these, we consider also
classifiers based on the link indicators available on the UWB
radio, which have characteristics complementary to ML
classifiers. Moreover, we exploit the synergic combination of these
two types of classifiers via several novel hybrid solutions we
designed exploiting alternative strategies. Second, the few works in
the WNLOS literature characterize the impact of signal attenuation and
evaluate classification techniques based on a single human subject, a
single environment, a specific radio configuration, and few occlusion
configurations without accurate ground truth. In contrast, our
experiments encompass 3~channels, 3~radio optimizations,
2~environments, and 5~human subjects, over a dataset of $1.5+$~million
ranging samples and corresponding CIRs. Finally, with the intent of
providing results of practical relevance, we focus on techniques
supporting classification on-device, a key requirement in
human-centric applications, and consequently evaluate also their
runtime and energy consumption. Moreover, in addition to
classification accuracy we also evaluate the ranging error of samples
classified as LOS, accounting for the practical impact of
misclassifications.

Our comparison of a total of $21$ classifiers reveals that they
\emph{all} achieve good accuracy, on average in the range
\perval{85.8\text{--}93.1}, yet with very different trade-offs when
ranging errors, energy consumption, and runtime are considered, and
when different radio configurations are used. We studied in detail the
resulting trade-offs along many dimensions, and distilled several
findings. Moreover, to facilitate their analysis and practical
application, we created \rectool, a recommender tool selecting the
best-performing classifier based on given performance requirements.

More work is required to investigate in further depth the
peculiarities of WNLOS. For instance, although we experimented with
5~subjects with different characteristics rather than a single one as
in the literature, we could not identify clear trade-offs related to
their body structure. Yet, a study involving a much larger cohort may
uncover subject-dependent effects; similar arguments hold concerning
alternative wearing positions, \eg on the wrist or inside a
pocket. Moreover, although we experimented with different and
representative environments and experimental setups, radically
different ones may elicit different tradeoffs, and similarly the use
of other radio configurations or entirely different UWB
transceivers. Finally, while we focused on discerning LOS \vs WNLOS
conditions, multi-label classification may elicit new insights.
Analyzing these dimensions would deserve dedicated studies, with a
multiplicative effect on the experimental effort, which was already
remarkable in our study. At the same time, we maintain that our study
not only constitutes a significant leap in knowledge and empirical
evidence \wrt the state of the art, but can also serve to
\emph{inform} these follow-up studies, providing a methodological and
technical foundation for experimentation and analysis. Consequently,
we openly release both our massive dataset and the \libra tool, to
benefit researchers and practitioners alike in follow-up endeavors.

%% file: content/acks.tex
\section*{Acknowledgments}
\label{sec:acknowledgment}

This work was partially supported by the European Union---Next
Generation EU under the Italian National Recovery and Resilience Plan
(NRRP), Mission 4, Component 2, Investment 1.3, CUP~B53C22003970001,
partnership on ``Telecommunications of the Future''
(PE00000001---program ``RESTART'') and, under the same NRRP program,
by the ICSC National Research Centre for High Performance Computing,
Big Data and Quantum Computing (CN00000013).  This work was also
partially executed within the SPiDR (``Secure, Performant, Dependable,
and Resilient Wireless Mesh Networks'') and SPIDR$^2$ (``Secure,
Performant, Intelligent, Dependable, Reliable, and Resilient Wireless
Systems'') projects financed by the Technology Innovation Institute.

\section*{AI disclosure statement}
Substantial portions of \libra source code, including trainer,
evaluator, and advisor modules, analysis utilities, and plotting
scripts, were refactored to enhance ease of use with assistance from
Google Gemini~3.1~Pro and Gemini~3.5~Flash within the Google
Antigravity IDE.  The authors reviewed, integrated, tested, and
modified the generated code, and remain responsible for the artifacts
and their conclusions.  The authors have also used ChatGPT for
linguistic refinement and grammar correction. All intellectual content
and final revisions were managed entirely by the authors.

\section*{CRediT author statement.} 
\emph{M. Gallacher}: conceptualization, methodology, software, validation, formal analysis, investigation, data curation, writing (original draft), visualization; 
\emph{V.A. Minh Le}: conceptualization, methodology, software, 
validation, formal analysis, investigation, data curation, writing (original draft), visualization; 
\emph{D. Vecchia}: conceptualization, methodology, 
formal analysis, visualization, writing (original draft); 
\emph{M. Trobinger}: conceptualization, methodology, 
formal analysis, visualization, writing (original draft); 
\emph{C.A. Boano}: conceptualization, methodology, writing (reviewing \& editing), supervision, project administration, funding acquisition. 
\emph{G.P. Picco}: conceptualization, methodology, writing (reviewing \& editing), supervision, project administration, funding acquisition.


%% file: references_long.bib
@ARTICLE{geng2013modeling,
  title = {{Modeling the Effect of Human Body on TOA Based Indoor Human Tracking}},
  author = {Geng, Yishuang and He, Jie and Pahlavan, Kaveh},
  journal = {Int. J. of Wireless Information Networks},
  volume = {20},
  number = {4},
  pages = {306--317},
  year = {2013},
  month = sep,
  publisher = {Springer}, 
  doi = {10.1007/s10776-013-0227-3},
}

@MISC{marketandmarkets,
  author = {{Markets \& Markets Research Private Ltd.}},
  title = {{Ultra-Wideband Market Size and Share}},
  year = {2025},
  month = oct,
  url = {https://www.marketsandmarkets.com/Market-Reports/ultra-wideband-market-200905786.html},
  note = {Last accessed: 2026-08-15},
}

@ARTICLE{breiman2001RandomForests,
  title = {Random {{Forests}}},
  author = {Breiman, Leo},
  date = {2001-10-01},
  journaltitle = {Machine Learning},
  shortjournal = {Mach. Learn.},
  volume = {45},
  number = {1},
  pages = {5--32},
  issn = {1573-0565},
  doi = {10.1023/A:1010933404324},
  url = {https://doi.org/10.1023/A:1010933404324},
}

@MISC{UWBAlliance_Applications,
  author = {{UWB Alliance}},		
  title = {{Applications}},
  url = {https://uwballiance.org/applications/},
  note = {Last accessed: 2026-08-15},
  year = {2026},
}

@MISC{FiRaConsortium_UseCases,
  author = {{FiRa Consortium}},		
  title = {{UWB Use Cases}},
  url = {https://www.firaconsortium.org/discover/use-cases},
  note = {Last accessed: 2026-08-15},  
  year = {2026},
}

@MISC{nordic-ppk2,
  author = {{Nordic Semiconductor}},		
  title = {{Power Profiler Kit II}},
  url = {https://www.nordicsemi.com/Products/Development-hardware/Power-Profiler-Kit-2},
  note = {Last accessed: 2026-08-15},
  year = {2026},
}

@MISC{dw1000-nlos-2,
  author = {{DecaWave Ltd.}},		
  title = {{APS006 Part 2 Application Note, v1.6: Non Line of Sight Operation and Optimizations to Improve Performance in DW1000 Based Systems}},
  year = 2024,
  month = may,
}

@MISC{dw1000:metrics,
  author = {{DecaWave Ltd.}},		
  title = {{APS006 Part 3 Application Note, v1.2: DW1000 Metrics for Estimation of Non Line Of Sight Operating Conditions}},
  year = 2024,
  month = may,
}

@MISC{dw1000:aps011,
  author = {{DecaWave Ltd.}},		
  title = {{APS011 Application Note, v1.2: Sources of Error in DW1000-based Two-Way Ranging (TWR) Schemes}},
  year = 2024,
  month = may,
}

@MISC{dw1000-datasheet,
  author = {{DecaWave Ltd.}},		
  title = {{DW1000 Datasheet, v2.22}},
  year = 2020,
  month = may,
}

@MISC{dw1000-manual,
  author = {{DecaWave Ltd.}},		
  title = {{DW1000 User Manual, v2.18}},
  year = 2017,
  month = apr,
}

@MISC{dwm1001-datasheet,
  author = {{DecaWave Ltd.}},		
  title = {{DWM1001C Datasheet, v1.7}},
  year = 2021,
  month = nov,
}

@MISC{dw3000-datasheet,
  author = {{DecaWave Ltd.}},		
  title = {{DW3000 Datasheet, v1.3}},
  year = 2022,
  month = sep,
}

@MISC{qorvo_mdek1001,
  author = {{Qorvo}},
  title = {{MDEK1001 Ultra-Wideband (UWB) Transceiver Development Kit}},
  url = {https://www.qorvo.com/products/p/MDEK1001},
  note = {Last accessed: 2026-08-15}, 
  year = {2026},
}

@MISC{optitrack,
  author = {{NaturalPoint, Inc.}},
  title = {{OptiTrack Documentation}},
  url = {https://docs.optitrack.com/},
  note = {Last accessed: 2026-08-15}, 
  year = {2026},
}

@MISC{motive,
  author = {{NaturalPoint, Inc.}},
  title = {{Motive Documentation}},
  url = {https://docs.optitrack.com/motive/},
  note = {Last accessed: 2026-08-15}, 
  year = {2026},
}

@MISC{rtt,
  author = {{SEGGER Microcontroller GmbH}},
  title = {{Real Time Transfer (RTT)}},
  url = {https://www.segger.com/products/debug-probes/j-link/technology/about-real-time-transfer/},
  note = {Last accessed: 2026-08-15}, 
  year = {2026},
}

@ARTICLE{wang23survey,
  author = {Wang, Fang and Tang, Hai and Chen, Jialei},
  title = {{Survey on NLOS Identification and Error Mitigation for UWB Indoor Positioning}},
  journal = {Electronics},
  volume = {12},
  number = {7},
  pages = {},
  year = {2023},
  month = apr,
  publisher = {MDPI},
  doi = {10.3390/electronics12071678},
}

@INPROCEEDINGS{kalyanaraman2020keyless,
  author = {Kalyanaraman, Avinash and Zeng, Yunze and Rakshit, Sushanta and Jain, Vivek},
  title = {{CaraoKey: Car States Sensing via the Ultra-Wideband Keyless Infrastructure}}, 
  booktitle = {Proc. of 17th Int. Conf. on Sensing, Communication, and Networking ({SECON})},
  location = {Como, Italy},
  pages = {1--9},
  year = {2020}, 
  month = jun,
  publisher = {IEEE},
  doi = {10.1109/SECON48991.2020.9158440},
}

@INPROCEEDINGS{2020_biri_socitrack,
  author = {Biri, Andreas and Jackson, Neal and Thiele, Lothar and Pannuto, Pat and Dutta, Prabal},
  title = {{SociTrack: Infrastructure-Free Interaction Tracking through Mobile Sensor Networks}},
  booktitle = {Proc. of 26th Int. Conf. on Mobile Computing and Networking ({MobiCom})},
  location = {Virtual event},
  pages = {1--14},
  year = {2020},
  month = apr,
  publisher = {ACM},
  doi = {https://doi.org/10.1145/3372224.3419190},
}

@INPROCEEDINGS{2016_ledergerber_robot,
  author = {Anton Ledergerber and Michael Hamer and Raffaello~D'Andrea},
  title = {{A Robot Self-Localization System using One-Way Ultra-Wideband Communication}},
  booktitle = {Proc. of Int. Conf. on Intelligent Robots and Systems~({IROS})},
  location = {Hamburg, Germany},
  pages = {3131--3137},
  year = 2015, 
  month = sep,
  publisher = {IEEE/RSJ},
  doi = {10.1109/IROS.2015.7353810},
}

@INPROCEEDINGS{yu2021robot,
  author = {Xianjia, Yu and Qingqing, Li and Queralta, Jorge Peña and Heikkonen, Jukka and Westerlund, Tomi},
  title = {{Cooperative UWB-Based Localization for Outdoors Positioning and Navigation of UAVs aided by Ground Robots}},
  booktitle = {Proc. of Int. Conf. on Autonomous Systems ({ICAS})},
  location = {Montreal, QC, Canada},
  pages = {1--5},
  year = {2021}, 
  month = aug,
  publisher = {IEEE},
  doi = {10.1109/ICAS49788.2021.9551177},
}

@INPROCEEDINGS{iya2011ObstructedPropogation,
  author = {Iya, Nuruddeen and Muqaibel, Ali and Johar, Umar and Landolsi, Mohamed Adnan},
  title = {{Ultra-Wideband Characterization of Obstructed Propagation}}, 
  booktitle = {Proc. of 7th Int. Wireless Communications and Mobile Computing Conf. (IWCMC)},     
  location = {Istanbul, Turkey},
  pages = {624--629},
  year = {2011},
  month = jul, 
  publisher = {IEEE},
  doi = {10.1109/IWCMC.2011.5982618},
}

@ARTICLE{marano2010NLOSIdentificationMit,
  author = {Stefano Maran{\`o} and Wesley M. Gifford and Henk Wymeersch and Moe Z. Win},
  title = {{NLOS Identification and Mitigation for Localization Based on UWB Experimental Data}},
  journal = {IEEE J. on Selected Areas in Communications},
  volume = {28},
  number = {7},
  pages = {1026--1035},
  year = {2010},
  month = sep,
  publisher = {IEEE},
  doi = {10.1109/JSAC.2010.100907},
}

@INPROCEEDINGS{stocker2021PerformanceofSupport,
  author = {Michael Stocker and Markus Gallacher and Carlo Alberto Boano and Kay R\"{o}mer},
  title = {{Performance of Support Vector Regression in Correcting UWB Ranging Measurements under LOS/NLOS Conditions}},
  _booktitle = {Proc. of 4th Int. Wkshp. on Benchmarking Cyber-Physical Systems and Internet of Things ({CPS-IoTBench})},  
  booktitle = {Proc. of 4th {CPS-IoTBench} Wkshp.},  
  location = {Virtual event},   
  pages = {6--11},	
  year = {2021},
  month = may,	
  publisher = {IEEE},
  doi = {10.1145/3458473.3458820},	
}

@INPROCEEDINGS{gallacher2023InSight,
  author = {Markus Gallacher and Michael Stocker and Michael Baddeley and Kay R\"{o}mer and Carlo Alberto Boano},
  title = {{InSight: Enabling NLOS Classification, Error Correction, and Anchor Selection on Resource-Constrained UWB Devices}},
  booktitle = {Proc. of 20th Int. Conf. on Embedded Wireless Systems and Networks ({EWSN})},  
  location = {Rende, Italy}, 
  pages = {132--144},	  
  year = {2023},
  month = sep,	
  publisher = {ACM},
  doi = {10.5555/3639940.3639959},	
}

@ARTICLE{Tian_2019_HumanBodyShadowing,
  author = {Tian, Qinglin and Wang, Kevin I-Kai and Salcic, Zoran},
  title = {{Human Body Shadowing Effect on UWB-Based Ranging System for Pedestrian Tracking}}, 
  journal = {IEEE Trans. on Instrumentation and Measurement (TIM)}, 
  volume = {68},
  number = {10},
  pages = {4028--4037},
  year = {2019},
  month = oct,
  publisher = {IEEE},  
  doi={10.1109/TIM.2018.2884605},
}

@INPROCEEDINGS{le22hnlos,
  author = {Anh Minh Le, Vu and Trobinger, Matteo and Vecchia, Davide and Picco, Gian Pietro},
  title = {{Human Occlusion in Ultra-wideband Ranging: What Can the Radio Do for You?}},
  booktitle = {Proc. of 18th Int. Conf. on Mobility, Sensing and Networking (MSN)},
  location = {Guangzhou, China},
  pages = {1--10},
  year = {2022},
  month = dec,
  publisher = {IEEE},
  doi = {10.1109/MSN57253.2022.00016},
}

@INPROCEEDINGS{Rodrigues_2023_WearableSmartSensing,
  author = {Rodrigues, Mariana Jacob and Postolache, Octavian and Cercas, Francisco},
  title = {{Wearable Smart Sensing and UWB System for Fall Detection in AAL Environments}}, 
  booktitle = {Proc. of 18th Sensors Applications Symp. (SAS)}, 
  location = {Ottawa, ON, Canada},
  pages = {1--6},
  year = {2023},
  month = jul, 
  publisher = {IEEE},
  doi = {10.1109/SAS58821.2023.10254065},
}

@ARTICLE{leoni22covid,
  author = {Leoni, Elia and Cencetti, Giulia and Santin, Gabriele and Istomin, Timofei and Molteni, Davide and Picco, Gian Pietro and Farella, Elisabetta and Lepri, Bruno and Murphy, Amy L.},
  title = {{Measuring Close Proximity Interactions in Summer Camps during the COVID-19 Pandemic}},
  journal = {EPJ Data Science},
  volume = {11},
  number = {5},
  pages = {},
  year = {2022},
  month = jan,
  publisher = {Springer},
  doi = {10.1140/epjds/s13688-022-00316-y},
}

@ARTICLE{Hulka_02092020,
  author = {Hulka, Karel and Strniste, Matej and Prycl, David},
  title = {{Accuracy and Reliability of Sage Analytics Tracking System based on UWB Technology for Indoor Team Sports}},
  journal = {Int. J. of Performance Analysis in Sport},
  volume = {20},
  number = {5},
  pages = {800--807},
  year = {2020},
  month = jul,
  publisher = {Routledge},
  doi = {10.1080/24748668.2020.1788349},
}

@ARTICLE{hachem25tosn,
  author = {Hachem, Fatima and Vecchia, Davide and Damiani, Maria Luisa and Picco, Gian Pietro},
  title = {{Fine-Grained Stop-Move Detection with UWB: Quality Metrics and Real-World Evaluation}},
  journal = {ACM Trans. on Sensor Networks (TOSN)},  
  volume = {21},
  number = {4},
  pages = {1--33},
  year = {2025},
  month = jul, 
  publisher = {ACM},
  doi = {10.1145/3735558},
}

@INPROCEEDINGS{makellaraki25material,
  author = {Makellaraki, Angeliki and Di Pietra, Vincenzo and Dabove, Paolo},
  title = {{Assessing Material Impacts in NLOS UWB Ranging Errors: Characterization for Museum Environments}}, 
  booktitle = {Proc. of Position, Location and Navigation Symp. (PLANS)},
  location = {Salt Lake City, UT, USA},
  pages = {1297--1306},
  year = {2025},
  month = apr,
  publisher = {IEEE},
  doi={10.1109/PLANS61210.2025.11028487},
}

@INPROCEEDINGS{cao20206FitAPartProtocolPhysical,
  author = {Cao, Yifeng and Dhekne, Ashutosh and Ammar, Mostafa},
  title = {{6Fit-A-Part: A Protocol for Physical Distancing on a Custom Wearable Device}}, 
  booktitle = {Proc. of 28th Int. Conf. on Network Protocols (ICNP)}, 
  location = {Madrid, Spain},
  pages = {1--12},
  year = {2020},
  month = oct,
  publisher = {IEEE},
  doi={10.1109/ICNP49622.2020.9259374},
}

@ARTICLE{bregar2021AssistedLiving,
    author = {Klemen Bregar and Andrej Hrovat and Mihael Mohor\v{c}i\v{c}},
    title = {{UWB Radio-Based Motion Detection System for Assisted Living}},
    journal = {Sensors},
    volume = {21},
    number = {11},
    pages = {},
    year = {2021},
	month = may,
	publisher = {MDPI},
    doi={10.3390/s21113631}
}

@INPROCEEDINGS{locklin20geofencing,
  author = {Löcklin, Andreas and Ruppert, Tamás and Jakab, László and Libert, Robert and Jazdi, Nasser and Weyrich, Michael},
  title = {{Trajectory Prediction of Humans in Factories and Warehouses with Real-Time Locating Systems}}, 
  booktitle = {Proc. of 25th Int. Conf. on Emerging Technologies and Factory Automation (ETFA)}, 
  location = {Vienna, Austria},
  pages = {1317--1320},
  year = {2020},
  month = sep,
  publisher = {IEEE},
  doi={10.1109/ETFA46521.2020.9211913},
}

@MASTERSTHESIS{Liukkonen25geofencing,
  author = {Liukkonen, Aino}, 
  title = {{Geofencing for Wireless Human-Machine Interfaces in Packaging Machinery: From Analysis to Implementation}},
  school = {Lappeenranta–Lahti University of Technology, Finland},
  year = {2025},
}

@INPROCEEDINGS{contigiani2016SmartRetail,
  author = {Contigiani, Marco and Pietrini, Rocco and Mancini, Adriano and Zingaretti, Primo},
  title = {{Implementation of a Tracking System based on UWB Technology in a Retail Environment}}, 
  booktitle = {Proc. of 12th Int. Conf. on Mechatronic and Embedded Systems and Applications (MESA)}, 
  location = {Auckland, New Zealand},
  pages = {1--6},
  year = {2016},
  month = aug,
  publisher = {IEEE},
  doi={10.1109/MESA.2016.7587123},
}

@ARTICLE{yang25tennis,
    author = {Yang, Wenpu and Wang, Jinzheng and Zhao, Zichen and Cui, Yixiong},
    title = {{Accuracy of an Ultra-Wideband-Based Tracking System for Time–Motion Analysis in Tennis}},
    journal = {Sensors},
    volume = {25},
    number = {4},
    pages = {},
    year = {2025},
	month = feb,
	publisher = {MDPI},
    doi={10.3390/s25041031}
}

@ARTICLE{Ferreira_2021_FeatureSelection,
  author = {Ferreira, André G. and Fernandes, Duarte and Branco, Sérgio and Catarino, André Paulo and Monteiro, João L.},
  title = {{Feature Selection for Real-Time NLOS Identification and Mitigation for Body-Mounted UWB Transceivers}}, 
  journal = {IEEE Trans. on Instrumentation and Measurement}, 
  volume={70},
  number={},
  pages={},
  year={2021},
  month = apr,
  publisher = {IEEE},
  doi={10.1109/TIM.2021.3070619}
}

@ARTICLE{Otim_2019_ImpactofBodyWearable,
  author = {Otim, Timothy and Bahillo, Alfonso and Díez, Luis Enrique and Lopez-Iturri, Peio and Falcone, Francisco},
  title = {{Impact of Body Wearable Sensor Positions on UWB Ranging}}, 
  journal = {IEEE Sensors J.}, 
  volume = {19},
  number = {23},
  pages = {11449--11457},
  year = {2019},
  month = dec,
  publisher = {IEEE},  
  doi={10.1109/JSEN.2019.2935634}
}

@ARTICLE{Kim_2023_UWBPositioning,
  author = {Kim, Dae-Ho and Farhad, Arshad and Pyun, Jae-Young},
  title = {{UWB Positioning System Based on LSTM Classification With Mitigated NLOS Effects}}, 
  journal = {IEEE Internet of Things J.}, 
  volume = {10},
  number = {2},
  pages = {1822--1835},
  year = {2023},
  month = jan,
  publisher = {IEEE},  
  doi={10.1109/JIOT.2022.3209735}
}

@ARTICLE{Mirama_2023_NLOSDetection,
  author = {Miramá, Víctor and Bahillo, Alfonso and Quintero, Víctor and Díez, Luis Enrique},  
  title = {{NLOS Detection Generated by Body Shadowing in a 6.5 GHz UWB Localization System Using Machine Learning}}, 
  journal = {IEEE Sensors J.}, 
  volume = {23},
  number = {17},
  pages = {20400--20411},
  year = {2023},
  month = sep,
  publisher = {IEEE},
  doi = {10.1109/JSEN.2023.3298843}
}

@ARTICLE{Han_2025_RobustUWBPositioning,
  author = {Han, Miseon and Kim, Jeongtae},
  title = {{Robust UWB Positioning System using Switching Tracking Filters with Estimated States-based Non-Line-of-Sight Classifier}}, 
  journal = {IEEE Access}, 
  volume = {13},
  number = {},
  pages = {71101--71119},
  year = {2025},
  month = apr,
  publisher = {IEEE},
  doi = {10.1109/ACCESS.2025.3561842}
}

@ARTICLE{Wei_2022_NLOSIdentification,
  author = {Junyu Wei and Haowen Wang and Shaojing Su and Ying Tang and Xiaojun Guo and Xiaoyong Sun},
  title = {{NLOS Identification using Parallel Deep Learning Model and Time-frequency Information in UWB-based Positioning System}},
  journal = {Measurement},
  volume = {195},
  number = {},
  pages = {},
  year = {2022},
  month = may,
  publisher = {Elsevier}, 
  doi = {https://doi.org/10.1016/j.measurement.2022.111191},
}

@ARTICLE{chenC_10422747,
  author = {Chen, Changwei and Kia, Solmaz S.},
  title = {{OptiTrack-Aided Supervised Learning for Neural Network-Based Ultra-Wideband Ranging Bias Correction}}, 
  journal = {IEEE Sensors J.}, 
  volume = {24},
  number = {6},
  pages = {8484--8492},
  year = {2024},
  month = mar,
  publisher = {IEEE},
  doi = {10.1109/JSEN.2024.3360187}
}

@INPROCEEDINGS{Ember_ImprovingTheCorrection_10767096,
  author = {Ember, Elisei and Pestana, Jesus and Krisper, Michael and Stocker, Michael and Römer, Kay and Boano, Carlo Alberto and Pelegrín, Pablo Corbalán},
  title = {{Improving the Correction of NLoS-Induced Ranging Errors in UWB Systems through Enhanced Labeling}}, 
  booktitle = {Proc. of 15th Wireless and Mobile Networking Conf. (WMNC)}, 
  location = {Venice, Italy},
  pages = {100--107},
  year = {2024},
  month = nov,
  publisher = {IEEE},
  doi = {10.52545/3-16}
}

@INPROCEEDINGS{ember23feature,
  author = {Ember, Elisei and Pestana, Jesus and Krisper, Michael and Stocker, Michael and Römer, Kay and Boano, Carlo Alberto and Pelegrín, Pablo Corbalán},  
  title = {{Impact of Feature Selection and CIR Window Length on NLoS Classification for UWB Systems}}, 
  booktitle = {Proc. of 19th Int. Conf. on Mobility, Sensing and Networking (MSN)}, 
  location = {Nanjing, China},
  pages = {72--80},
  year = {2023},
  month = dec, 
  publisher = {IEEE},
  doi={10.1109/MSN60784.2023.00024},
}

@INPROCEEDINGS{Albaidhani16nlos,
  author = {Albaidhani, Abbas and Morell, Antoni and Vicario, Jose Lopez},  
  title = {{Ranging in UWB Using Commercial Radio Modules: Experimental Validation and NLOS Mitigation}}, 
  booktitle = {Proc. of Int. Conf. on Indoor Positioning and Indoor Navigation (IPIN)}, 
  location = {Alcala de Henares, Spain},
  pages = {1--7},
  year={2016},
  month = oct,
  publisher = {IEEE},
  doi = {10.1109/IPIN.2016.7743639},
}

@ARTICLE{bregar2018CNN,
  author = {Klemen Bregar and Mihael Mohor\v{c}i\v{c}},
  title = {{Improving Indoor Localization Using Convolutional Neural Networks on Computationally Restricted Devices}},
  journal = {IEEE Access},
  volume = {6},
  number = {},
  pages = {17429--17441},
  year = {2018},
  month = mar,
  publisher = {IEEE},
  doi = {10.1109/ACCESS.2018.2817800},
}

@ARTICLE{jiang_9108193,
  author = {Jiang, Changhui and Shen, Jichun and Chen, Shuai and Chen, Yuwei and Liu, Di and Bo, Yuming},
  title = {{UWB NLOS/LOS Classification Using Deep Learning Method}}, 
  journal = {IEEE Communications Letters}, 
  volume = {24},
  number = {10},
  pages = {2226--2230},
  year = {2020},
  month = oct, 
  publisher = {IEEE},
  doi={10.1109/LCOMM.2020.2999904}, 
}

@ARTICLE{angarano2021REMNet,
	author = {Simone Angarano and Vittorio Mazzia and Francesco Salvetti and Giovanni Fantin and Marcello Chiaberge},
	title = {{Robust Ultra-Wideband Range Error Mitigation with Deep Learning at the Edge}},
	journal = {Engineering Applications of Artificial Intelligence},
	volume = {102},
	number = {},
    pages = {},
	year = {2021},
	month = jun,
	publisher = {Elsevier},
	doi = {10.1016/j.engappai.2021.104278},
}

@ARTICLE{Hwang_2025_UltrawidebandNonLineofSightClassification,
  author = {Hwang, Hae-Ji and Jeong, Seon-Geun and Hwang, won-Joo},
  title = {{Ultrawideband Non-Line-of-Sight Classification Using Transformer-Convolutional Neural Networks}},
  journal = {IEEE Access},
  volume = {13},
  number = {},
  pages = {86323--86332},
  year = {2025},
  month = may,
  publisher = {IEEE},
  doi = {10.1109/ACCESS.2025.3568830},
}

@ARTICLE{Deng_2023_UWBNLOSIdentificationAndMitigation,
  author = {Deng, Bowen and Xu, Tangwen and Yan, Maode},
  title = {{UWB NLOS Identification and Mitigation Based on Gramian Angular Field and Parallel Deep Learning Model}}, 
  journal = {IEEE Sensors J.}, 
  volume = {23},
  number = {22},
  pages = {28513--28525},
  year = {2023},
  month = oct,
  publisher = {IEEE},
  doi = {10.1109/JSEN.2023.3323564}, 
}

@INPROCEEDINGS{malajner2015UWBRangingAccuracy,
  author = {Malajner, Marko and Planinšič, Peter and Gleich, Dušan},
  title = {{UWB Ranging Accuracy}}, 
  booktitle = {Proc. of Int. Conf. on Systems, Signals and Image Processing (IWSSIP)}, 
  location = {London, UK},
  pages = {61--64},
  year = {2015},
  month = sep,
  publisher = {IEEE},
  doi = {10.1109/IWSSIP.2015.7314177},
}

@ARTICLE{Bharadwaj_2022_ExperimentalInvestigation,
  author = {Bharadwaj, Richa and Alomainy, Akram and Koul, Shiban K.},
  title = {{Experimental Investigation of Body-Centric Indoor Localization Using Compact Wearable Antennas and Machine Learning Algorithms}}, 
  journal = {IEEE Trans. on Antennas and Propagation}, 
  volume = {70},
  number = {2},
  pages = {1344--1354},
  year = {2022},
  month = feb,
  publisher = {IEEE},
  doi = {10.1109/TAP.2021.3111308},
}

@ARTICLE{yogesh2024UWBDistanceEstimationa,  
  author = {Yogesh, Vinish and Rook, Jan Willem A. and Keizers, Thomas and Voort, Carsten and Buurke, Jaap H. and Veltink, Peter H. and Baten, Chris T.M.},
  title = {{UWB Distance Estimation Errors in (Non-)Line of Sight Situations within the Context of 3D Analysis of Human Movement}},
  journal = {Engineering Research Express},
  volume = {6},
  number = {4},
  pages = {},
  year = {2024},
  month = oct,
  publisher = {IOP Publishing},
  doi = {10.1088/2631-8695/ad7e7e},
}

@ARTICLE{Yang_2025_FuzzyTransformer,
  author = {Yang, Hongchao and Wang, Yunjia and Seow, Chee Kiat and Sun, Meng and Coene, Sander and Huang, Lu and Joseph, Wout and Plets, David},
  title = {{Fuzzy Transformer Machine Learning for UWB NLOS Identification and Ranging Mitigation}}, 
  journal = {IEEE Trans. on Instrumentation and Measurement}, 
  volume = {74},
  number={},
  pages = {1--17},
  year = {2025},
  month = mar,
  publisher = {IEEE},
  doi = {10.1109/TIM.2025.3548180}, 
}

@ARTICLE{sung23position,
  author = {Sung, Sangmo and Kim, Hokeun and Jung, Jae-Il},
  title = {{Accurate Indoor Positioning for UWB-Based Personal Devices Using Deep Learning}}, 
  journal = {IEEE Access}, 
  volume = {11},
  number = {},
  pages = {20095--20113},
  year = {2023},
  month = feb, 
  publisher = {IEEE},
  doi={10.1109/ACCESS.2023.3250180},
}

@INPROCEEDINGS{gururaj2017RssFpplDetection,
  author = {Gururaj, Karthikeyan and Rajendra, Anojh Kumaran and Song, Yang and Law, Choi Look and Cai, Guofa},
  title = {{Real-time Identification of NLOS Range Measurements for Enhanced UWB Localization}}, 
  booktitle = {Proc. of Int. Conf. on Indoor Positioning and Indoor Navigation (IPIN)}, 
  location = {Sapporo, Japan},
  pages={1--7},
  year={2017},
  month = sep,
  publisher = {IEEE},
  doi={10.1109/IPIN.2017.8115877},
}

@ARTICLE{chenY_10.1145/3657639,
  author = {Chen, Yijie and Wang, Jiliang and Yang, Jing},
  title = {{Exploiting Anchor Links for NLOS Combating in UWB Localization}},
  journal = {ACM Trans. on Sensor Networks (TOSN)},
  volume = {20},
  number = {3},
  pages = {},
  year = {2024},
  month = may,
  publisher = {ACM},
  doi = {10.1145/3657639},
}

@INPROCEEDINGS{heydariaan18nlosbenchmark,
  author = {Heydariaan, Milad and Mohammadmoradi, Hessam and Gnawali, Omprakash},
  title = {{Toward Standard Non-Line-of-Sight Benchmarking of Ultra-Wideband Radio-Based Localization}},
  booktitle = {Proc. of 1st Int. Wkshp. on Benchmarking Cyber-Physical Networks and Systems (CPSBench)},
  location = {Porto, Portugal}, 
  pages = {19--24},	
  year = {2018},
  month = apr,	
  publisher = {IEEE},
  doi = {10.1109/CPSBench.2018.00010},	
}

@INPROCEEDINGS{stocker23nlos,
  author = {Michael Stocker and Markus Gallacher and Carlo Alberto Boano and Kay R\"{o}mer},
  title = {{Applying NLOS Classification and Error Correction Techniques to UWB Systems: Lessons Learned and Recommendations}},
  _booktitle = {Proc. of 6th Int. Wkshp. on Benchmarking Cyber-Physical Systems and Internet of Things ({CPS-IoTBench})},
  booktitle = {Proc. of 6th {CPS-IoTBench} Wkshp.},  
  location = {San Antonio, TX, USA}, 
  pages = {78-–83},	
  year = {2023},
  month = may,	
  publisher = {ACM},
  doi = {10.1145/3576914.3587522},	
}

@INPROCEEDINGS{Shalihan_NLOSRanging_2022_9926650,
  author = {Shalihan, Muhammad and Liu, Ran and Yuen, Chau},
  title = {{NLOS Ranging Mitigation with Neural Network Model for UWB Localization}}, 
  booktitle = {Proc. of 18th Int. Conf. on Automation Science and Engineering (CASE)}, 
  location = {Mexico City, Mexico},
  pages = {1370--1376},
  year = {2022},
  month = aug, 
  publisher = {IEEE},
  doi = {10.1109/CASE49997.2022.9926650},
}

@ARTICLE{Li_2024_NLOSOcclusionRecognition,
  author = {Li, Wenfeng and Yang, Anning and Zhou, Jinglong and Zhu, Yulei},
  title = {{NLOS Occlusion Recognition Method to Improve UWB Spatial Sensing}}, 
  journal = {IEEE Internet of Things J.}, 
  volume = {11},
  number = {19},
  pages = {31715--31729},
  year = {2024},
  month = oct,
  publisher = {IEEE},
  doi = {10.1109/JIOT.2024.3419796}, 
}

@ARTICLE{arsuaga2024NovelClassificationMethod,
  author = {Arsuaga, Meritxell and {Ochoa-De-Eribe-Landaberea}, Aitor and {Zamora-Cadenas}, Leticia and Arrizabalaga, Saioa and Velez, Igone},
  title = {{Novel Classification Method to Predict the Accuracy of UWB Ranging Estimates}},
  journal = {IEEE Access},
  volume = {12},
  number = {},
  pages = {33659--33670},
  year = {2024},
  month = feb,
  publisher = {IEEE},
  doi = {10.1109/ACCESS.2024.3371907},
}

@ARTICLE{barralNLOSIdentificationMitigation2019,
  author = {Barral, Valent{\'i}n and Escudero, Carlos J. and {Garc{\'i}a-Naya}, Jos{\'e} A. and {Maneiro-Catoira}, Roberto},
  title = {{NLOS Identification and Mitigation Using Low-Cost UWB Devices}},
  journal = {Sensors},
  volume = {19},
  number = {16},
  pages = {},
  year = {2019},
  month = jan,
  publisher = {MDPI},
  doi = {10.3390/s19163464},
}

@INPROCEEDINGS{Zeng_2019_UWBNLOSIdentification,
  author = {Zeng, Zhuoqi and Liu, Steven and Wang, Lei},
  title = {{UWB NLOS Identification with Feature Combination Selection based on Genetic Algorithm}}, 
  booktitle = {Proc. of IEEE Int. Conf. on Consumer Electronics (ICCE)}, 
  location = {Las Vegas, NV, USA},
  pages = {1--5},
  year = {2019},
  month = jan,
  publisher = {IEEE},
  doi = {10.1109/ICCE.2019.8662065},
}

@INPROCEEDINGS{flueratoru_9662532,
  author = {Flueratoru, Laura and Lohan, Elena Simona and Niculescu, Dragoş},
  title = {{Self-Learning Detection and Mitigation of Non-Line-of-Sight Measurements in Ultra-Wideband Localization}}, 
  booktitle = {Proc. of Int. Conf. on Indoor Positioning and Indoor Navigation (IPIN)}, 
  location = {Lloret de Mar, Spain},
  pages = {1--8},
  year = {2021},
  month = nov,
  publisher = {IEEE},
  doi = {10.1109/IPIN51156.2021.9662532},
}

@ARTICLE{Pei03072024,
  author = {Yu, Pei and Ruizhi, Chen and Deren, Li and Xiongwu, Xiao and Xingyu, Zheng},
  title = {{FCN-Attention: A Deep Learning UWB NLOS/LOS Classification Algorithm Using Fully Convolution Neural Network with Self-attention Mechanism}},
  journal = {Geo-spatial Information Science},
  volume = {27},
  number = {4},
  pages = {1162--1181},
  year = {2023},
  month = mar,
  publisher = {Taylor \& Francis},
  doi = {10.1080/10095020.2023.2178334}, 
}

@INPROCEEDINGS{Jones23TinyButMighty,
  author = {Jones, Ben and Raza, Usman and Khan, Aftab},
  title = {{Tiny but Mighty: Embedded Machine Learning for Indoor Wireless Localization}}, 
  booktitle = {Proc. of 20th Consumer Communications \& Networking Conf. (CCNC)}, 
  location = {Las Vegas, NV, USA},
  pages = {176--181},
  year = {2023},
  month = jan,
  publisher = {IEEE},
  doi = {10.1109/CCNC51644.2023.10060797},
}

@INPROCEEDINGS{zeng_8631718,
  author = {Zeng, Zhuoqi and Liu, Steven and Wang, Lei},
  title = {{NLOS Identification for UWB Based on Channel Impulse Response}}, 
  booktitle = {Proc. of 12th Int. Conf. on Signal Processing and Communication Systems (ICSPCS)}, 
  location = {Cairns, QLD, Australia},
  pages = {1--6},
  year = {2018},
  month = dec,
  publisher = {IEEE},
  doi = {10.1109/ICSPCS.2018.8631718}, 
}

@ARTICLE{Sang_2020_IdentificationOfNLOS,
  author = {Sang, Cung Lian and Steinhagen, Bastian and Homburg, Jonas Dominik and Adams, Michael and Hesse, Marc and Rückert, Ulrich},
  title = {{Identification of NLOS and Multi-Path Conditions in UWB Localization Using Machine Learning Methods}},
  journal = {Applied Sciences},
  volume = {10},
  number = {11},
  pages = {},
  year = {2020},
  month = jun,
  publisher = {MDPI},
  doi = {https://doi.org/10.3390/app10113980},
}

@ARTICLE{VanNguyen15ML,
  author = {Van Nguyen, Thang and Jeong, Youngmin and Shin, Hyundong and Win, Moe Z.},
  title = {{Machine Learning for Wideband Localization}}, 
  journal = {IEEE J. on Selected Areas in Communications}, 
  volume = {33},
  number = {7},
  pages = {1357--1380},
  year = {2015},
  month = jul,
  publisher = {IEEE},
  doi={10.1109/JSAC.2015.2430191},
}

@ARTICLE{Wang_2025_ImprovedUWBBased,
  author = {Shoude Wang and Nur Syazreen Ahmad},
  title = {{Improved UWB-based Indoor Positioning System via NLOS Classification and Error Mitigation}},
  journal = {Engineering Science and Technology},
  volume = {63},
  number = {},
  pages = {},
  year = {2025},
  month = may,
  publisher = {Elsevier},
  doi = {10.1016/j.jestch.2025.101979},
}

@ARTICLE{WIERZBICKI1982391,
  author = {Andrzej P. Wierzbicki},
  title = {{A Mathematical Basis for Satisficing Decision Making}},
  journal = {Mathematical Modelling},
  volume = {3},
  number = {5},
  pages = {391--405},
  year = {1982},
  publisher = {Elsevier},
  doi = {10.1016/0270-0255(82)90038-0},
}

@ARTICLE{ALLMENDINGER2022105857,
  author = {Richard Allmendinger and Andrzej Jaszkiewicz and Arnaud Liefooghe and Christiane Tammer},
  title = {{What If We Increase the Number of Objectives? Theoretical and Empirical Implications for Many-objective Combinatorial Optimization}},
  journal = {Computers \& Operations Research},
  volume = {145},
  number = {},
  pages = {},
  year = {2022},
  month = sep,
  publisher = {Elsevier},
  doi = {10.1016/j.cor.2022.105857},
}

@ARTICLE{yang24nlos,
  author = {Yang, Hongchao and Wang, Yunjia and Xu, Shenglei and Bi, Jingxue and Jia, Haonan and Seow, Cheekiat},
  title = {{Ultra-Wideband Ranging Error Mitigation with Novel Channel Impulse Response Feature Parameters and Two-Step Non-Line-of-Sight Identification}},
  journal = {Sensors},
  volume = {24},
  number = {5},
  pages = {},
  year = {2024},
  month = mar,
  publisher = {MDPI},
  doi = {10.3390/s24051703},
}

@INPROCEEDINGS{stahlke20nlos,
  author = {Stahlke, Maximilian and Kram, Sebastian and Mutschler, Christopher and Mahr, Thomas},
  title = {{NLOS Detection using UWB Channel Impulse Responses and Convolutional Neural Networks}},
  booktitle = {Proc. of Int. Conf. on Localization and GNSS (ICL-GNSS)},
  location = {Tampere, Finland},
  pages = {1--6},
  year = {2020},
  month = jun,
  publisher = {IEEE},
  doi = {10.1109/ICL-GNSS49876.2020.9115498},
}

@INPROCEEDINGS{brunner22uwbwifi,
  author = {Hannah Brunner and Michael Stocker and Maximilian Schuh and Markus Schu{\ss} and Carlo Alberto Boano and Kay R\"{o}mer},
  title = {{Understanding and Mitigating the Impact of Wi-Fi 6E Interference on Ultra-Wideband Communications and Ranging}},
  booktitle = {Proc. of 21st Int. Conf. on Information Processing in Sensor Networks ({IPSN})},
  pages = {92--104},	
  location = {Virtual event}, 
  year = {2022},
  month = may,	
  publisher = {IEEE},
  doi = {10.1109/IPSN54338.2022.00015},	
}

@ARTICLE{Leitch_2023_OnIndoorLocalization,
  author = {Leitch, Samuel G. and Ahmed, Qasim Zeeshan and Abbas, Waqas Bin and Hafeez, Maryam and Laziridis, Pavlos I. and Sureephong, Pradorn and Alade, Temitope},
  title = {{On Indoor Localization Using WiFi, BLE, UWB, and IMU Technologies}},
  journal = {Sensors},
  volume = {23},
  number = {20},
  pages = {},
  year = {2023},
  month = oct,
  publisher = {MDPI},
  doi = {10.3390/s23208598},
}

@ARTICLE{Li_2024_HumanBodyImpact,
  author = {Li, Xueping and Gao, Yue and Yang, Yuan and Xiang, Wei and Jiang, Xiaolin and Yang, Yun and Zhang, Xiaoling},
  title = {{Human Body Impact and Error Correction for Indoor UWB Ranging}},
  journal = {IEEE Trans. on Consumer Electronics},
  volume = {70},
  number = {1},
  pages = {2480--2491},
  year = {2024},
  month = mar,
  publisher = {IEEE},
  doi = {10.1109/TCE.2024.3375513},
}

@INPROCEEDINGS{tianqi_xgboost,
  author = {Chen, Tianqi and Guestrin, Carlos},
  title = {{XGBoost: A Scalable Tree Boosting System}},
  booktitle = {Proc. of 22nd ACM SIGKDD Int. Conf. on Knowledge Discovery and Data Mining ({KDD})},
  pages = {785--794},	
  location = {San Francisco, CA, USA},
  year = {2016},
  month = aug, 
  publisher = {ACM},
  doi = {10.1145/2939672.2939785},
}

@INPROCEEDINGS{Kumpuniemi14bodysize,
  author = {Kumpuniemi, Timo and H\"{a}m\"{a}l\"{a}inen, Matti and Yazdandoost, Kamya Yekeh and Iinatti, Jari},
  title = {{Human Body Size and Shape Effect on UWB On-body WBAN Radio Channels -- Preliminary Results}},
  booktitle = {Proc. of 9th Int. Conf. on Body Area Networks ({BodyNets})},
  pages = {372--378},	
  location = {London, United Kingdom},
  year = {2014},
  month = sep, 
  publisher = {ICST},
  doi = {10.4108/icst.bodynets.2014.256996},
}

@ARTICLE{ridolfi_sport2018,
  author = {Matteo Ridolfi and Stef Vandermeeren and Jense Defraye and Heidi Steendam and Joeri Gerlo and Dirk De Clercq and Jeroen Hoebeke and Eli De Poorter},
  title = {{Experimental Evaluation of UWB Indoor Positioning for Sport Postures}},
  journal = {MDPI Sensors},
  volume = {18},
  number = {1},
  pages = {},
  year = {2018},
  month = jan,
  publisher = {MDPI},
  doi = {10.3390/s18010168},
}

@INPROCEEDINGS{yusoff09absorption,
  author = {Yusoff, N.I.M. and Khatun, S. and AlShehri, S.A.},
  title = {{Characterization of Absorption Loss for UWB Body Tissue Propagation Model}},
  booktitle = {Proc. of 9th Malaysia International Conference on Communications ({MICC})},
  pages = {254--258},	
  location = {Kuala Lumpur, Malaysia},
  year = {2009},
  month = dec, 
  publisher = {IEEE},
  doi = {10.1109/MICC.2009.5431507},
}

@ARTICLE{klemm06absorption,
  author = {Maciej Klemm and Gerhard Troester},
  _author = {M. Klemm and G. Troester},
  title = {{EM Energy Absorption in the Human Body Tissues Due to UWB Antennas}},
  journal = {Progress In Electromagnetics Research (PIER)},
  volume = {62},
  number = {},
  pages = {261--280},
  year = {2006},
  month = jul,
  publisher = {},
  doi = {10.2528/PIER06040601},
}

@ARTICLE{christ06tissue,
  _author = {A. Christ and T. Samaras and A, Klingenb\"{o}ck and N. Kuster},
  author = {Andreas Christ and Theodoros Samaras and Anja Klingenb\"{o}ck and Niels Kuster}, 
  title = {{Characterization of the Electromagnetic Near-field Absorption in Layered Biological Tissue in the Frequency Range from 30 MHz to 6000 MHz}},
  journal = {Physics in Medicine \& Biology},
  volume = {51},
  number = {19},
  pages = {},
  year = {2006},
  month = sep,
  publisher = {IOP Science},
  doi = {10.1088/0031-9155/51/19/014},
}

@inproceedings{schuh24complex,
  author = {Maximilian Schuh and Michael Baddeley and Kay R\"{o}mer and Carlo Alberto Boano},
  _author = {M. Schuh and M. Baddeley and K. R\"{o}mer and C.A. Boano},
  _author = {Maximilian Schuh and others},
  title = {{Understanding Concurrent Transmissions over Ultra-Wideband Complex Channels}},
  booktitle = {Proc. of the 22nd ACM International Conference on Embedded Networked Sensor Systems ({SenSys})},
  _booktitle = {Proc. of the 22nd ACM SenSys Conference},
  pages = {757--770},	
  location = {Hangzhou, China}, 
  month = nov,	
  year = {2024},
  publisher = {ACM},
  doi = {10.1145/3666025.3699372},	
}

@Book{Goodfellow2016,
  author = {Ian Goodfellow and Yoshua Bengio and Aaron Courville},
  title = {{Deep Learning, Section 7.8 (Early Stopping)}},
  publisher = {MIT Press},
  year = {2016},
  month = nov,
  isbn = {978-0-262-03561-3},
}

@Inbook{Prechelt1998,
  author = {Prechelt, Lutz},
  title = {{Early Stopping -- But When?}},
  booktitle = {Neural Networks: Tricks of the Trade}, 
  year = {1998},
  publisher = {Springer Berlin Heidelberg},
  pages = {55--69},
  _isbn = {978-3-540-49430-0},
  doi = {10.1007/3-540-49430-8_3}
}
